\documentclass[twocolumn,letterpaper]{IEEEtran}
\usepackage{multirow}
\usepackage{amsmath,amsfonts}
\usepackage{algorithmic}
\usepackage{algorithm}
\usepackage{array}
\usepackage{textcomp}
\usepackage{stfloats}
\usepackage{bm}
\usepackage{cuted}
\usepackage{subfigure}
\usepackage{wrapfig}
\usepackage{url}
\usepackage{verbatim}
\usepackage{graphicx}
\usepackage{cite}
\usepackage{xcolor}
\usepackage{booktabs}
\usepackage[flushleft]{threeparttable}
\usepackage{bookmark}
\usepackage[most]{tcolorbox}
\usepackage{microtype}
\usepackage{yhmath}
\usepackage{threeparttable}

\def\s{\mathop{\rm s}\nolimits}
\def\c{\mathop{\rm c}\nolimits}
\newcommand{\bs}[1]{\ensuremath{{\boldsymbol{#1}}}}

\DeclareRobustCommand{\xunjie}[1]{{\textcolor{black}{#1}}}

\begin{document}

\title{A Foot Resistive Force Model for Legged Locomotion on Muddy Terrains}

\author{Xunjie Chen\thanks{X. Chen, X. Huang,  J. Shan, and J. Yi are with the Department of Mechanical and Aerospace Engineering, Rutgers University, Piscataway, NJ 08854 USA (email: xc337@rutgers.edu; xh301@scarletmail.rutgers.edu; jshan@soe.rutgers.edu; jgyi@rutgers.edu). ({\em Corresponding author: Jingang~Yi})}, Liuyin Wang\thanks{L. Wang and Y. Shen are with the Department of Electrical and Biomedical Engineering, University of Nevada, Reno, NV 89557 USA (liuyinw@unr.edu; ytshen@unr.edu).}, Xinyan Huang, Jerry Shan, Yantao Shen, and Jingang Yi}


\maketitle

\begin{abstract}
Legged robots \xunjie{face significant} challenges in \xunjie{moving and navigating} on deformable and highly yielding terrain such as mud. We present a resistive force model for legged foot-mud interactions. The model captures rheological behaviors such as visco-elasticity, thixotropy of the mud suspension and retractive suction. One attractive property of this new model lies in its effective, uniform formulation to provide underlying physical interpretation and accurate resistive force predictions. We further take advantage of the \xunjie{resistive} force model to design a new morphing robotic foot for effective and efficient legged locomotion. We conduct extensive experiments to validate the force model and the results demonstrate that the morphing foot enhances not only the locomotion mobility but also energy-efficiency of walking in mud. The new resistive force model can be further used to develop data-driven simulation and locomotion control of legged robots on muddy terrains.
\end{abstract}


\section{Introduction}
\label{sec:intro}

Legged robots are increasingly used to navigate in \xunjie{natural} environments with deformable, high-yielding terrain such as mud. It is however challenging to generate energy-efficient, stable gait over muddy terrain~\cite{AguilarRPP2016,GodonFRAI2023} because mud propulsive response significantly varies by mud substrate composition and water content. Most developed foot resistive force theories \xunjie{(RFT)} \xunjie{have been developed and mainly focused} on dry granular media (e.g., sand)~\cite{li2013terradynamics,treersRAL2021granular,ChenICRA24,zhu2025JBE,ChenTMECH2025Sand}. Legged robot locomotion on muddy terrain is, however, rarely explored. Unlike granular media, mud rheology is \xunjie{difficult to predict} from its components~\cite{coussot1994behavior}. Constitutive models have been developed to describe complex mud rheology in terms of visco-elasticity and thixotropy~\cite{mewis2009thixotropy,larson2015constitutive}. However, they are not ready to be applied \xunjie{for} robot locomotion. \xunjie{Basic} kinematics and kinetics principles need to be investigated and developed for general robot-mud interaction force modeling.

A few studies \xunjie{have been} reported on legged robot locomotion on muddy terrain. In~\cite{renAIM2013experimental}, the locomotion performance of a quadruped equipped with elliptic-curved legs \xunjie{was} analyzed in terms of propulsive efficiency and optimal gait speed. A simple robophysics model of a flipper-driven robot was proposed in~\cite{liuRAL2023adaptation} to reveal two distinct locomotion failure mechanisms, slippage and entrapment. Few works \xunjie{have} addressed the modeling of mud behavior under vertical loads of foot intrusion with featured characteristics such as hysteresis and water content-dependent mud strength~\cite{GodonRAL2022,ChenAIM24}. Qualitative relationships between mud propulsive/resistive response (e.g., maximum mud strength and suction) and locomotion velocity, water content, and foot shapes \xunjie{have not been fully} understood. Real-time, proprioceptive sensing brings additional challenges to \xunjie{achieving} efficient, stable locomotion control on muddy terrains~\cite{LiuRSS2025,godonTMECH2025walking}. Deformable foot design has demonstrated the advantages of enhanced stable contact on rocks and slopes~\cite{ranjanTMECH2024design}. Multi-segmented bird-inspired feet \xunjie{have been} designed to resist slipping and sinking on soft sandy terrains~\cite{ChatterjeeICRA2023}. For mud locomotion, the work in~\cite{godonBB2024robotic} proposed an ungulate-inspired soft foot with split hooves and tests on a quadruped showed advantages in reducing sinking and energy cost. However, there is no systematic modeling and guidance available yet for the above-mentioned robotic foot design.

In this paper, we propose a new resistive force model of the foot-mud interactions. The new model captures the mud rheology behaviors in three-dimensional (3D) space, including visco-elasticity, thixotropy, and hysteresis effects. These behaviors are directly related to the foot intrusion/retraction velocity and acceleration. We explicitly introduce a structure parameter related to the microstructure state of mud substrate to capture the mud yield strength and stress relaxation. A pressure model is also included by evaluating the sealing state of the cavity underneath the foot contact interface, and therefore, predicting the suction forces during the foot retracting stage. Using the new foot-mud interactions model, we further design a morphing foot with variable stiffness mechanisms by passively enlarging the contact surface in foot intrusion process and quickly retracting the supporting pads in \xunjie{the} foot swing phase, thus reducing the mud suction force significantly. We conduct extensive experiments to validate the proposed resistive force model. Stepping and walking locomotion experiments with new morphing foot are also conducted and compared with regular foot to demonstrate the advantages of mobility in terms of preventing appendage entrapment and reducing energy cost.

The main contributions of this work are twofold. The proposed resistive force model is one of the first such developments that are built on physical principles and provide accurate foot-mud interaction force predictions by considering variations of physical conditions. Compared with the existing work~\cite{GodonRAL2022, ChenAIM24}, enhanced developments \xunjie{of the present work} are significant in several aspects such as \xunjie{the extension to} 3D \xunjie{forces} and the capture of intrinsic mud thixotropic properties. The model leverages terrain knowledge of mud characteristics and can be used as an effective computational tool. Second, based on the model, we have developed a morphing foot mechanism that counters sinking and appendage entrapment in mud. This new foot design not only enhances the locomotion mobility but also improves energy efficiency.


\section{Foot-Mud Interaction Model}
\label{sec:model}

\subsection{Foot-Mud Interactions}

\begin{figure}[h!]
	\centering
		\includegraphics[width=2.9in]{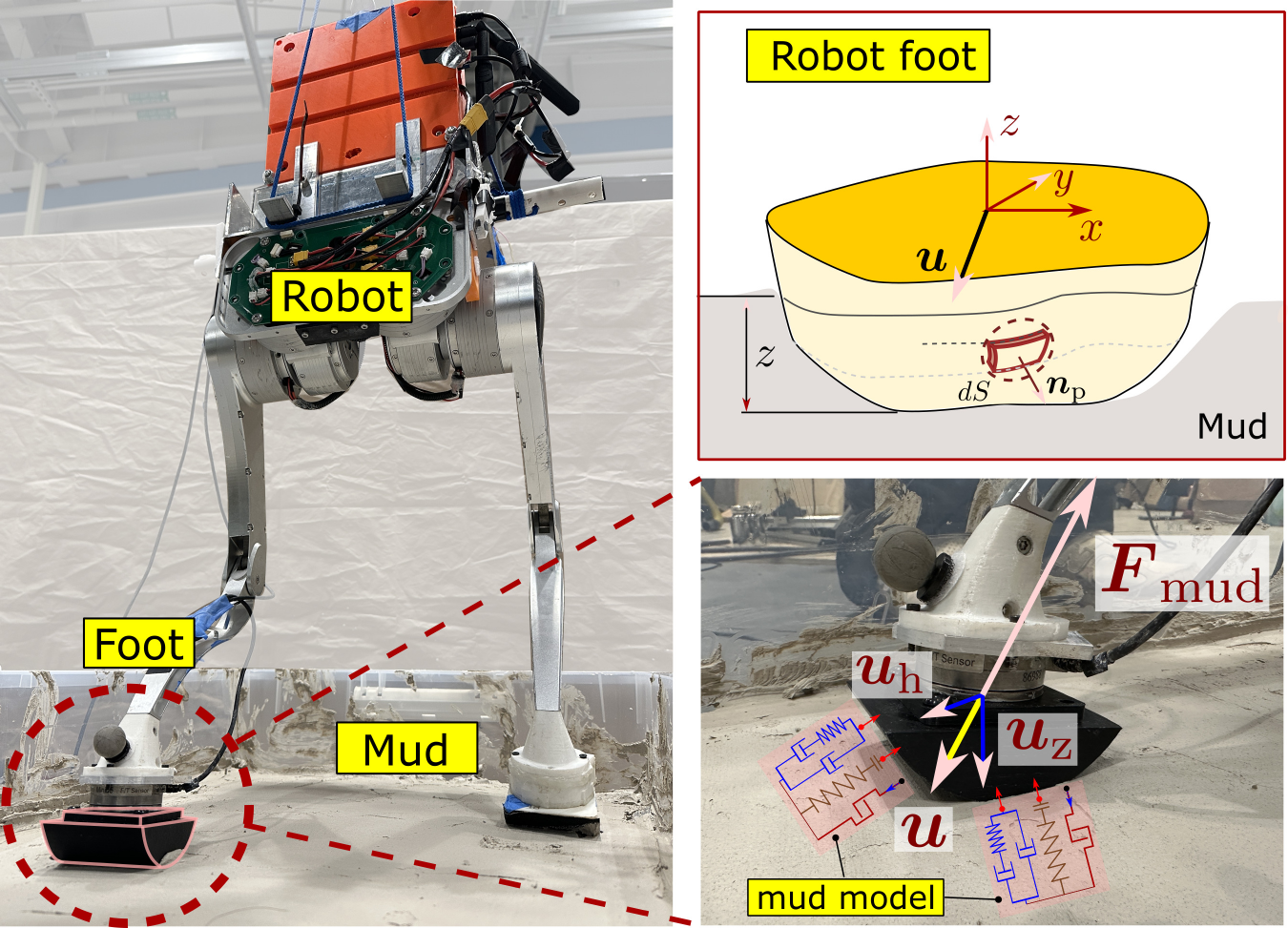}
\vspace{-1mm}
	\caption{A bipedal robot stepping on muddy terrain and a schematic of robot foot–mud contact. The horizontal and vertical projections of the foot are denoted by $\bs{u}_{\mathrm{h}}$ and $\bs{u}_{z}$, respectively.}
		\label{fig:probConfig}
\vspace{-1mm}
\end{figure}

Fig.~\ref{fig:probConfig} shows a robotic foot stepping on \xunjie{even} muddy terrain. An inertial frame $\mathcal{N}(x,y,z)$ is selected on the terrain and the foot velocity is denoted as $\bs{u}=[u_x \, u_y \, u_z]^\top$. For an arbitrary shape of the foot, the resultant reaction force is denoted by $\bs{F}_{\mathrm{mud}}=[F_x \,F_y \,F_z]^\top$, where $F_x$, $F_y$, and $F_z$ represent the longitudinal, lateral and vertical forces, respectively. We focus on the longitudinal and lateral forces in the horizontal plane (i.e., $xoy$-plane), where the foot bulldozes the media leading to the front surface of the foot.

\begin{figure}[h!]
	\centering
		\includegraphics[width=3.5in]{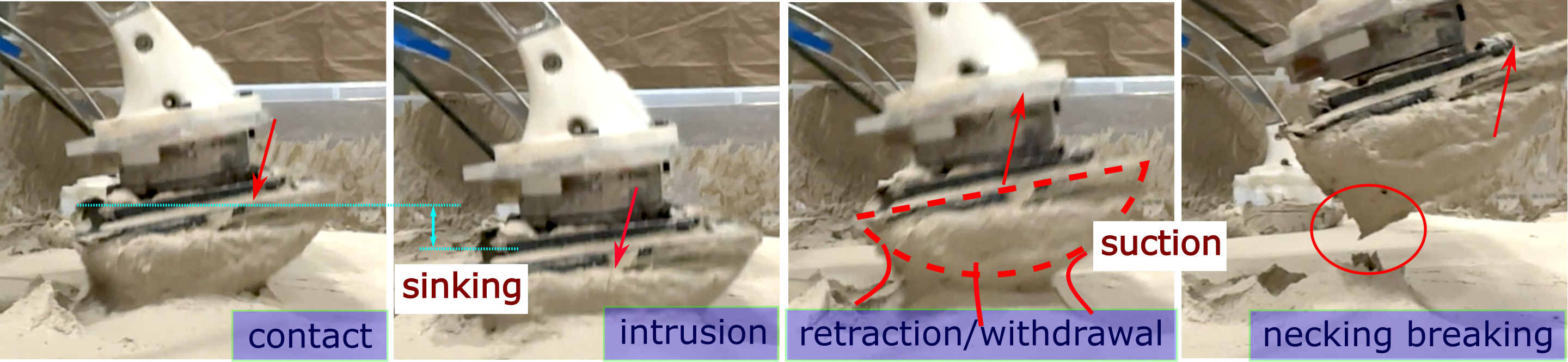}
	\caption{Sequential snapshots of the foot intrusion and retraction over mud.}
		\label{fig:footSnapshot}
\vspace{-5mm}
\end{figure}

Fig.~\ref{fig:footSnapshot} shows a set of \xunjie{snapshots} of foot-mud interactions. As the foot penetrates into the media, the mud substrates are squeezed and compacted underneath the foot, which provides the supporting force. Different from the dry granular media, once the foot retracts from mud, the media generates significant suction force that resists the foot's motion. Finally, when foot is leaving the mud media, the necking effect \xunjie{generates} residual resistive force. We next propose a unified mud constitutive model that represents the \xunjie{principal} force generation mechanism in any directions.

\subsection{Foot-Mud Constitutive Model}
\label{subsec:stress}
\begin{figure}[h!]
	\hspace{-2mm}
	\subfigure[]{
		\includegraphics[width=1.52in]{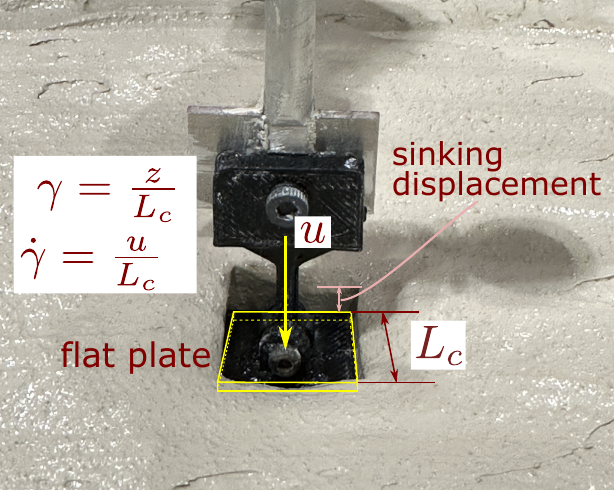}
		\label{fig:simpleIntrusion}}
	\hspace{-2.5mm}
	\subfigure[]{
		\includegraphics[width=1.86in]{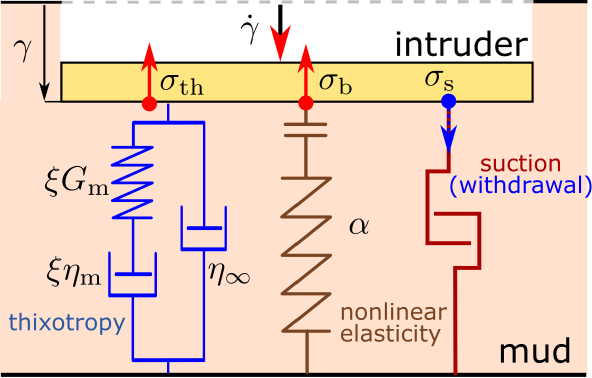}
		\label{fig:mud_model}}
	\caption{(a) A simple flat-plate intrusion into mud. (b) The schematic of the foot-mud constitutive model.}
\end{figure}

Fig.~\ref{fig:simpleIntrusion} shows a simple plate-intrusion setup into the mud and Fig.~\ref{fig:mud_model} illustrates the corresponding schematic of the mud model for the principal stress on an infinitesimal small intruder in mud. We denote a dimensionless displacement and rate of the intruder by $\gamma=\frac{z}{L_c}$ and $\dot{\gamma}=\frac{u}{L_c}$, respectively, where $z$ ($u=\dot{z}$) is the intruder depth (velocity) and $L_c$ is the characteristic length of the plate.
\xunjie{We assume a locally flat surface, and therefore, the intrusion depth is uniform.}
The mud model consists of three elementary parts, namely, the resistive model represented by the nonlinear elastic spring element, the mud thixotropic model by a combination of spring and damping elements, and a suction model element (only for foot retraction).
Including both the intrusion and retraction process, the total stress, denoted by $\sigma_{\text{tot}}$, on the intruder is expressed as
\begin{equation}
\label{eqn:totalStress}
\sigma_{\text{tot}} = \sigma_{\mathrm{b}} + \sigma_{\mathrm{th}}  + H_{\nu}(\dot{\gamma})\sigma_{\mathrm{s}},
\end{equation}
where $\sigma_{\mathrm{b}}$ is the immediate resistive stress responding to volume change of mud media beneath the foot, $\sigma_{\mathrm{th}}$ is the mud thixotropic stress, and $\sigma_{\mathrm{s}}$ is the suction stress. The continuous function $H_{\nu}(\dot{\gamma})=\frac{1}{2}\left(1-\tanh\frac{\dot{\gamma}}{\nu}\right)$ is an index to distinguish the intrusion ($H_{\nu}(\dot{\gamma})\approx 0$) and retraction ($H_{\nu}(\dot{\gamma}) \approx 1$) motion, where $\nu$ is a small constant. Next, we compute each stress component in~\eqref{eqn:totalStress}.
\subsubsection{Immediate resistance $\sigma_{\mathrm{b}}$}
We consider a nonlinear elasticity model for immediate resistive stress~\cite{ChenAIM24}
\begin{equation}
\label{eqn:sigma_b}
  \sigma_{\mathrm{b}} = \alpha \gamma ^ {n},
\end{equation}
where $\alpha$ is defined as the stiffness related to direct depth of intrusion in the mud and $n \in (0, 1)$ is a constant.

\subsubsection{Mud thixotropic stress $\sigma_{\mathrm{th}}$}
A viscoelastic thixotropic model is used to capture the mud suspensions’ rheological behavior~\cite{larson2019review}. The stress evolution uses a structural parameter $\xi  \in (0,1)$ to relate the mud micro-structure to the macroscopic viscosity property~\cite{dullaert2006structural}. The thixotropic stress is described as
\begin{subequations}
\label{eqn:thixotropy}
  \begin{align}
  & \dot{\sigma}_{\mathrm{th}}= \frac{1}{\lambda}f_{\xi}(\dot{\gamma},\ddot{\gamma})-\frac{1}{\lambda}\sigma_{\mathrm{th}}, \label{eqn:sigma_th}  \\
  & \dot{\xi} =k_a(1-\xi) - k_r|\dot{\gamma}|\xi, \label{eqn:xi_ODE}
\end{align}
\end{subequations}
where $k_{a}$ and $k_{r}$ are constant, function $f_{\xi}(\dot{\gamma}, \ddot{\gamma})=(\eta_{\infty} + \xi\eta_{\mathrm{m}}) \dot{\gamma} + \lambda \eta_{\infty}\ddot{\gamma}$ considers intruder velocity and acceleration to bring the foot inertial effect into the mud rheological behavior, $\lambda = \eta_\mathrm{m}/G_\mathrm{m}$ is a relaxation time constant, and $\eta_\mathrm{m}$, $\eta_{\infty}$, and $G_\mathrm{m}$ are the structural viscosity, shear viscosity, and the elastic modulus of mud, respectively.
In~\eqref{eqn:xi_ODE}, $\xi$ is associated with the evolution of substrate microstructure, namely, ``aging'' (building-up process) and ``rejuvenation'' (flow-induced breakdown), represented respectively by $k_{a}$ and $k_{r}$. Aging happens if the mud is undisturbed and rejuvenation when the mud is sheared. It is straightforward from~\eqref{eqn:xi_ODE} to obtain the steady-state $\xi_{\mathrm{ss}} = k_a/(k_a + k_r|\dot{\gamma}|)=\dot{\gamma}_c/(\dot{\gamma}_c+|\dot{\gamma}|)$, $\dot{\gamma}_c = \frac{k_a}{k_r}$. The corresponding steady-state thixotropic stress $\sigma_{\mathrm{th},\mathrm{ss}}$ and yield stress $\sigma_Y$ are estimated as
\begin{equation}
\label{eqn:yieldStress}
\sigma_{\mathrm{th},\mathrm{ss}} = (\eta_{\infty} + \xi_{\mathrm{ss}} \eta_{\mathrm{m}}) \dot{\gamma}, \; \sigma_Y = (\eta_{\mathrm{m}}-\eta_{\infty})\dot{\gamma}_c,
\end{equation}
where $\dot{\gamma}_c$ is a critical displacement rate at which the mud starts to fully fluidize.

\subsubsection{Suction pressure $\sigma_{\mathrm{s}}$}
\xunjie{Upon retraction, a cavity underneath the intruder forms due to insufficient mud fluid replenishment. This leads to a significant pressure gap compared to the ambient, resulting in the suction effect.
} The \xunjie{rate of the suction pressure change} $\dot{\sigma}_{\mathrm{s}}$ is assumed proportional to the cavity-volume change \xunjie{related to mud flow that} is proportional to the pressure gap. We therefore describe the dynamic suction pressure as the resultant ``build'' and ``leak'' effects as
\begin{equation}
\label{eqn:suction_ODE}
\dot{\sigma}_{\mathrm{s}} = \underbrace{-\frac{\phi_{\varepsilon}(\gamma)}{\tau_{\mathrm{build}}}(\sigma_{\mathrm{s}}+\sigma_Y)}_{\mathrm{build}}
-\underbrace{\frac{1-\phi_{\varepsilon}(\gamma)}{\tau_{\mathrm{leak}}}\sigma_{\mathrm{s}}}_{\mathrm{leak}},
\end{equation}
where $\tau_{\mathrm{build}}$ and $\tau_{\mathrm{leak}}$ are characteristic time scales \xunjie{which provide a compact description of the competing suction establishment and break-down process simultaneously.} Similar to $H_{\nu}(\dot{\gamma})$, the displacement-dependent factor, $\phi_{\varepsilon}(\gamma)=\frac{1}{2}\left(1-\tanh \frac{\gamma-\gamma_0}{\varepsilon} \right)$, represents the sealing state at the contact interface, where $\gamma_0$ is the dimensionless displacement when the intruder starts to retract and $\varepsilon > 0$ is displacement-scaling constant. When \xunjie{$\phi_{\varepsilon}=1$}, the interface is fully sealed. Otherwise, it is partially leaking with slow refill of mud and air. Unlike constant $\nu$ in $H_{\nu}(\dot{\gamma})$, $\varepsilon$ varies with mud properties like substrate composition and water content and is identified via experiments. In~\eqref{eqn:suction_ODE}, $-\sigma_Y$ is the yield suction pressure beyond which mud strength is so weak that suction fails to hold \xunjie{and $\phi_{\varepsilon}$ reflects how fast this transient state happens. Generally, the proposed suction model~\eqref{eqn:suction_ODE} provides a computationally tractable representation that can match the experimental profile, rather than directly measuring the cavity pressure.}


\subsection{3D Foot-Mud Interaction Forces}
\label{subsec:FootMudInteraction}

\begin{figure}[h!]
	\centering
	\includegraphics[width=2.95in]{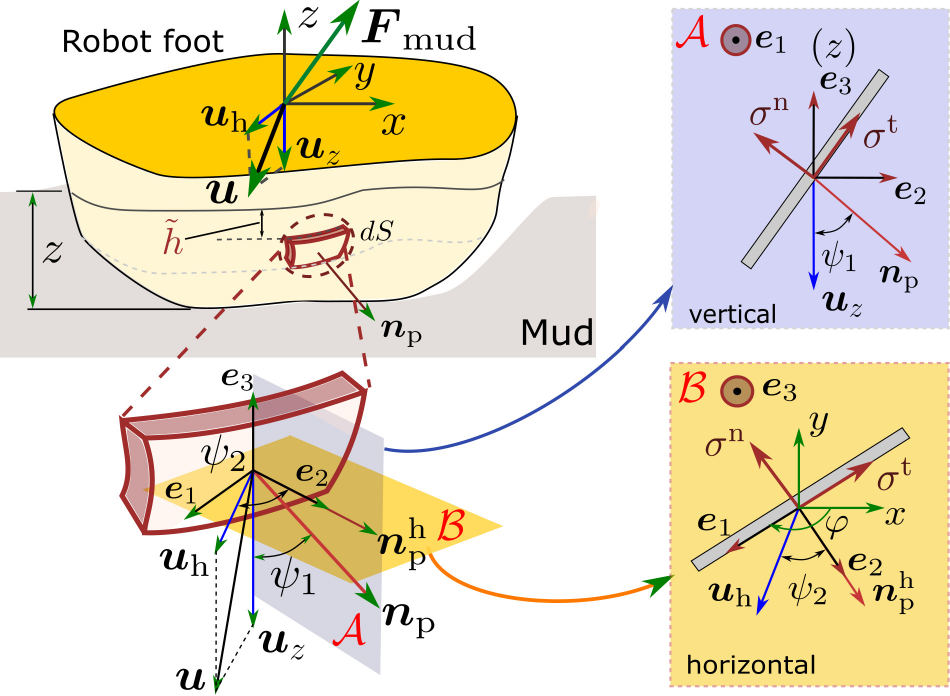}
	\caption{The general 3D configuration of an arbitrary robotic foot consisting of infinitesimal plates.}
	\label{fig:3D_config}
\vspace{-2mm}
\end{figure}

For an arbitrary foot shape, the net 3D mud resistive-force model is obtained by integration of surface stresses on infinitesimal plates. Fig.~\ref{fig:3D_config} illustrates the general 3D configuration of the robotic foot in $\mathcal{N}$. We consider a sinking downward velocity $\bs{u}_{z}$ and a sliding horizontal velocity $\bs{u}_{\mathrm{h}}$ on the highlighted infinitesimal foot surface $dS$ with the normal vector $\bs{n}_{\mathrm{p}}$ pointing outward. A local plate coordinate system is described by unit vectors $(\bs{e}_1,\bs{e}_2,\bs{e}_3)$, where $\bs{e}_3$ is upward along the $z$-axis, $\bs{e}_1 = \left(\bs{n}_\mathrm{p} \times \bs{e}_3\right)/\|\bs{n}_\mathrm{p} \times \bs{e}_3\|$, and $\bs{e}_2 = \bs{e}_3 \times \bs{e}_1$. The vertical plane formed by $\bs{n}_{\mathrm{p}}$ and $\bs{u}_{z}$ is defined as $\mathcal{A}$, while the horizontal plane by $\bs{e}_1$ and $\bs{e}_2$ as $\mathcal{B}$; see Fig.~\ref{fig:3D_config}. We compute the resistive forces on $\mathcal{A}$ and $\mathcal{B}$.

We denote $\psi_{1}$ as the angle between $\bs{n}_{\mathrm{p}}$ and $\bs{u}_{z}$ on $\mathcal{A}$. We further denote $\varphi$ as the angle between $\bs{n}_{\mathrm{p}}^{\mathrm{h}}$ and the $x$-axis and $\psi_2$ as the angle between the $\bs{e}_2$-axis and $\bs{u}_{\mathrm{h}}$ on $\mathcal{B}$. The force $d\bs{F}=[dF_x \,dF_y\,dF_z]^\top$ on $dS$ is computed by
\begin{subequations}
\label{eqn:dF}
    \begin{align}
      & dF_{x} = (-\sigma^{\mathrm{n}}_{\mathrm{h}}\c_{\varphi}+\sigma_{\mathrm{h}}^\mathrm{t}\s_{\varphi})dS, \; dF_{y} = (\sigma^\mathrm{n}_{\mathrm{h}}\s_{\varphi}+\sigma^\mathrm{t}_{\mathrm{h}}\c_{\varphi})dS, \label{eqn:df_y}\\
      & dF_{z} = (\sigma^\mathrm{n}_{\mathrm{v}}\c_{\psi_1}+\sigma^\mathrm{t}_{\mathrm{v}}\s_{\psi_1})dS, \label{eqn:df_z}
    \end{align}
\end{subequations}
where $\sigma^{\mathrm{n}}_i$ ($\sigma^{\mathrm{t}}_i$), $i= \mathrm{h,v}$, are the normal (tangential) stress in the horizontal (h) and vertical (v) planes. In~\eqref{eqn:dF}, notation convention $\s_\varphi:=\sin \varphi$ and $\c_\varphi:=\cos \varphi$ is used for $\varphi$ and other angles throughout this paper.

We use~\eqref{eqn:totalStress} to compute the horizontal stress $\sigma_{\mathrm{h}}$ (as function of rate $\dot{\gamma}_h=|\bs{u}_{\mathrm{h}}|/L_c$) and vertical stress $\sigma_{\mathrm{v}}$ (as function of rate $\dot{\gamma}_v=|\bs{u}_{\mathrm{v}}|/L_c$) . Considering \xunjie{the non-aligned
shear motion with orientation by $\psi_{1}$ and $\psi_2$}, two dimensionless scaling factors are used to weight the normal and tangential \xunjie{stress projection:}
\begin{equation}
\label{eqn:scaling}
\sigma_{\mathrm{v}}^{i}=f_{i}(\psi_1)\sigma_{\mathrm{v}}, \; \sigma_{\mathrm{h}}^{i}=f_{i}(\psi_2)\sigma_{\mathrm{h}}, \; i=\mathrm{t, n},
\end{equation}
where $f_{i}(\psi_{1})$ and $f_{i}(\psi_{2})$ are the scaling factors that are determined by experiments.
We propose to use $f_{\mathrm{n}}(\psi_{i}) = \frac{1}{2}(1+\cos{2\psi_{i}})$ and  $f_{\mathrm{t}}(\psi_i) = \frac{1}{2}(1-\cos{2\psi_i})$, $i=1,2$\xunjie{, to extend the local mud stress to the resultant 3D force integration.} Detailed calibrations are discussed in Section~\ref{sec:Results}.

Finally, we integrate~\eqref{eqn:dF} to obtain the resultant resistive force $\bs{F}_{\mathrm{mud}}$ on the robotic foot, namely, \xunjie{$\bs{F}_{\mathrm{mud}}=\int_{S}d\bs{F}$}.

\section{\xunjie{Resultant Forces and Robotic Foot}}
\label{force}


\subsection{Force Calculation for Regular Shape Foot}

We first present solutions of~\eqref{eqn:thixotropy} and~\eqref{eqn:suction_ODE}. Considering the \xunjie{quasi-static condition of the foot motion and neglecting the acceleration term of $f_{\xi}(\dot{\gamma},\ddot{\gamma})$ in mud thixotropy, i.e.,} $\ddot{\gamma} = 0$, the solution of~\eqref{eqn:sigma_th} is expressed as
\begin{equation}
\label{eqn:sigma_th_approx}
    \sigma_{\mathrm{th}}=\left(1-e^{-\frac{t}{\lambda}}\right)\sigma_{\mathrm{th,ss}},
\end{equation}
where $\sigma_{\mathrm{th,ss}}$ is given by~\eqref{eqn:yieldStress}. We also provide the suction pressure in~\eqref{eqn:suction_ODE} as
\begin{equation}\label{eqn:suction_p_approx}
    \sigma_{\mathrm{s}}=\left(1-e^{-a\Delta t}\right)\sigma_{\mathrm{s,ss}},~ \Delta t = t-t_{\mathrm{w}},
\end{equation}
where $t_{\mathrm{w}}$ is the time moment when the foot retraction begins, and the steady-state suction stress $\sigma_{\mathrm{s,ss}}$ and time constant $a$ are given by
\begin{equation}
\label{eqn:sigma_steadySuction}
\sigma_{\mathrm{s,ss}}= -\frac{\phi(\gamma)}{a\tau_{\mathrm{build}}}\sigma_Y,\;
  a=\frac{\phi(\gamma)}{\tau_{\mathrm{build}}} + \frac{1-\phi(\gamma)}{\tau_{\mathrm{leak}}}.
  \vspace{-1mm}
\end{equation}
From~\eqref{eqn:sigma_steadySuction}, when the cavity at the interface is completely sealed, $\phi(\gamma)=1$, $\sigma_{\mathrm{s,ss}}=-\sigma_Y$, that is, the suction pressure rapidly increases to the mud limitation stress. With the full break of sealing, $\phi(\gamma)=0$ and $\sigma_{\mathrm{s,ss}}=0$. \xunjie{Details of the above derivations and corresponding sensitivity and identifiability of mud-intrinsic parameters are found in the Supplementary Materials.}

We now present closed-form calculations of resistive forces $\bs{F}_{\mathrm{mud}}$ for three representative foot shapes: flat, semi-cylindrical, and semi-spherical feet. The details of derivations and calculations are given in the Supplementary Materials. Fig.~\ref{fig:forcesample} shows the schematic of the semi-cylindrical- and semi-spherical-foot in the mud. The angle between foot horizontal velocity $\bs{u}_{\mathrm{h}}$ and the $x$-axis is denoted by $\varphi$ (see Fig.~\ref{fig:3D_config}) and the depth of the foot bottom tip into mud is denoted by $z$.

\begin{figure}[h!]
	\centering
	\includegraphics[width=3.5in]{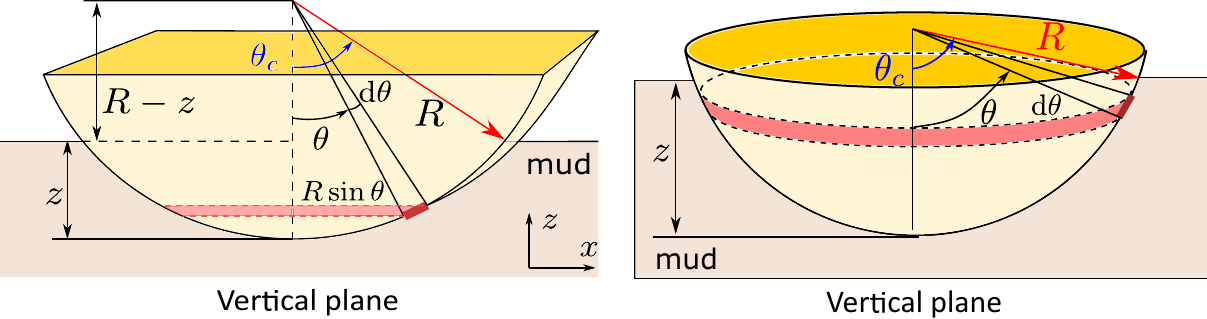}
	\caption{Schematic diagrams for 3D resultant force calculation for semi-cylindrical foot (left) and semi-spherical foot (right).}
	\label{fig:forcesample}
\end{figure}

\subsubsection{Flat foot}
For a rectangular-shape flat foot with length $L$ and width $W$, we obtain
\begin{subequations}
\label{eqn:Flat_Forces}
  \begin{align}
   & F_x = z\left[Lf_{\mathrm{n}}(\varphi)\sigma_x +Wf_{\mathrm{t}}(\varphi_c)\sigma_y\right],  \label{eqn:Flat_Fx} \\
   & F_y =z\left[Lf_{\mathrm{t}}(\varphi)\sigma_x+Wf_{\mathrm{n}}(\varphi_c)\sigma_y\right], \; F_z = S_e\sigma_z, \label{eqn:Flat_Fz}
  \end{align}
\end{subequations}
where $S_e=LW$ is the foot effective area, $\varphi_c=\frac{\pi}{2}-\varphi$, $\sigma_{i} = \sigma^{i}_{\mathrm{th}}+\sigma^{i}_{\mathrm{s}} +\sigma_{\mathrm{b}}^i$ for three directions $i=x,y,z$. Each component is computed by the corresponding displacement $\gamma$ and rate $\dot{\gamma}$ by~\eqref{eqn:sigma_b}, ~\eqref{eqn:sigma_th_approx}, and~\eqref{eqn:suction_p_approx}, respectively.

\subsubsection{Semi-cylindrical foot}
With radius $R$ and width $W$, the resultant resistive forces are
\begin{subequations}
\label{eqn:SemiCylind_Forces}
\begin{align}
    \label{eqn:SemiCylind_Fx} & \text{\hspace{-3mm}} F_x =\frac{S_e}{2}\Bigl[\frac{R}{W}\left(\frac{\theta_c}{\s_{\theta_c}}-\c_{\theta_c}\right)f_\mathrm{n}(\varphi)\sigma_x+\frac{1-\c_{\theta_c}}{\s_{\theta_c}}f_{\mathrm{t}}(\varphi_c)\sigma_y\Bigr], \\
    \label{eqn:SemiCylind_Fy} & \text{\hspace{-3mm}} F_y =\frac{S_e}{2}\Bigl[\frac{R}{W}\left(\frac{\theta_c}{\s_{\theta_c}}-\c_{\theta_c}\right)f_\mathrm{t}(\varphi)\sigma_x+\frac{1-\c_{\theta_c}}{\s_{\theta_c}}f_{\mathrm{n}}(\varphi_c)\sigma_y\Bigr], \\
    \label{eqn:SemiCylind_Fz} & \text{\hspace{-3mm}} F_z = \frac{1}{3}S_e \left(2+\c^2_{\theta_c}\right)\sigma_z,
  \end{align}
\end{subequations}
where $\theta_c =\cos^{-1}(1-z/R)$, $S_e=2RW \s_{\theta_c}$ is foot effective area, and $\sigma_{i}, i= x,y,z$, are the same as in~\eqref{eqn:Flat_Forces}.

\subsubsection{Semi-spherical foot}
With radius $R$, the resultant resistive forces are
\begin{subequations}
\label{eqn:SemiSphere_Forces}
  \begin{align}
    \label{eqn:SemiSphere_Fx} & F_x  =\frac{4S_e}{3\pi\s_{\theta_c}}\Bigl(\frac{2\theta_c}{\s_{\theta_c}}-1\Bigr)\c_{\varphi}\sigma_x, \, F_y  = \frac{4S_e}{3\pi\s_{\theta_c}}\Bigl(\frac{2\theta_c}{\s_{\theta_c}}-1\Bigr)\s_{\varphi}\sigma_y, \\
    \label{eqn:SemiSphere_Fz} & F_z = \frac{S_e}{4}\left(2+\c^2_{\theta_c}+\frac{2\c^3_{\theta_c}}{\s_{\theta_c}}-5\frac{\c_{\theta_c}}{\s_{\theta_c}}+\frac{3\theta_c}{\s^2_{\theta_c}}\right)\sigma_z,
  \end{align}
\end{subequations}
where $S_e=\pi R^2 \s^2_{\theta_c}$ is the foot effective area and $\theta_c$ is the same as in~\eqref{eqn:SemiCylind_Forces}.

\subsection{Morphing Foot Design}

The force calculations in~\eqref{eqn:Flat_Forces}-\eqref{eqn:SemiSphere_Forces} clearly indicate that the vertical resultant force $F_z$ is proportional to the foot effective area $S_e$. To generate an effective, sustainable supporting force with restricted sinking, the foot design should have a sufficiently large contact area in intrusion and small effective surface area during retraction gait (i.e., pulling out of mud). Thus, instead of a fixed shape, we propose a foot design that change its morphology depending on whether it penetrates or retracts.

\begin{figure}[h!]
	\centering
		\includegraphics[width=3.5in]{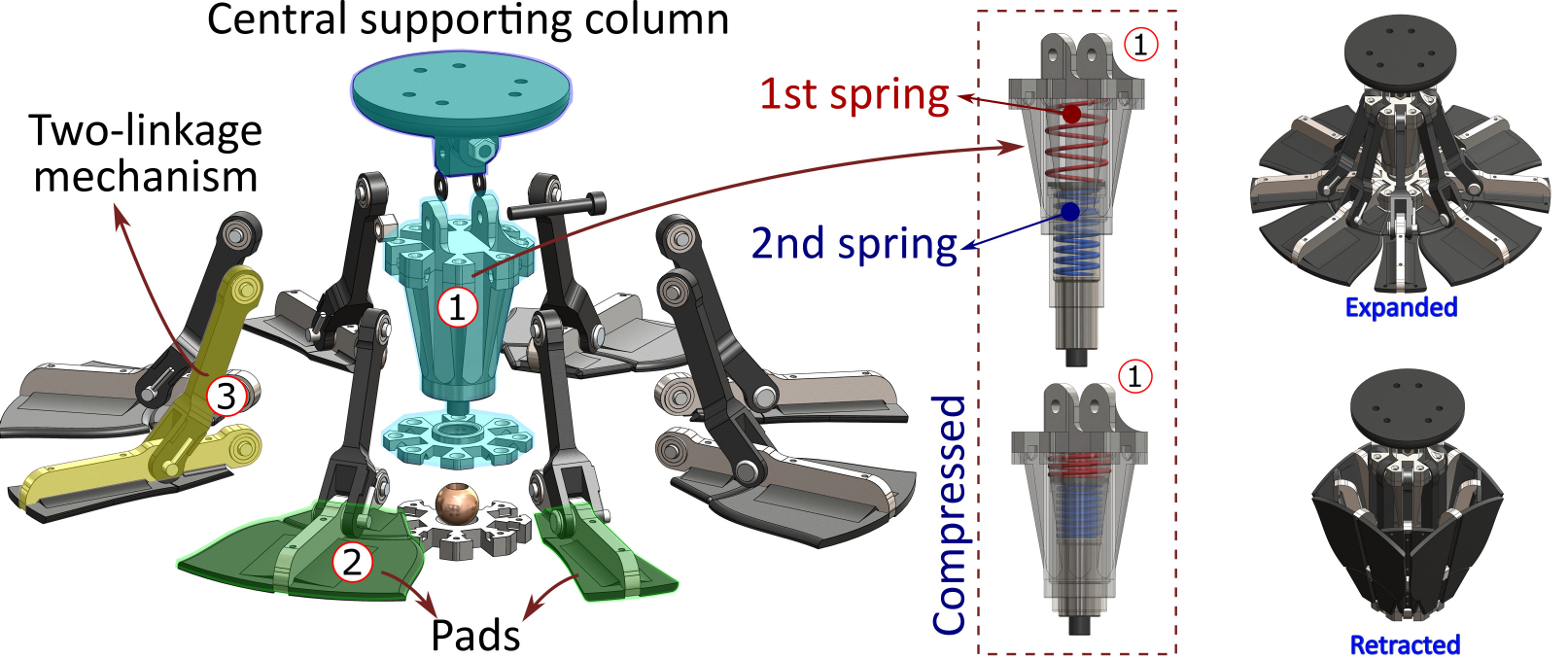}
		\vspace{-5mm}
		\caption{Left: A 3D model of the morphing foot showing three core components. Right: Two modes of the morphing foot. Top: fully-expanded; Bottom: fully retracted.}
		\label{fig:foot}
		\vspace{-1mm}
	\end{figure}

Fig.~\ref{fig:foot} illustrates the morphing-foot design. The morphing mechanism is composed of three components: a central telescope-type supporting column with springs, the surrounding expandable pads, and a set of eight linkages.
\xunjie{The full pad-covered shape is disk-like for structural simplicity and packaging compactness.}
The central supporting column is designed as a two-cascade hollow telescope whose upper end moves freely downward once the bottom end contacts firmly on the ground. Both the upper and lower part of the central column have eight poles evenly to connect a two-linkage mechanism through pin bearings. Eight pads are attached on all lower links so that they expand \xunjie{outward,} \xunjie{ensuring} a large contact surface, as shown in Fig.~\ref{fig:foot}. The geometry of the linkage system is carefully designed such that when the column is completely compacted, the pads spread out (top right in the figure); while the support column is not compressed, the pads curl up tightly pointing almost upward (bottom right in the figure).

The two embedded springs are selected with different stiffnesses and connected in series inside the telescope column: a soft spring in the first upper cascade and a hard one in the second lower cascade. During foot intrusion into the mud, the soft spring is first compressed quickly and then the hard spring takes the further load. This variable-stiffness design provides a desirable actuation property that the pads spread out rapidly even under a small external supporting force. Conversely, the lower links are pulled up rapidly by the compressed hard spring when the foot lifts from mud media and the suction pressure breaks.

\begin{figure*}[t!]
    \centering
	\subfigure[]{
		\label{fig:exp:a}
		\includegraphics[width=1.85in]{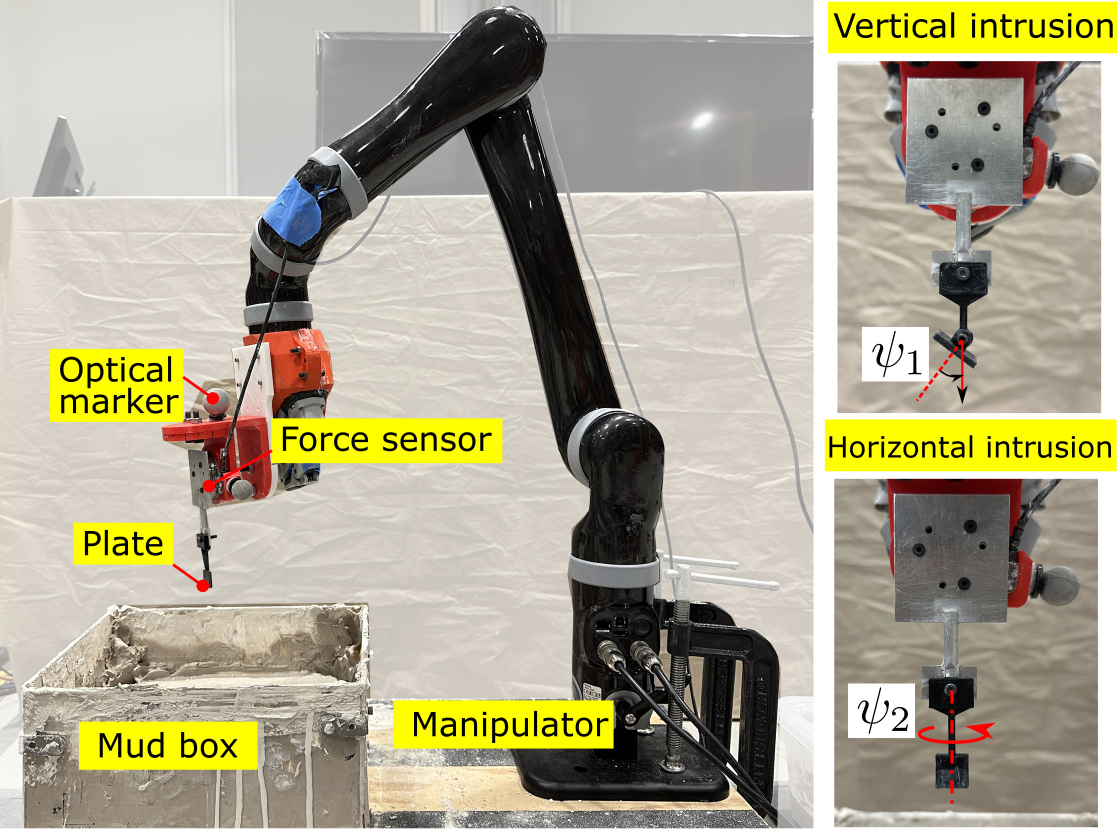}}
	\subfigure[]{
		\label{fig:exp:b}
		\includegraphics[width=2.34in]{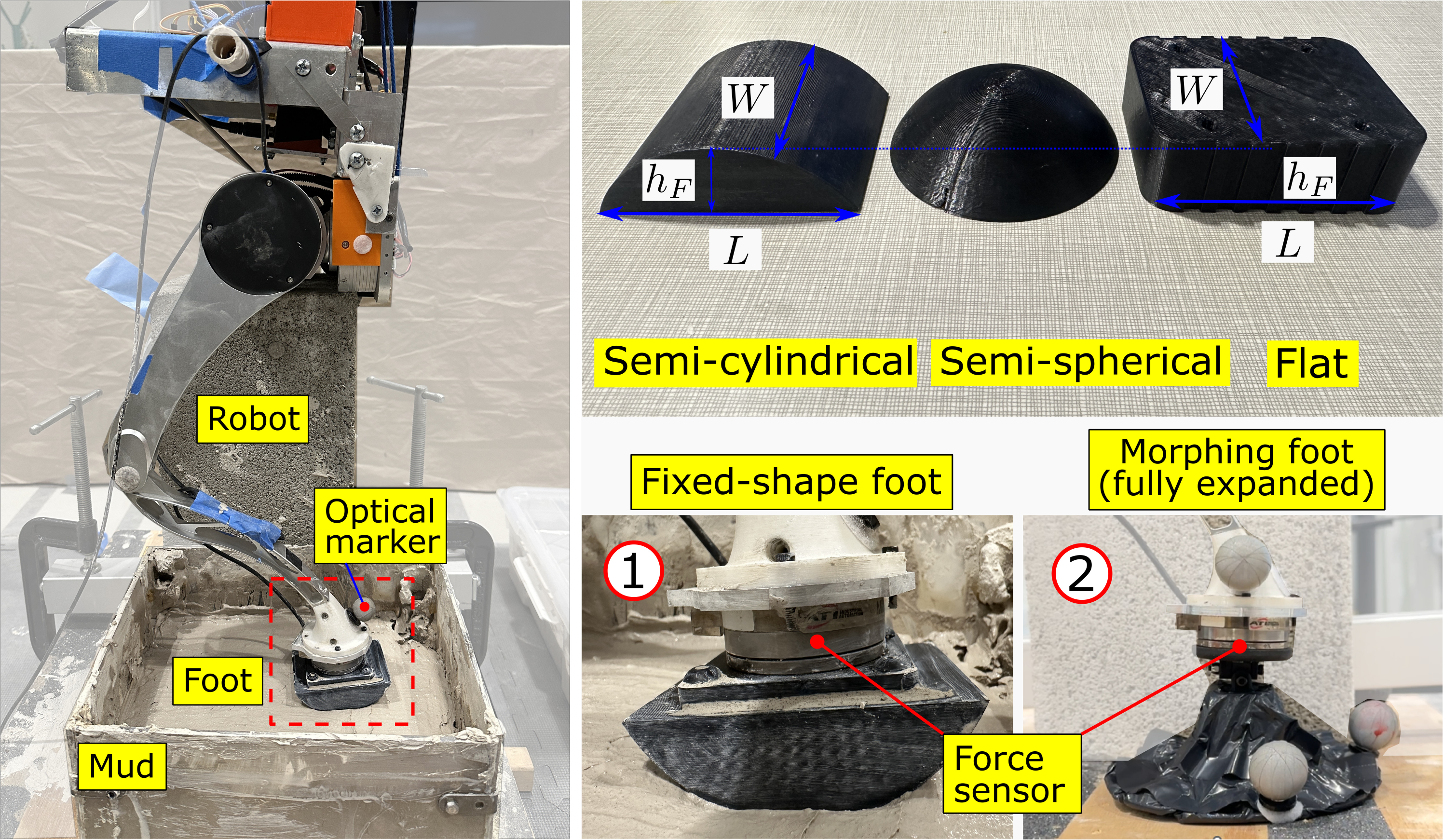}}
	\subfigure[]{
		\label{fig:exp:c}
		\includegraphics[width=2.40in]{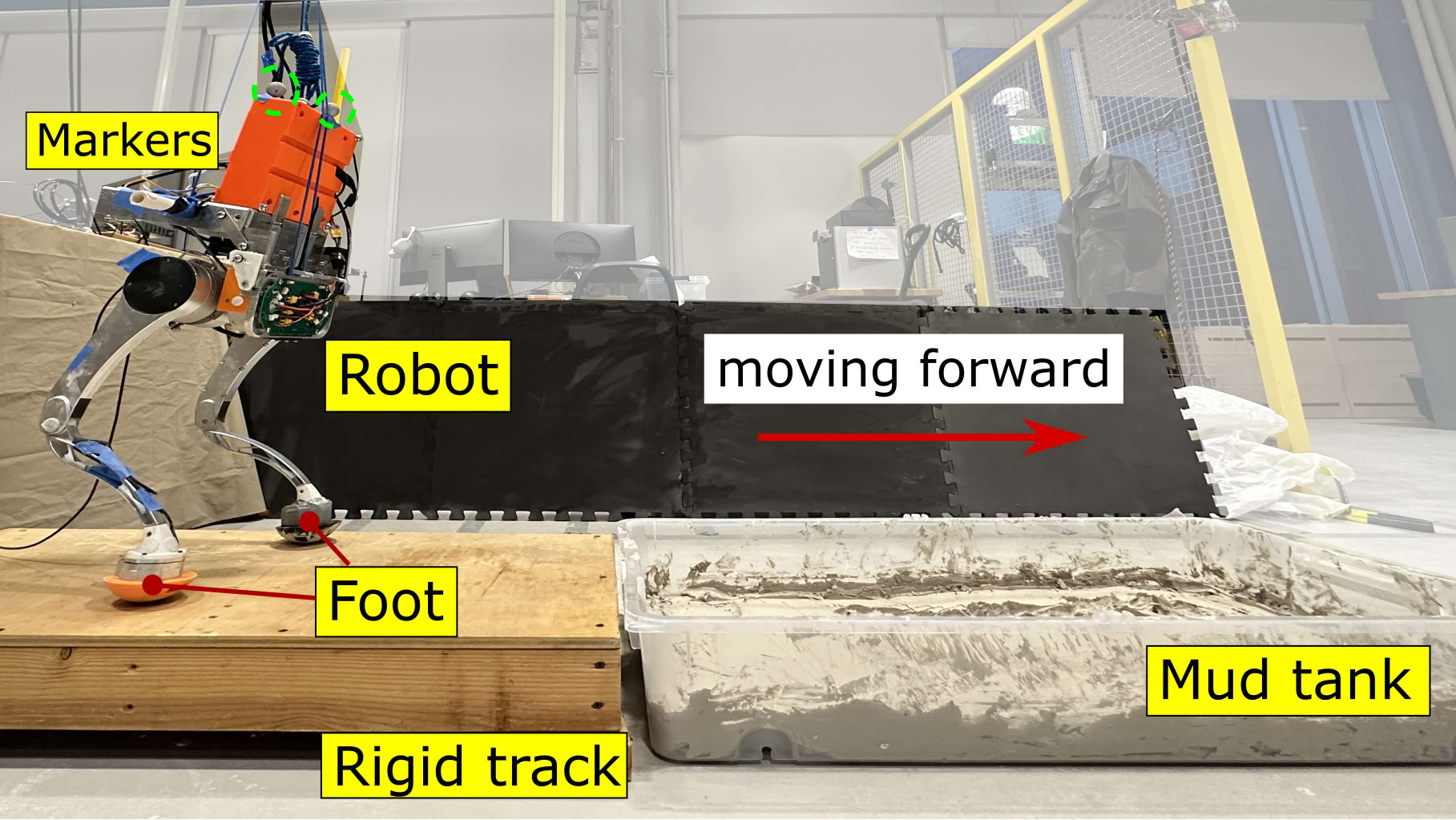}}
	\vspace{-2mm}
	\caption{(a) Experimental setup for vertical and horizontal penetration tests for mud rheology behavior calibrations using a robotic manipulator. (b) Robotic foot penetration tests with fixed shaped feet and the morphing foot. (c) Biped walking tests with various types of feet.}
	\label{fig:exp}
	\vspace{-2mm}
\end{figure*}
\vspace{-2mm}
\section{Experimental Results}
\label{sec:Exp}


\subsection{Experimental Setups}

We \xunjie{prepared} mud substrates by mixing controlled volumes of clay (Sea Mix 6 from Seattle Pottery Supply), sand ($150-600~\mu$m graded standard silica sand from Gilson Inc.), and water in the desired proportions. For all experiments reported in this work, the ratio of clay-to-sand was selected to be $3$-to-$1$ \xunjie{ and the mud water content was defined as the liquid-to-overall volume ratio to provide consistent results that are comparable to the existing work~\cite{liuRAL2023adaptation,ChenAIM24}.} By varying the water content, the mud rheological properties related to mud deformation and yielding were controlled.

We first conducted planar plate intrusion tests to capture the mud rheological behavior for various working conditions (e.g., penetration angles and mud water content). A container with the size of $28 \times 23 \times 15$~cm was used. Fig.~\ref{fig:exp:a} shows the experimental setup. To check the shape-variant effect, three different shapes of the intruders, namely, square (S), semi-circle (SC), and triangle (T), were selected. The penetration angle of the plate was adjusted from $0$ to $90$ deg with a $22.5$-deg increment to obtain the tangential and normal stresses applied on the plate. In an intrusion test, the plate was first controlled to penetrate in mud with a prescribed constant velocity and then stopped to allow capture of the relaxation of the reaction forces. Finally, the plate was retracted in the reversed direction with the same velocity as intrusion. The water content of mud (denoted by $W$) for all tests \xunjie{was} from $15\%$ to $35\%$. For each working condition, three trials were repeated and the mud surface was flattened before each trial. Table~\ref{tab:plate_experiments} lists the planar intrusion tests for the different conditions.

\renewcommand{\arraystretch}{1.2}
\setlength{\tabcolsep}{0.055in}
\begin{table}[h!]
	\centering
	\caption{List of planar intrusion experiments (HI/VI: horizontal/vertical intrusion; $W$: water content)}
\vspace{-2mm}
	\label{tab:plate_experiments}
	\begin{tabular}{ccccc}
	\toprule[1.2 pt]
	{\bf Exps.} & {\bf Description} & {\bf $W$ ($\%$)} & {\bf Penetration angle} & {\bf Plate shape} \\
			\midrule
		E1 & VI & $25$ & $0^{\circ}$ & S, SC, T \\
		E2 & HI & $25$ & $0^{\circ}$ & S, SC, T. \\
		E3 & VI & $15,25,35$ & $0, 22.5, 45, 67.5, 90^{\circ}$ & S \\
        E4 & HI & $15,25,35$ & $0, 22.5, 45, 67.5, 90^{\circ}$ & S \\
	\bottomrule[1.2pt]
	\end{tabular}
\vspace{-3mm}
\end{table}

\begin{figure*}[t!]
	\hspace{-2mm}
	\subfigure[]{
		\includegraphics[width=1.67in]{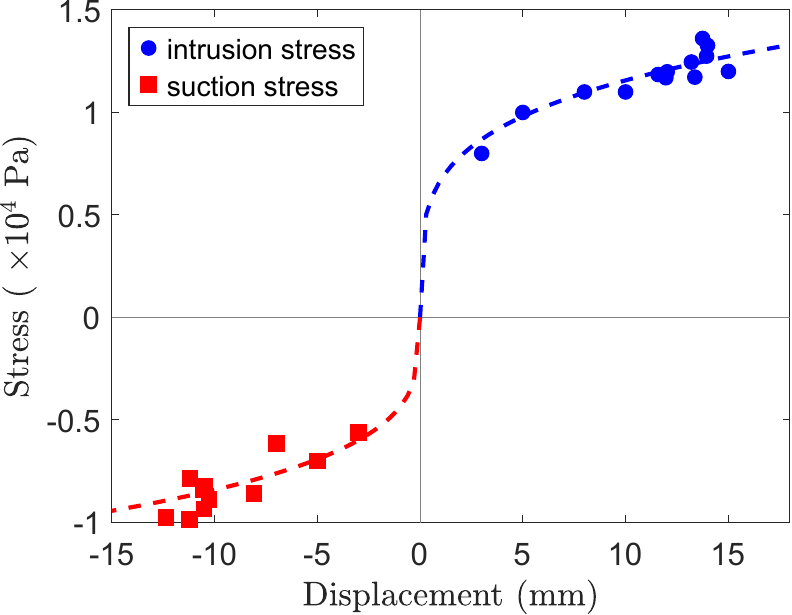}
		\label{fig:HIVI:a}}
	\hspace{-4mm}
	\subfigure[]{
		\includegraphics[width=1.67in]{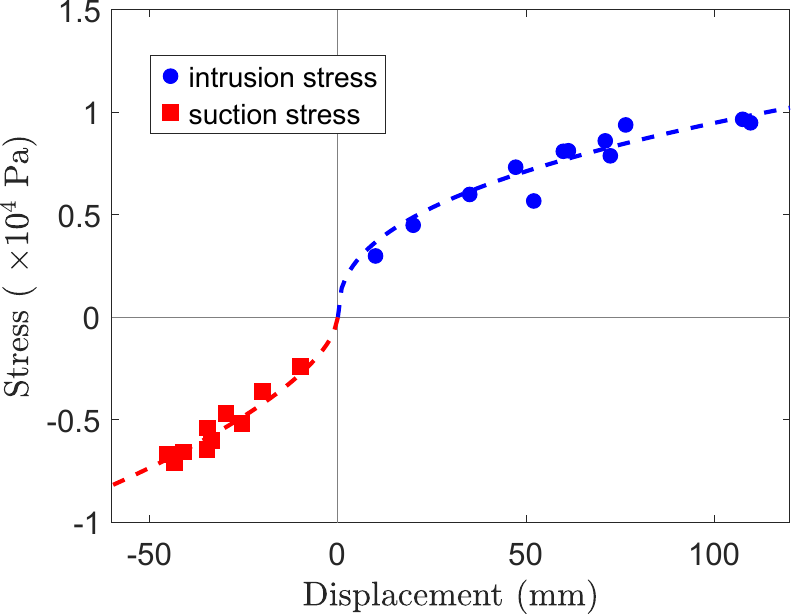}
		\label{fig:HIVI:b}}
	\hspace{-4mm}
	\subfigure[]{
		\includegraphics[width=1.8in]{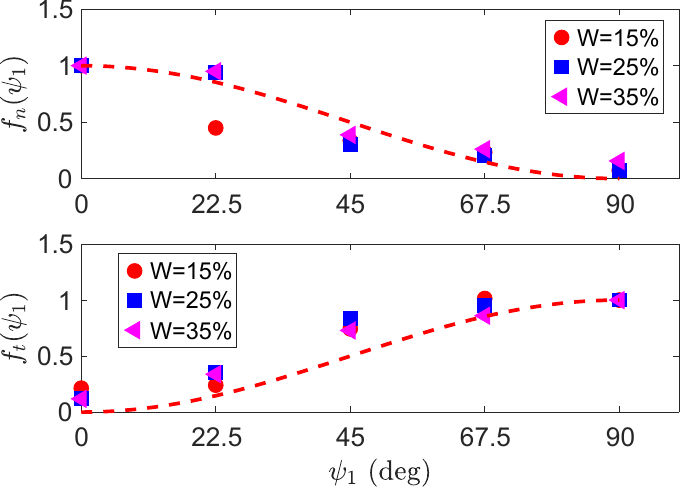}
		\label{fig:fnft:a}}
	\hspace{-4mm}
	\subfigure[]{
		\includegraphics[width=1.8in]{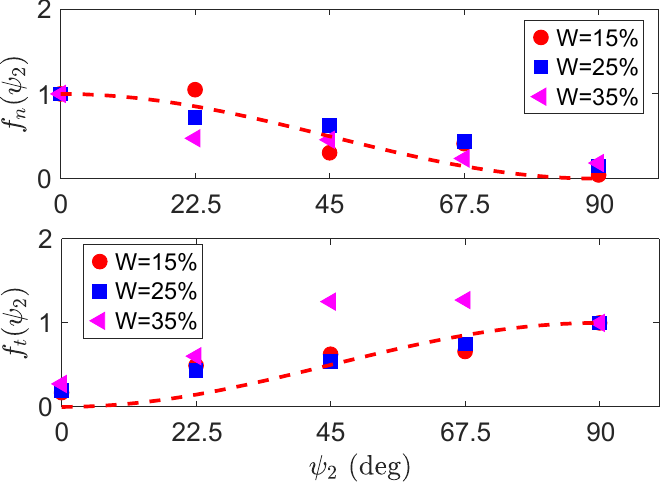}
		\label{fig:fnft:b}}
	\vspace{-3mm}
	\caption{Results of the vertical and horizontal intrusion E3 and E4 using a square plate. (a)-(b) Vertical and horizontal stress versus intrusion displacement. Blue and red dashed curves are fitting curves of $\sigma_{\mathrm{b}}$ using $\sigma_{b}=\alpha \gamma ^n$ for the penetration and suction stress, respectively. (c)-(d) Two dimensionless scaling factors $f_\mathrm{n}$ and $f_\mathrm{t}$ versus penetration angle $\psi_1$ and $\psi_2$ for the vertical and horizontal intrusion, respectively. Red dashed curves are empirical modeling of $f_\mathrm{n}$ and $f_\mathrm{t}$.}
	\label{fig:result:fn_ft}
	\vspace{-0mm}
\end{figure*}

\begin{figure*}[t!]
	\centering
	\subfigure[]{
		\includegraphics[width=2.25in]{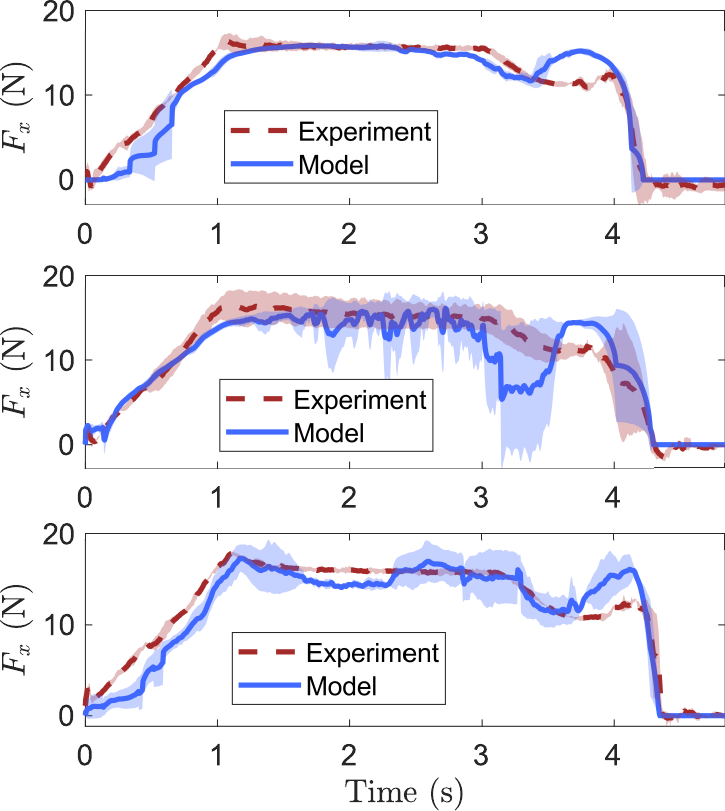}
		\label{fig:model_validation:a}}
	\hspace{-3mm}
	\subfigure[]{
		\includegraphics[width=2.25in]{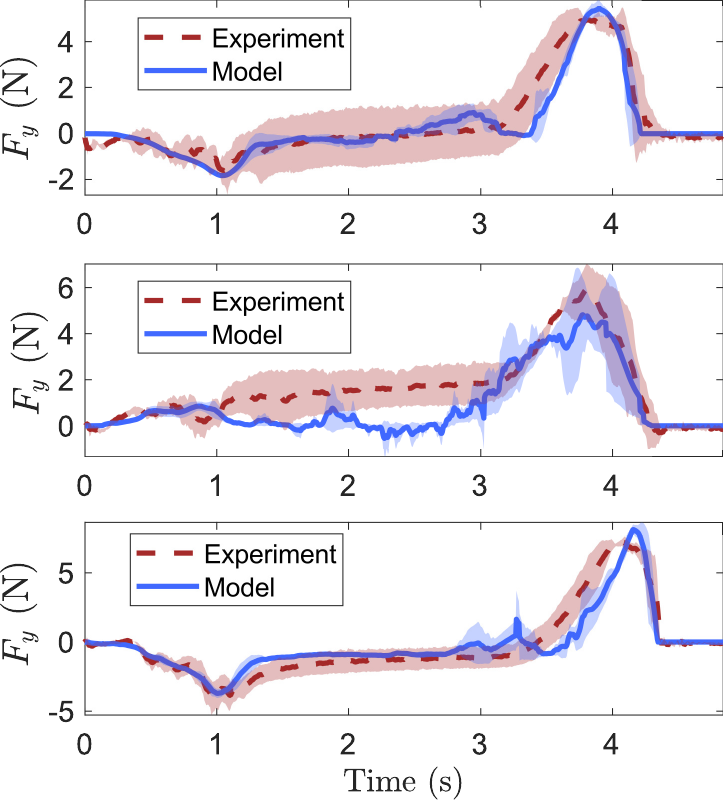}
		\label{fig:model_validation:b}}
	\hspace{-3mm}
	\subfigure[]{
		\includegraphics[width=2.25in]{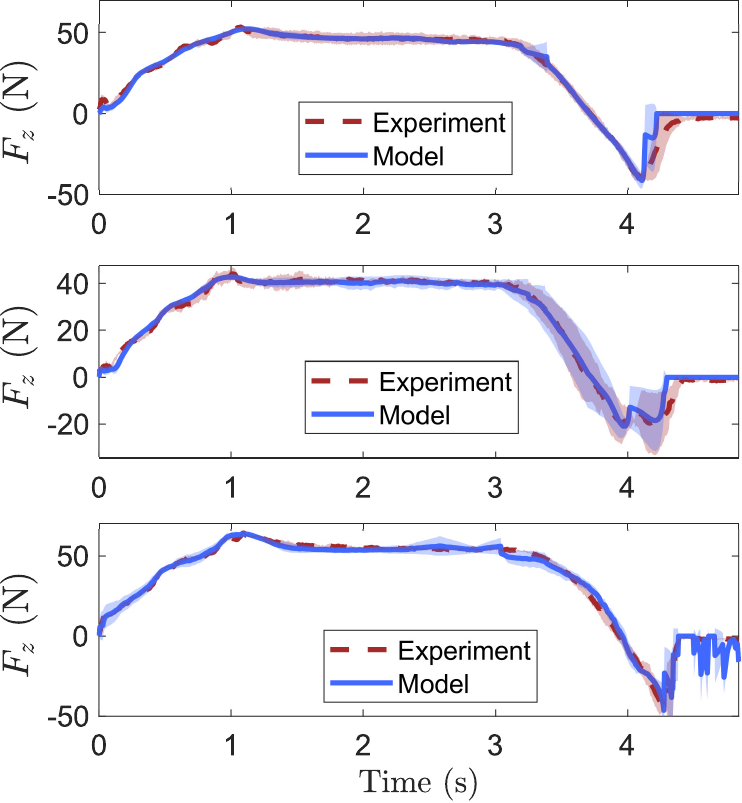}
		\label{fig:model_validation:c}}
	\vspace{-2mm}
	\caption{Comparison of model predictions and experimental results for three representative feet (top row: semi-cylindrical; middle row: semi-spherical; bottom row: flat) at medium intrusion speed ($0.2$~m/s) and water content of $25\%$. (a) $F_x$. (b) $F_y$. (c) $F_z$.}
	\label{fig:model_validation}
\end{figure*}

\renewcommand{\arraystretch}{1.2}
\setlength{\tabcolsep}{0.065in}
\begin{table*}[h!]
	\centering
\caption{Comparison of the relative RMSEs (mean and standard deviation) of three foot shapes at three different velocities.}
\label{tab:RMSE_fig11}
\vspace{-2mm}
\begin{tabular}{c|ccccccccc}
	\toprule[1.2pt]
	\multicolumn{1}{c|}{\multirow{2}{*}{RMSE }} & \multicolumn{3}{c|}{\bf{Flat}} & \multicolumn{3}{c|}{\bf{Semi-cylindrical}} & \multicolumn{3}{c}{\bf{Semi-spherical}} \\ \cline{2-10}
	\multicolumn{1}{c|}{} & $F_z$ ($\%$) & $F_x$ ($\%$) & \multicolumn{1}{c|}{$F_y$ ($\%$)} & $F_z$ ($\%$) & $F_x$ ($\%$) & \multicolumn{1}{c|}{$F_y$ ($\%$)} & $F_z$ ($\%$) & $F_x$ ($\%$) & $F_y$ ($\%$) \\ \midrule
	
	\bf{Slow} & ${9.03\pm 1.10}$ & ${8.72 \pm 1.84}$ & ${12.50 \pm 3.90}$ & ${5.11 \pm 1.09}$ & ${7.88 \pm 4.72}$ & $9.60\pm 2.20$ & ${4.59\pm 0.37}$ & ${12.32 \pm 1.36}$ & ${13.10 \pm 2.20}$ \\
	
	\bf{Medium} & $7.69\pm 2.07$ & $9.80 \pm 1.48$ & $12.80 \pm 3.90$ & $6.81 \pm 1.16$ & $7.64 \pm 1.84$ & ${11.20 \pm 3.10}$ & $5.10\pm 0.87$ & $11.60\pm 6.88$ &  $12.30 \pm 5.40$ \\
	
	\bf{Fast} & $9.21 \pm 1.67$ & $8.88 \pm 2.20$& $18.09 \pm 3.30$ & $4.19 \pm 0.77$ & $6.08 \pm 1.08$ & $12.10 \pm 3.10$ & $9.79 \pm 0.83$ & $15.72 \pm 4.71$ & $10.60 \pm 6.50$ \\
	\bottomrule[1.2pt]
\end{tabular}
\vspace{-2mm}
\end{table*}

Fig.~\ref{fig:exp:b} shows the setup of robotic foot-mud interaction experiments. A bipedal robot platform (model YoboGo-10S from Yobotics Inc., China) was used and the motion control was implemented with an update rate of $500$~Hz on an on-board computer (Jetson TX2 from Nvidia Corp.) A modularized foot embedded with the force/torque sensor (Mini45 from ATI Inc.) was mounted at the ankle. Inverse kinematics of the robot were calculated for the leg joint control given a certain foot intrusion profile. The foot position was obtained through the aforementioned motion capture system at $100$~Hz.

We selected fixed-shape feet to validate the mud resistive force model. Fig.~\ref{fig:exp:b} shows the dimensions of three representative foot shapes: semi-cylindrical ($L\times W\times h_F=80 \times 65 \times 26$~mm, $R=45$~mm), spherical ($R=45$~mm, $h_F=26$~mm), and flat ($L \times W \times h_F= 80 \times 65 \times 26$~mm). A 3D foot intrusion and retraction trajectory was designed based on the normal walking gait of the biped~\cite{zhu2025JBE}. Three different gait velocities were used in experiments: slow ($0.13$~m/s), medium ($0.2$~m/s) and fast ($0.26$~m/s). For each foot-shape condition, we ran three trials. Before each trial we manually blended the mud uniformly and then leveled the surface. Different water-content conditions were considered. The morphing foot was also used in experiments. We further conducted biped walking tests with different feet on muddy terrain. Fig.~\ref{fig:exp:c} shows the experimental setup. The biped was commanded to step forward traversing the mud at a slow speed using the model predictive control and whole-body control (MPC-WBC) framework to ensure the stability of locomotion~\cite{MihalecTMech2023}. Angular velocities and joint torques were recorded to calculate the energy consumption during walking.

\subsection{Results}
\label{sec:Results}

Initial experiments E1 and E2 mainly confirm the average resistive stress of single-plate intrusion in both the vertical and horizontal directions. Detailed descriptions of the results are provided in the Supplementary Materials. We found that the mud resistive stress does not possess significant shape-dependency. This provides a justification for computing resistive forces regardless of the shape by decomposing the surface into infinitesimal plates . Therefore, we selected the square plate for model calibrations.

The E3 and E4 experiments shown in Figs.~\ref{fig:HIVI:a} and~\ref{fig:HIVI:b} reveal the immediate resistive stress $\sigma_{\mathrm{b}}$ under $25$\% water content in the vertical and horizontal directions, respectively. The results confirm that the proposed model matches well with experimental data. Using a square plate with different penetration angles $\psi_i$, we computed two dimensionless scaling factors $f_{\mathrm{n}}(\psi_{i})$ and $f_{\mathrm{t}}(\psi_{i}) $, $i=1,2$. We also considered three representative water contents, $15$, $25$, and $35\%$. Figs.~\ref{fig:fnft:a} and~\ref{fig:fnft:b} show that measurements for all three conditions collapsed into two master curves (red dashed curves) for both $f_{\mathrm{n}}$ and $f_{\mathrm{t}}$, confirming the proposed form of the scaling factors.

For concise presentation, we showcase model predictions of three fix-shaped feet with medium gait velocity using $25\%$ water-content mud. Fig.~\ref{fig:model_validation} shows the comparisons between the model predictions and experimental measurements of force (mean and one-standard deviation). For all three foot shapes, the closed-form models captured the resistive response of mud with good prediction accuracy. Table~\ref{tab:RMSE_fig11} lists the relative root-mean-square errors (RMSEs) of force prediction for different gait velocities with different foot shapes.

\begin{figure*}[t!]
\vspace{-2mm}
  \centering
	\subfigure[]{
	\includegraphics[width=2.32in]{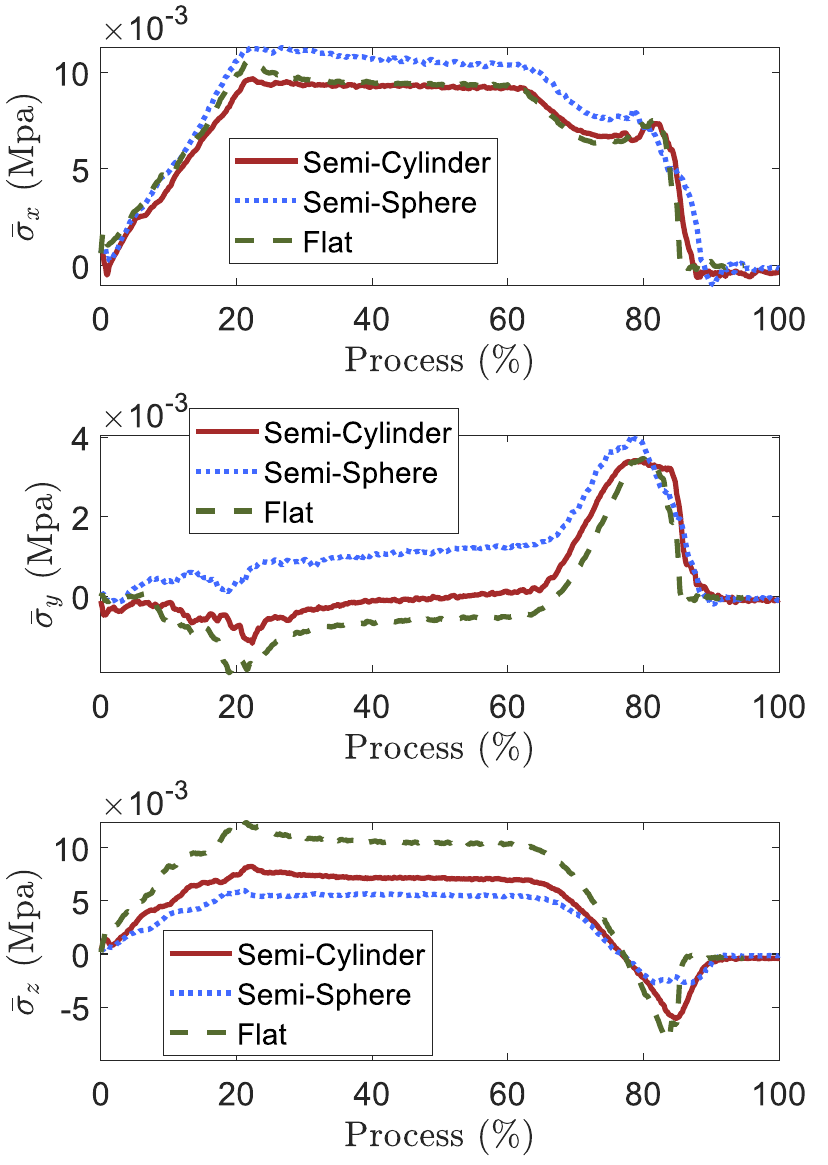}
\label{fig:work_condition_validation:a}}
\hspace{-5mm}
	\subfigure[]{
	\includegraphics[width=2.35in]{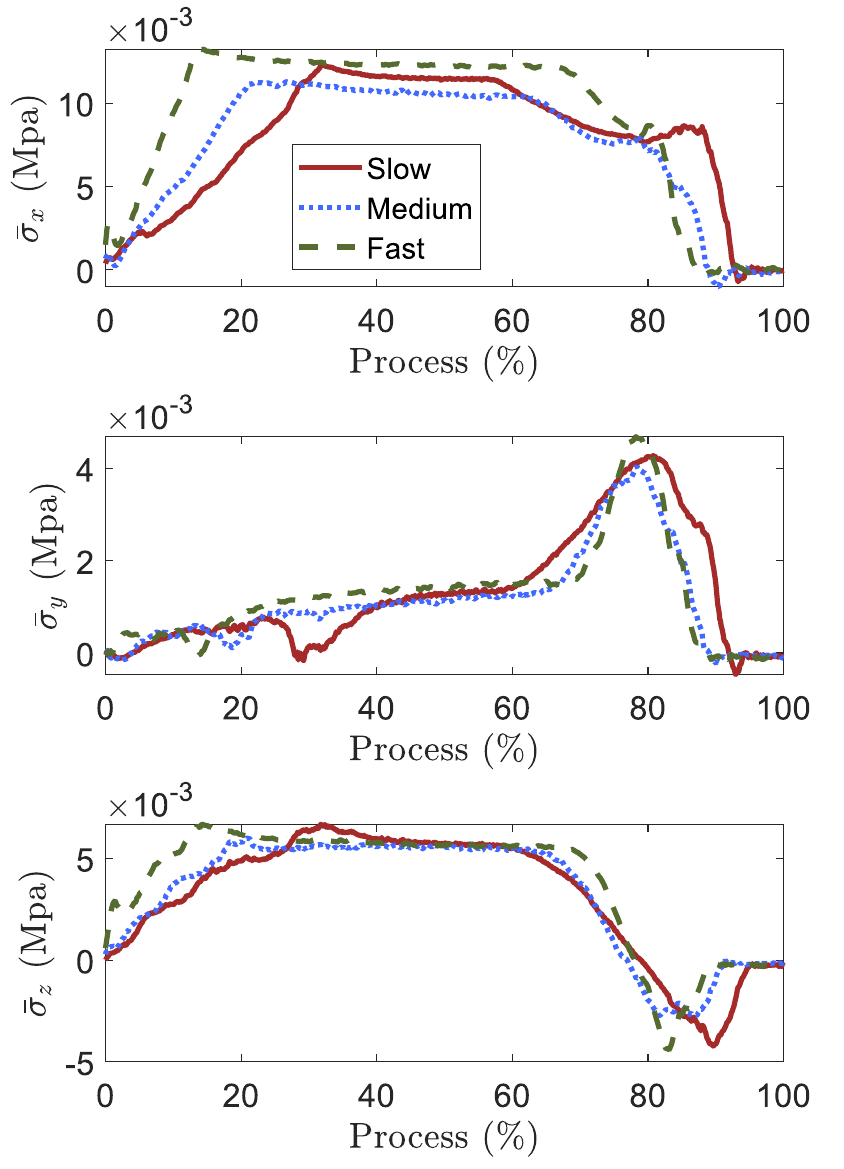}
	\label{fig:work_condition_validation:b}}
\hspace{-5mm}
	\subfigure[]{
	\includegraphics[width=2.25in]{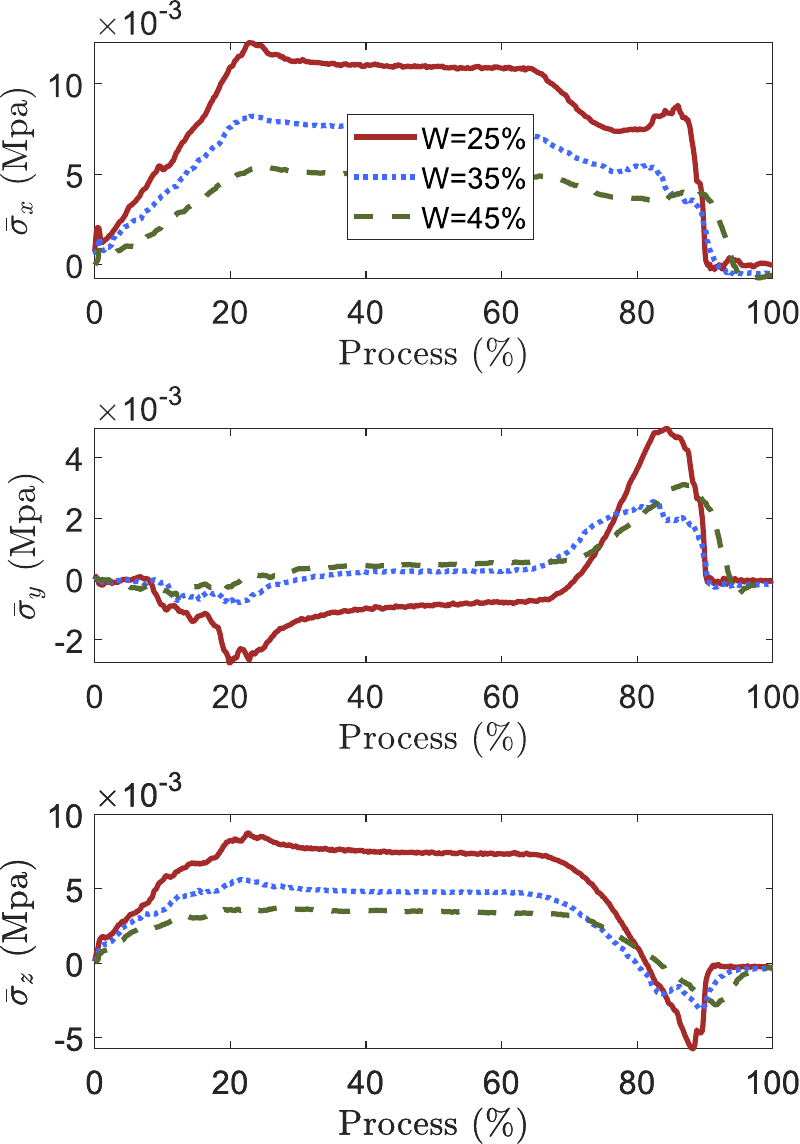}
	\label{fig:work_condition_validation:c}}
\vspace{-2mm}
  \caption{Normalized 3D resistive stresses under various conditions. (a) Different foot shapes. (b) Different intrusion velocities. (c) Different water content.}
 \label{fig:work_condition_validation}
\vspace{-1mm}
\end{figure*}

\setcounter{figure}{10}
\begin{figure*}[t!]
	\hspace{-2mm}
	\subfigure[]{
		\includegraphics[width=2.25in]{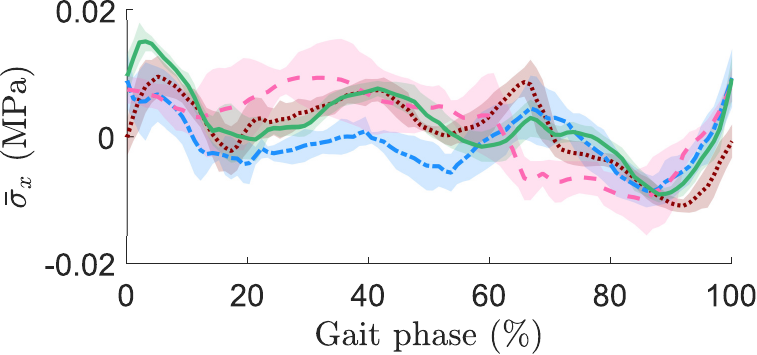}
		\label{fig:dynamicWalking:a}}
	\hspace{-2mm}
	\subfigure[]{
		\includegraphics[width=2.25in]{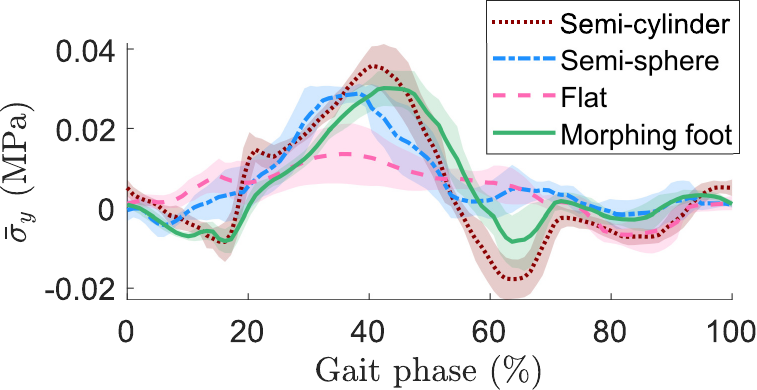}
		\label{fig:dynamicWalking:b}}
	\hspace{-2mm}
	\subfigure[]{
		\includegraphics[width=2.25in]{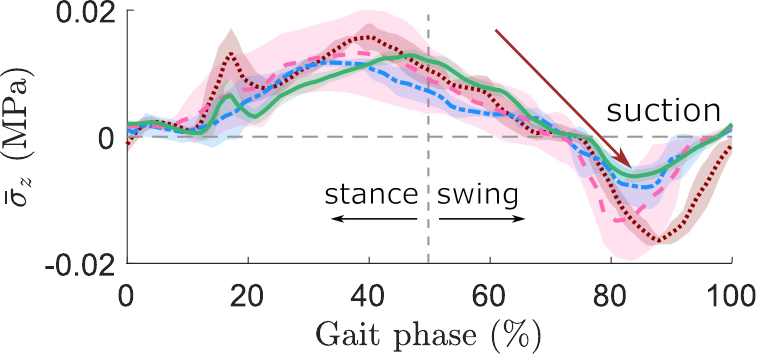}
		\label{fig:dynamicWalking:c}}
	\vspace{-2mm}
	\caption{Average mud resistive stress for four different foot shapes within one walking gait. (a) Stress $\bar{\sigma}_x$. (b) Stress $\bar{\sigma}_y$. (c) Stress $\bar{\sigma}_z$.}
	\label{fig:dynamicWalking}
	\vspace{-3mm}
\end{figure*}

To clearly compare the results among three foot shapes, we normalized the resultant force in each direction by dividing the magnitude of the corresponding cross-section area. Fig.~\ref{fig:work_condition_validation} shows a comparison of the results for different foot shapes, intrusion velocities and mud water content. For the vertical force, the flat foot provided the largest resistive stress followed by the semi-cylindrical foot and semi-spherical foot; see Fig.~\ref{fig:work_condition_validation:a}. The flat foot, however, also suffered from the largest suction force during retraction, while the semi-spherical foot endured the least. The intrusion velocity did not change the resistive force (Fig.~\ref{fig:work_condition_validation:b}). From Fig.~\ref{fig:work_condition_validation:c}, water content also significantly influenced the mud rheology. With increased water content, the mud strength clearly decreased.


Fig.~\ref{fig:dynamicWalking} shows comparisons of the mud reaction forces (stresses) over one entire gait phase of dynamic walking, i.e., half stance phase plus half swing phase, using three fix-shaped feet and the morphing foot. It is clear to see that the morphing foot generated a smaller suction force compared to other three feet; see Fig.~\ref{fig:dynamicWalking:c}. For the three fix-shaped feet, the suction force was almost as large as the supporting force in the intrusion phase, while for the morphing foot, it was only $40 \%$ of intrusion resistance. Fig.~\ref{fig:morphingStiffness} shows a comparison of the vertical resistance stress $\sigma_z$ for the three fix-shaped feet and the morphing foot on $25\%$-water-content mud. It is clearly found that the morphing foot enabled not only the smallest suction stress but also the equivalently largest supporting stress, which resulted in a minimum sinking depth compared with the fixed-shape feet.

\setcounter{figure}{11}
\begin{figure}[h!]
	\centering
		\includegraphics[width=2.8in]{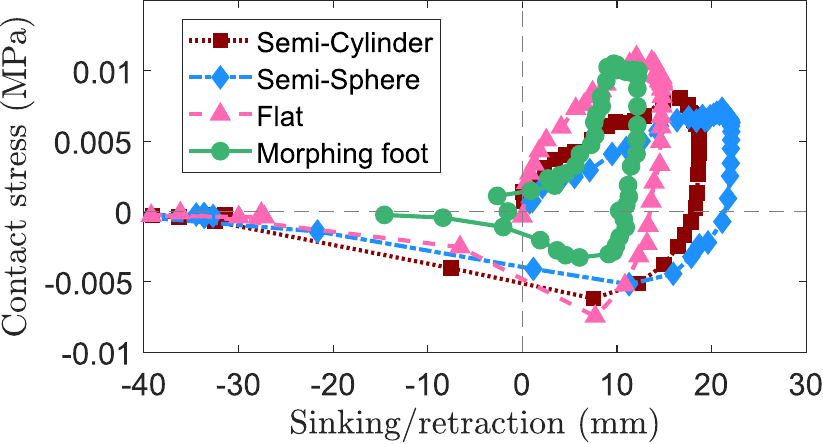}
	\vspace{-2mm}
	\caption{Comparisons of the mud vertical resistance stress of three fix-shaped feet and the morphing foot for mud with $25\%$ water content.}
	\label{fig:morphingStiffness}
	\vspace{-2mm}
\end{figure}

\xunjie{We focused on evaluation of local foot-mud interaction performance in terms of sufficient resistive forces, mitigating suction, and reducing gait energy consumption.} Fig.~\ref{fig:suction_energy_comparison:a} shows a comparison of the maximum suction force and consumed energy for the various feet.
\xunjie{Fig.~\ref{fig:suction_energy_comparison:b} shows a further comprehensive spider-chart comparison of the impulse in three directions, along with the maximum suction and energy cost.} The morphing foot showed the smallest suction force and the lowest energy among the four types of foot. In terms of relative reduction, both the suction force in the vertical direction and the energy consumption of the morphing foot showed a $57$\% and $44$\% decrease, respectively, compared to the semi-cylindrical foot. Compared to the flat foot, the reductions were $32$\% and $26$\%, respectively.
The reduction of the energy consumption using the morphing foot can be interpreted as arising from the reduction of the hysteresis area shown in Fig.~\ref{fig:morphingStiffness}. \xunjie{The balance of the biped is also important and the corresponding balance-related locomotion performance of the morphing foot is discussed in the Supplementary Materials.} These results clearly demonstrate the superior performance of the morphing foot on \xunjie{mud}.
\setcounter{figure}{12}
\begin{figure}[h!]
\hspace{-2mm}
\subfigure[]{
	\includegraphics[width=1.48in]{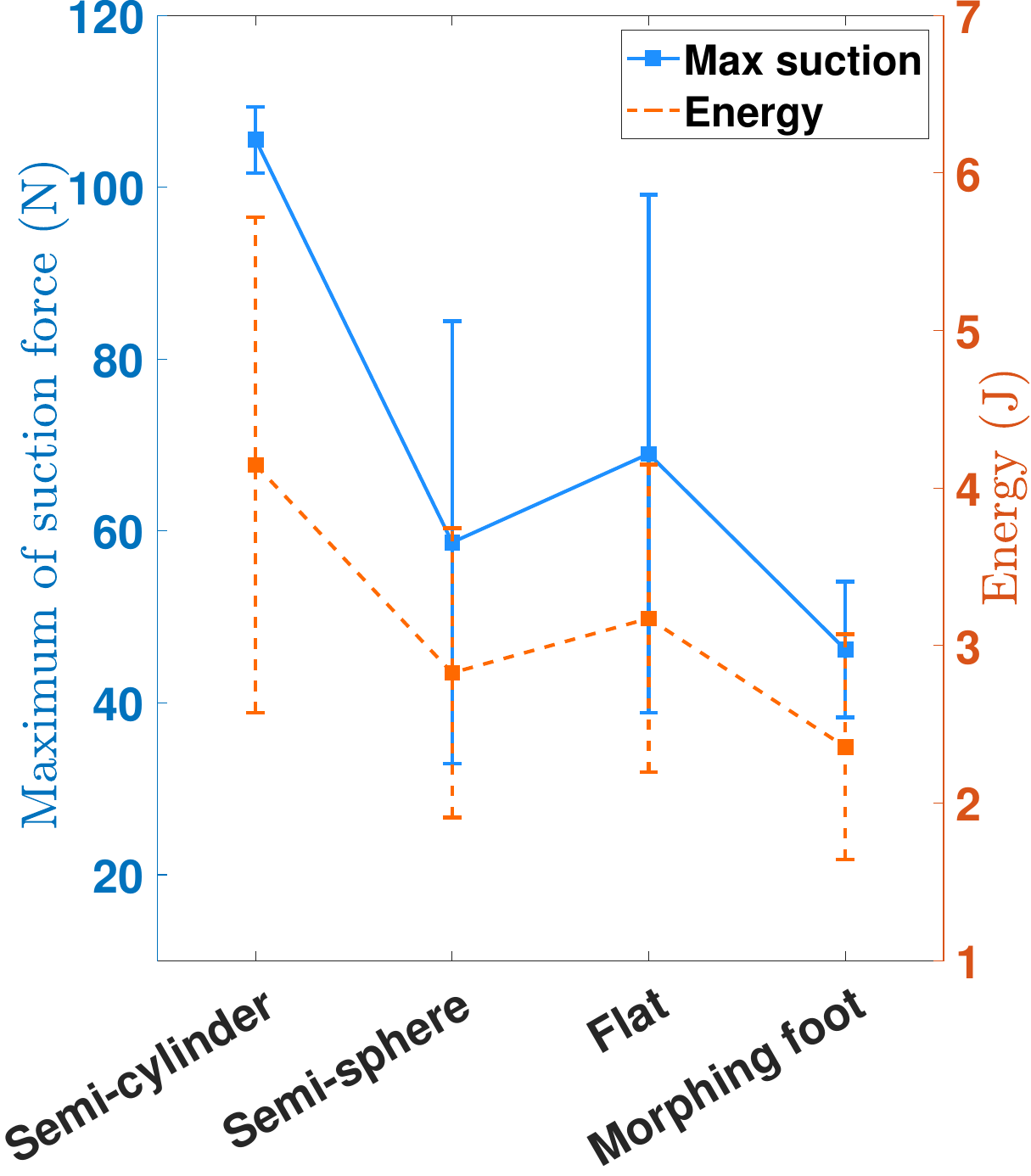}
		\label{fig:suction_energy_comparison:a}}
\hspace{-3mm}
\subfigure[]{
	\includegraphics[width=2in]{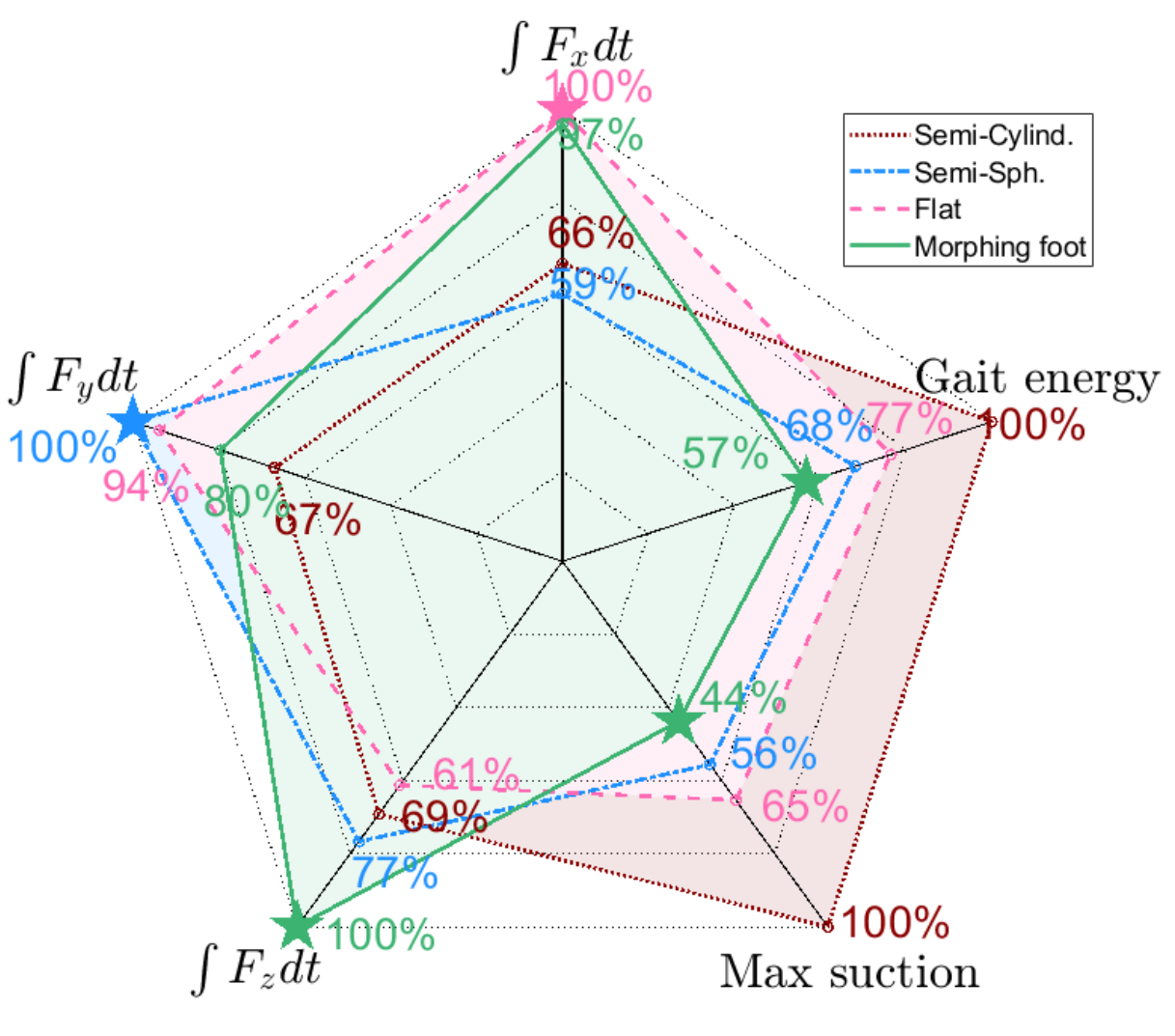}
	\label{fig:suction_energy_comparison:b}}
	\vspace{-3mm}
  \caption{(a) Comparisons of the maximum suction force and energy consumption with various feet. (b) Comprehensive comparison of impulse, maximum suction force and gait energy consumption for different feet.}
  \label{fig:suction_energy_comparison}
	\vspace{-3mm}
\end{figure}

\subsection{Discussion}
Previous work on analytical models for foot-mud interactions has been limited, and this work is the first to present a 3D resistive force model. Compared with the existing models in~\cite{GodonRAL2022,ChenAIM24,LiuRSS2025} used for bipedal robot locomotion, the proposed model not only significantly extends its applicability from 1D vertical intrusion to 3D motion, but also allows physical interpretation of mud rheology, along with accurate force predictions. Instead of simply using the concept of stiffness of mud as in~\cite{GodonRAL2022,LiuRSS2025}, this work fills a knowledge gap by accounting for suction in the foot-mud interactions, which has not been reported in other work. To the best knowledge of the authors, the proposed model is the first work to provide accurate closed-form 3D force formulations based on mud rheology. \xunjie{It should be noted that the proposed model was developed and validated for the intermediate water-content regime ($15\sim35\%$) instead of highly fluidized mud. In future work, it would be desirable to extend the model to higher-water-content cases where the mud may behave like a high-viscosity liquid, i.e., gel, and to identify the water volume fraction at which the rheology of mud changes significantly.}

Compared with walking on other yielding terrain, such as sand, a significant feature of \xunjie{mud} is the suction upon retraction. As shown in Figs.~\ref{fig:work_condition_validation:a} and~\ref{fig:morphingStiffness}, the magnitudes of suction forces in the vertical and lateral forces were as large as those of immediate resistance forces during the intrusion.
\xunjie{However, conventional RFT for granular media only accounts for friction-dominated resistance proportional to the hydrostatic-like pressure, and neglects the suction and time-dependent stress relaxation which are important for mud~\cite{li2013terradynamics,ChenTMECH2025Sand}.} \xunjie{Therefore}, previous foot-sand resistive-force models cannot be applied to muddy terrain. We present \xunjie{additional comparison results in the Supplementary Materials}.


In spite of these contributions, there are some limitations of this work.
The model \xunjie{focuses on the foot-mud interaction in the stance phase, and therefore,} does not capture the effect of added mud mass on the foot after pulling out from the media. \xunjie{This inertial effect is likely to influence the swing dynamics and could even potentially affect the effective foot shape at the next stance. Additional rigorous modeling for higher water-content scenarios is also among future work.}

\vspace{-2mm}
\section{Conclusions}
\label{sec:conclusion}

We have presented a new 3D resistive force model for legged robot locomotion on mud. The model accounts for several main sources of mud resistance, namely visco-elasticity, thixotropy, and retractive suction. One attractive feature of the model is that it facilitates underlying physical interpretation in addition to providing accurate resistive force predictions. Closed forms of 3D resultant forces were further developed for flat-, semi-cylindrical- and semi-spherical-shape feet. Using the new foot-mud interaction model, a passively morphing foot was designed such that it reduced foot sinking and saved energy on muddy terrain. Extensive experiments were conducted to demonstrate the accuracy and efficacy of the mud model and improved locomotion mobility and energy-saving by the morphing foot. \xunjie{Two ongoing research directions are the enhancement of the model implementation for more complex outdoor mud terrain and the optimal design of the morphing foot with online terrain tactile sensing capability using the testbed developed in~\cite{GongTMech2024}.}

\bibliography{ChenAIM_TMECH_Ref}
\end{document}


\bibliographystyle{IEEEtran}

\title{\vspace{-15mm} Supplementary Materials for\\
\vspace{4mm}
\textbf{A Foot Resistive Force Model for Legged Locomotion \\on Muddy Terrains}}
\date{}
\author{Xunjie Chen\thanks{X. Chen, X. Huang,  J. Shan, and J. Yi are with the Department of Mechanical and Aerospace Engineering, Rutgers University, Piscataway, NJ 08854 USA (email: xc337@rutgers.edu; xh301@scarletmail.rutgers.edu; jgyi@rutgers.edu; jshan@soe.rutgers.edu). ({\em Corresponding author: Jingang~Yi})}, Liuyin Wang\thanks{L. Wang and Y. Shen are with the Department of Electrical and Biomedical Engineering, University of Nevada, Reno, NV 89557 USA (liuyinw@unr.edu; ytshen@unr.edu).}, Xinyan Huang, Jerry Shan, Yantao Shen, and Jingang Yi}
\maketitle


This supplementary document contains three sections. Section~\ref{Sec:ODEs} presents the derivations of the closed-form formulation for mud thixotropic stress $\sigma_{\mathrm{th}}$ and suction stress $\sigma_{\mathrm{s}}$. Section~\ref{Sec:force} discusses the derivation of the resultant force calculations for three representative robotic feet, namely, semi-cylindrical, flat, and spherical shapes. Finally, Section~\ref{Sec:Exp_Results} presents additional experimental results \xunjie{in terms of calibrations, parameter sensitivity, and dynamic walking balance,} and model comparisons on different terrains, including sand and rigid ground surfaces.

\section{Mud Thixotropic and Suction Stresses Calculation}
\label{Sec:ODEs}

\subsection{Mud thixotropy stress}

Given the differential equations governing $\sigma_{\mathrm{th}}$ and $\xi$ by~(3) in the manuscript, we provide their steady-state solutions with zero initial conditions. At steady-state, $\ddot{\gamma}=0$, and~(3) in the manuscript becomes
\begin{subequations}
\begin{align}
& \dot{\sigma}_{\mathrm{th}}= -\frac{1}{\lambda}\sigma_{\mathrm{th}}+\frac{1}{\lambda}[(\eta_{\infty} + \xi\eta_{\mathrm{m}}) \dot{\gamma}], \label{eq0a} \\
& \dot{\xi} =-\frac{1}{\tau_{\xi}}\xi+k_a,	\label{eq0b}
\end{align}
\end{subequations}
where $\tau_{\xi} = 1/(k_a + k_r|\dot{\gamma}|)$ is the time constant for the structural parameter dynamics~\eqref{eq0b}. The solution of~\eqref{eq0b} is obtained as $\xi(t) = \left(1-e^{t/\tau_{\xi}}\right)\xi_{\mathrm{ss}}$, where the steady state of $\xi$ is defined as $\xi_{\mathrm{ss}} = k_a/(k_a+k_r|\dot{\gamma}|)=\frac{\dot{\gamma}_c}{\dot{\gamma}_c+|\dot{\gamma}|}$, $\dot{\gamma}_c = \frac{k_a}{k_r}$. We plug the above solution of $\xi(t)$ into~\eqref{eq0a} and obtain
\begin{align}
		\sigma_{\mathrm{th}}(t) & = \frac{1}{\lambda}e^{-\frac{t}{\lambda}}\int_{0}^{t}e^{\frac{s}{\lambda}}\left[\eta_{\infty}\dot{\gamma}+(1-e^{-\frac{s}{\tau_{\xi}}})\xi_{\mathrm{ss}}\eta_{\mathrm{m}}\dot{\gamma}\right]ds \nonumber \\
		&=\left(1-e^{-\frac{t}{\lambda}}\right)\left(\eta_{\infty}+\xi_{\mathrm{ss}}\eta_{\mathrm{m}}\right)\dot{\gamma}+\left(e^{-\frac{t}{\lambda}}-e^{-\frac{t}{\tau_{\xi}}}\right)\frac{\tau_{\xi}}{\tau_{\xi}-\lambda}\xi_{\mathrm{ss}}\eta_{\mathrm{m}}\dot{\gamma} \label{app:eqn:sigma_th}
\end{align}

Time constants $\tau_{\xi}$ and $\lambda$ are assumed to have the same scale, that is, $\tau_{\xi} \approx \lambda$. This is justified by the experimental results. Therefore, the second term in~\eqref{app:eqn:sigma_th} vanishes and we obtain the approximation of the thixotropic stress
\begin{equation}
\label{app:eqn:sigma_th_approx}
	\sigma_{\mathrm{th}}(t) =\left(1-e^{-\frac{t}{\lambda}}\right)\left(\eta_{\infty}+\xi_{\mathrm{ss}}\eta_{\mathrm{m}}\right)\dot{\gamma}.
\end{equation}
Notice that $\sigma_{\mathrm{th,ss}}(\dot{\gamma}) = \left(\eta_{\infty}+\xi_{\mathrm{ss}}\eta_{\mathrm{m}}\right)\dot{\gamma}$, we re-write~\eqref{app:eqn:sigma_th_approx} as $\sigma_{\mathrm{th}}(t) = \left(1-e^{-\frac{t}{\lambda}}\right)\sigma_{\mathrm{th,ss}}(\dot{\gamma})$ as~(9) in the manuscript.

\subsection{Suction stress}

The differential equation~(5) in the manuscript is rewritten as
\begin{equation}
	\dot{\sigma}_{\mathrm{s}}=-a(\gamma)\sigma_s(t) + b(\gamma),
\label{eq1}
\end{equation}
where $a = \frac{\phi(\gamma)}{\tau_{\mathrm{build}}}+\frac{1-\phi(\gamma)}{\tau_{\mathrm{leak}}}$ and $b=\frac{\phi(\gamma)}{\tau_{\mathrm{build}}}p_0$. We consider that the quasi-static condition of the motion input $\gamma$ is time-invariant at any instantaneous time. Therefore, the solution of~\eqref{eq1} is obtained as
\begin{equation*}
	\sigma_{\mathrm{s}}(t) = \frac{b}{a} + \left[\sigma_{\mathrm{s}}(t_0)-\frac{b}{a}\right]e^{-a(t-t_0)},
\end{equation*}
where the initial time is denoted as $t_0 = t_{\mathrm{w}}$, that is, the time moment when the intruder starts to retract from the mud media. Considering that $\sigma_{\mathrm{s}}(t_{\mathrm{w}})= 0$, we obtain the solution by substituting $a$, $b$ with $p_0=-\sigma_Y$ into the above equation
\begin{equation}
	\sigma_\mathrm{s}(t)=(1-e^{-a\Delta t})\sigma_{\mathrm{s,ss}},~\Delta t = t-t_{\mathrm{w}},
\end{equation}
where $\sigma_{\mathrm{s,ss}}=-\frac{\phi(\gamma)}{a\tau_{\mathrm{build}}}\sigma_Y$ is the steady-state suction stress.

\section{Closed-Form Calculation of Resistive Forces}
\label{Sec:force}

With the above stress calculation, we now present the closed-form formulations for the resultant resistive force on flat, semi-cylindrical, and semi-spherical feet.

\subsection{Flat foot}

For a rectangular flat foot with length $L$ and width $W$, using~(6) in the manuscript, the elementary forces in three directions are given by
\begin{displaymath}
	dF_x = f_\mathrm{n}(\varphi)\sigma_x Ld\tilde{h} + f_\mathrm{t}\left(\varphi_c\right)\sigma_y Wd\tilde{h}, \;
	dF_y = f_\mathrm{t}(\varphi)\sigma_x Ld\tilde{h} + f_\mathrm{n}\left(\varphi_c\right)\sigma_y Wd\tilde{h}, \;
	dF_z = LW\sigma_z.
\end{displaymath}
It is straightforward to obtain that the foot effective area $S_e=LW$ and integrating cross the entire plate yields the resultant forces as
\begin{equation}
F_x = z\left[Lf_{\mathrm{n}}(\varphi)\sigma_x +Wf_{\mathrm{t}}(\varphi_c)\sigma_y\right], \;
F_y =z\left[Lf_{\mathrm{t}}(\varphi)\sigma_x+Wf_{\mathrm{n}}(\varphi_c)\sigma_y\right], \; F_z = S_e\sigma_z. \label{eqn:Flat_Fz}
\end{equation}

\subsection{Semi-cylindrical foot}

\begin{figure}[h!]
	\centering
	\includegraphics[width=6in]{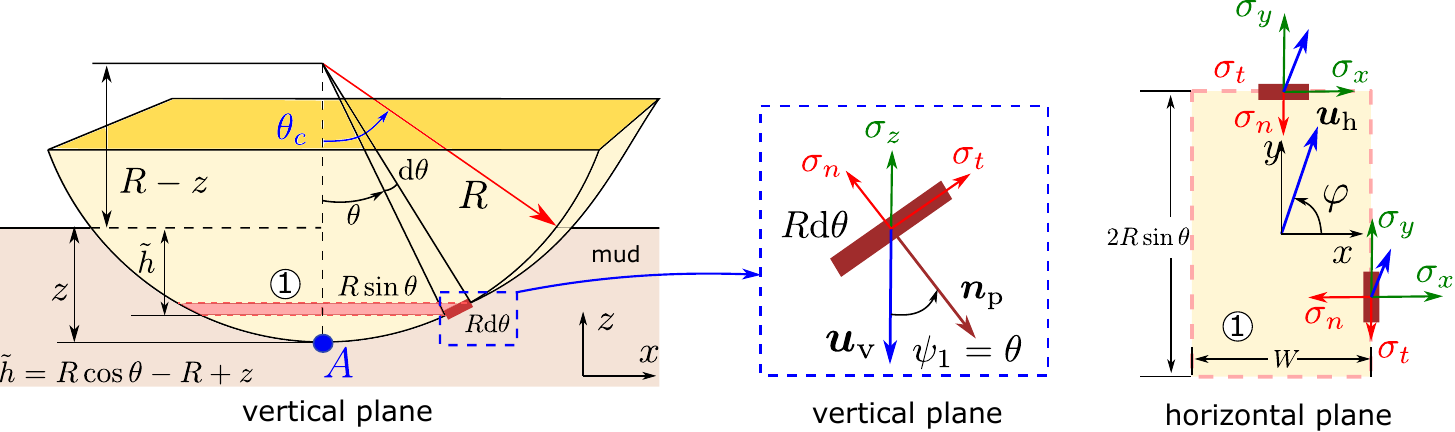}
	\caption{The schematics of a semi-cylindrical foot intruding in mud.}\label{fig:semiCylind}
\end{figure}

Fig.~\ref{fig:semiCylind} shows the schematic of a semi-cylindrical foot intruding into mud media. The horizontal sliding and vertical sinking positions of the foot bottom tip point (point $A$ in the figure) are denoted as $(x, y)$ and $z$, respectively. The horizontal projection and vertical component of the foot velocity are denoted by $\bs{u}_{\mathrm{h}} = [u_x\;u_y]^{\top}$ and $\bs{u}_{\mathrm{z}}$, respectively. We denote the angle between the foot motion and the longitudinal direction as $\varphi$. By introducing variable $\theta$ representing any infinitesimal plate location with respect to the vertical direction, the plate's depth is then $\tilde{h} = z + R\cos{\theta}-R$, where $R$ is the radius of the cross section of the foot.

We compute the longitudinal and lateral forces on the highlighted horizontal section with infinitesimal angle $d \theta$ (length $2R\sin{\theta}$ and width $W$) at the depth of $\tilde{h}$ in mud media as
\begin{equation}
dF_x =f_\mathrm{n}(\varphi)\sigma_x 2R\s_{\theta}d\tilde{h} + f_\mathrm{t}\left(\varphi_c\right)\sigma_y Wd\tilde{h},\; dF_y =f_\mathrm{t}(\varphi)\sigma_x 2R\s_{\theta}d\tilde{h} + f_\mathrm{n}\left(\varphi_c\right)\sigma_y Wd\tilde{h},
\label{eq10}
\end{equation}
where $\varphi_c=\frac{\pi}{2}-\varphi$, $\sigma_{x}$ ($\sigma_{y}$) is the total longitudinal (lateral) stress that is induced by the decomposed motion and is calculated by~(1) in the manuscript. In~\eqref{eq10}, notation convention $\s_\theta:=\sin \theta$ is used for $\theta$ and other angles throughout this document. We also have $d\tilde{h}= -R\s_{\theta}d\theta$, and from the above equation, obtain the total resultant forces as
\begin{equation*}
F_x =\int_{0}^{\theta_c} \left[2f_\mathrm{n}(\varphi)\sigma_x R^2\s^2_{\theta} + f_\mathrm{t}\left(\varphi_c\right)\sigma_y WR\s_{\theta}\right]d\theta,\;	F_y =\int_{0}^{\theta_c} \left[2f_\mathrm{t}(\varphi)\sigma_x R^2\s^2_{\theta} + f_\mathrm{n}\left(\varphi_c\right)\sigma_y WR\s_{\theta}\right]d\theta,
\end{equation*}
where $\theta_c = \cos^{-1}(1-z/R)$. By integration of the above equations, we finally obtain the closed-form of the resultant forces as
\begin{subequations}
\begin{align}
& F_x =f_\mathrm{n}(\varphi)\sigma_x R^2\left(\theta_c-\frac{\s_{2\theta_c}}{2}\right)+f_{\mathrm{t}}\left(\varphi_c\right)\sigma_y WR(1-\c_{\theta_c})=\frac{S_e}{2}\Bigl[\frac{R}{W}\left(\frac{\theta_c}{\s_{\theta_c}}-\c_{\theta_c}\right)f_\mathrm{n}(\varphi)\sigma_x+\frac{1-\c_{\theta_c}}{\s_{\theta_c}}f_{\mathrm{t}}(\varphi_c)\sigma_y\Bigr],\\
& F_y =f_\mathrm{t}(\varphi)\sigma_x R^2\left(\theta_c-\frac{\s_{2\theta_c}}{2}\right)+f_{\mathrm{n}}\left(\varphi_c\right)\sigma_y WR(1-\c_{\theta_c})=\frac{S_e}{2}\Bigl[\frac{R}{W}\left(\frac{\theta_c}{\s_{\theta_c}}-\c_{\theta_c}\right)f_\mathrm{t}(\varphi)\sigma_x+\frac{1-\c_{\theta_c}}{\s_{\theta_c}}f_{\mathrm{n}}(\varphi_c)\sigma_y\Bigr],
\end{align}
\end{subequations}
where $S_e=2RW \s_{\theta_c}$ is foot effective area in the $z$-axis direction and notation convention $\c_{\theta_c}:=\cos \theta_c$ is used for $\theta_c$ and other angles throughout this document.

For the vertical force, we consider the orientation of each infinitesimal plate, $\theta$, and we have
\begin{equation*}
	dF_z = \left[f_\mathrm{n}(\theta)\sigma_z\c_{\theta}+f_\mathrm{t}(\theta)\sigma_z\s_{\theta}\right]WRd\theta,
\end{equation*}
where $\sigma_{z}$ is the total stress induced by the vertical motion. We consider an empirical model to represent $f_{\mathrm{n}}(\theta) = \frac{1}{2}\left(1+\c_{2\theta}\right)$ and $f_{\mathrm{t}}(\theta) = \frac{1}{2}\left(1-\c_{2\theta}\right)$, which is discussed in the Section IV-B in the manuscript. We obtain the vertical force by integration of $dF_z$ as
\begin{align*}
F_z = \frac{WR\sigma_z}{2}\int_{-\theta_c}^{\theta_c}\bigg[\left(1+\c_{2\theta}\right)\cos{\theta}+\left(1-\c_{2\theta}\right)\s_{\theta}\bigg]d\theta =WR\sigma_z\left(\frac{4}{3}\s_{\theta_c}+\frac{2}{3}\s_{\theta_c}\c^2_{\theta_c}\right)= \frac{1}{3}S_e \left(2+\c^2_{\theta_c}\right)\sigma_z.
\end{align*}

\subsection{Semi-spherical foot}

\begin{figure}[h!]
	\centering
	\includegraphics[width=6in]{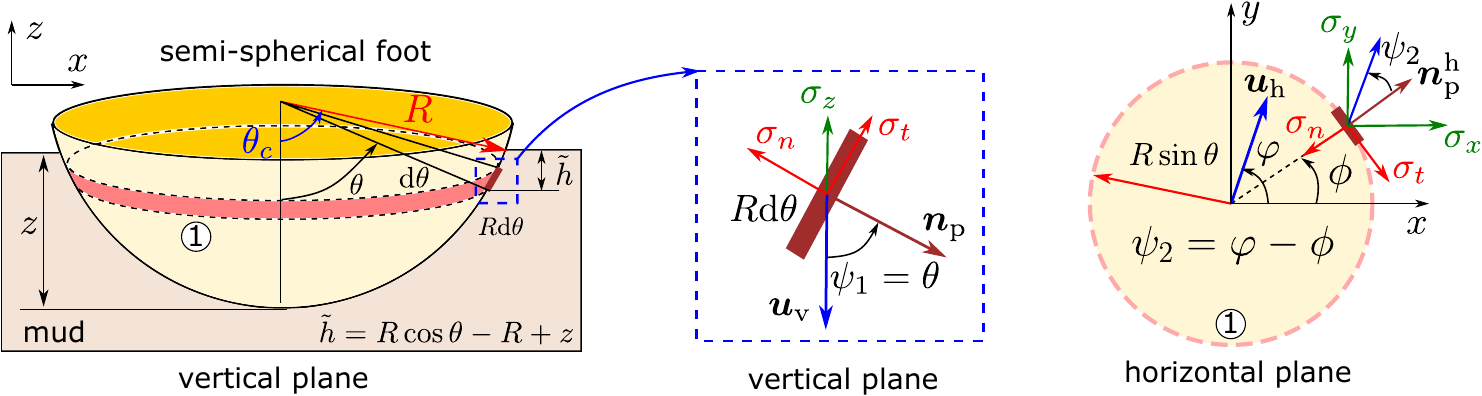}
	\caption{The schematics of a semi-spherical foot intruding in mud.}
	\label{fig:semiSphere}
\end{figure}

Fig.~\ref{fig:semiSphere} shows a schematic of a semi-spherical foot intruding in the mud media. Similar to the case of the semi-cylindrical foot, the elementary vertical force applied on an annulus at the depth of $\tilde{h}$ (with radius $Rs_{\theta}$) is computed as
\begin{equation*}
dF_z = \left[f_\mathrm{n}(\theta)\c_{\theta}+f_\mathrm{t}(\theta)\s_{\theta}\right]\sigma_z (Rd\theta) (2\pi R\s_{\theta}).
\end{equation*}
In the cross section, we use $f_{\mathrm{n}}(\theta) = \frac{1}{2}\left(1+\c_{2\theta}\right)$ and $f_{\mathrm{t}}(\theta) = \frac{1}{2}\left(1-\c_{2\theta}\right)$ and the resultant vertical force is
\begin{align}
F_z  &= 2\pi R^2\sigma_z\int_{0}^{\theta_c} \big[f_\mathrm{n}(\theta)\c_{\theta}+f_\mathrm{t}(\theta)\s_{\theta}\big]\s_{\theta}d\theta = \pi R^2\sigma_z\int_{0}^{\theta_c}\big[\left(1+\c_{2\theta}\right)\c_{\theta}\s_{\theta}+\left(1-\c_{2\theta}\right)\s^2_{\theta}\big]d\theta\nonumber\\
	&= \pi R^2\sigma_z \left(\frac{6\theta_c-5\s_{2\theta_c}}{8}+\frac{\c^3_{\theta_c}\s_{\theta_c}}{2}+\frac{1-\c^4_{\theta_c}}{2}\right)=\frac{S_e}{4}\left(2+\c^2_{\theta_c}+\frac{2\c^3_{\theta_c}}{\s_{\theta_c}}-5\frac{\c_{\theta_c}}{\s_{\theta_c}}+\frac{3\theta_c}{\s^2_{\theta_c}}\right)\sigma_z,
\end{align}
where $\theta_c = \cos^{-1}(1-z/R)$ and $S_e=\pi R^2 \s^2_{\theta_c}$ is foot effective area in the vertical direction.

For the annulus at the depth of $\tilde{h}$, we denote the location angle of an arbitrary plate on the annulus as $\varphi$, as shown in Fig.~\ref{fig:semiSphere}. Then, the longitudinal and lateral forces are calculated as
\begin{displaymath}
dF_x =-f_\mathrm{n}(\psi_2)\sigma_x \c_{\phi}+ f_\mathrm{t}(\psi_2)\sigma_x \s_{\phi}R_s d\phi d\tilde{h}, \;
dF_y =f_\mathrm{n}(\psi_2)\sigma_y \s_{\phi}+ f_\mathrm{t}(\psi_2)\sigma_y \c_{\phi}R_s d\phi d\tilde{h},
\end{displaymath}
where $\psi_2=\varphi-\phi$, $R_s = R\s_{\theta}$ and $d\tilde{h}= -R\s_{\theta}d\theta$. By integration of the above equations, we have
\begin{align}
F_x = & -\sigma_x R^2 \int_{0}^{\theta_c}\s^2_{\theta}d\theta\int_{0}^{2\pi}\big[f_\mathrm{n}(\psi_2) \c_{\phi}- f_\mathrm{t}(\psi_2)\s_{\phi}\big]d\phi, \label{app:eqn:Fx1}\\
	F_y = & -\sigma_y R^2 \int_{0}^{\theta_c}\s^2_{\theta}d\theta\int_{0}^{2\pi}\big[f_\mathrm{n}(\psi_2) \s_{\phi}+ f_\mathrm{t}(\psi_2)\c_{\phi}\big]d\phi. \label{app:eqn:Fy1}
\end{align}
To obtain the second integration regarding $\phi$ in the above equations, we need to extend $f_\mathrm{n}(\psi_2)$ and $f_\mathrm{t}(\psi_2)$ from $\psi_2\in[0, \frac{\pi}{2}]$ to $\psi_2\in[-\pi, \pi]$. Here, we provide the formulation of $f_\mathrm{n}(\psi_2)$ and $f_\mathrm{t}(\psi_2)$ as
\begin{equation*}
	f_\mathrm{n}(\psi_2) = \begin{cases}
		\frac{1}{2}(1+\c_{2\psi_2}), & |\psi_2|\leq\frac{\pi}{2}, \\
		-\frac{1}{2}(1+\c_{2\psi_2}), & |\psi_2|\in(\frac{\pi}{2},\pi],
	\end{cases}, \;\; \text{and} \;\;
	f_\mathrm{t}(\psi_2) = \begin{cases}
		\frac{1}{2}(1-\c_{2\psi_2}), & \psi_2\in[0,\pi], \\
		-\frac{1}{2}(1-\c_{2\psi_2}), & \psi_2\in[-\pi,0).
	\end{cases}
\end{equation*}
Then, we change the integration variable $\psi_2=\varphi-\phi$ of the second integration part in~\eqref{app:eqn:Fx1} and~\eqref{app:eqn:Fy1}, and obtain
\begin{align*}
&\int_{0}^{2\pi}\big[f_\mathrm{n}(\varphi-\phi) \c_{\phi}- f_\mathrm{t}(\varphi-\phi)\s_{\phi}\big]d\phi	=\int_{\varphi-\pi}^{\varphi+\pi}\big[f_\mathrm{n}(\phi) \c_{\varphi-\phi}-f_\mathrm{t}(\phi)\s_{\varphi-\phi}\big]d\phi, \\
&\int_{0}^{2\pi}\big[f_\mathrm{n}(\varphi-\phi) \s_{\phi}+ f_\mathrm{t}(\varphi-\phi)\c_{\phi}\big]d\phi=\int_{\varphi-\pi}^{\varphi+\pi}\big[f_\mathrm{n}(\phi) \s_{\varphi-\phi}+f_\mathrm{t}(\phi)\c_{\varphi-\phi}\big]d\phi.
\end{align*}
Without losing the generality, we consider that $\varphi \in(0,\pi/2)$, and therefore, simplify the integration above as
\begin{align*}
&\int_{\varphi-\pi}^{\varphi+\pi}\big[f_\mathrm{n}(\phi) \c_{\varphi-\phi}-f_\mathrm{t}(\phi)\s_{\varphi-\phi}\big]d\phi=\frac{32}{3}\c_{\varphi},\; \int_{\varphi-\pi}^{\varphi+\pi}\big[f_\mathrm{n}(\phi) \s_{\varphi-\phi}+f_\mathrm{t}(\phi)\c_{\varphi-\phi}\big]d\phi= \frac{32}{3}\s_{\varphi}.
\end{align*}
Finally, the resultant longitudinal and lateral forces are obtained as
\begin{align*}
& F_x  = \frac{1}{3}R^2\left(8\theta_c-4\s_{\theta_c}\right)\c_{\varphi}\sigma_x=\frac{4S_e}{3\pi\s_{\theta_c}}\Bigl(\frac{2\theta_c}{\s_{\theta_c}}-1\Bigr)\c_{\varphi}\sigma_x, \\
& F_y  = \frac{1}{3}R^2\left(8\theta_c-4\s_{\theta_c}\right)\s_{\varphi}\sigma_y=\frac{4S_e}{3\pi\s_{\theta_c}}\Bigl(\frac{2\theta_c}{\s_{\theta_c}}-1\Bigr)\s_{\varphi}\sigma_y.
\end{align*}

\section{Experiments and Results}
\label{Sec:Exp_Results}

In this section, we provide details of the experiments and additional results, including vertical and horizontal intrusions using a flat plate of different shapes, gait energy \xunjie{and balance} analysis using the morphing foot, and model comparisons on different types of terrains.

\subsection{Plate intrusion experiments and model calibration}

\begin{figure}[h!]
\hspace{-2mm}
\subfigure[]{
		\includegraphics[width=1.67in]{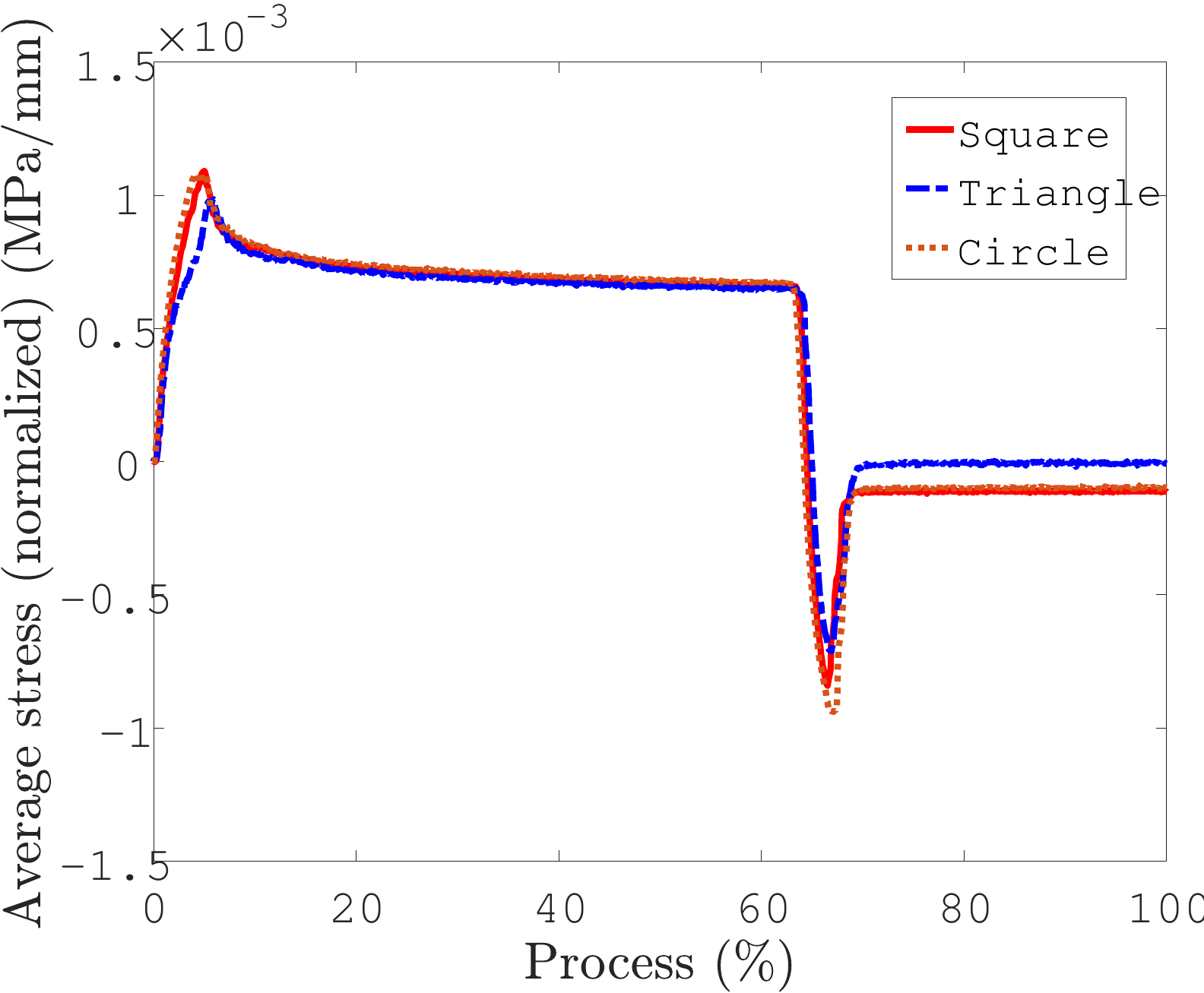}
		\label{fig:plate:a}}
		\hspace{-2mm}
	\subfigure[]{
		\includegraphics[width=1.67in]{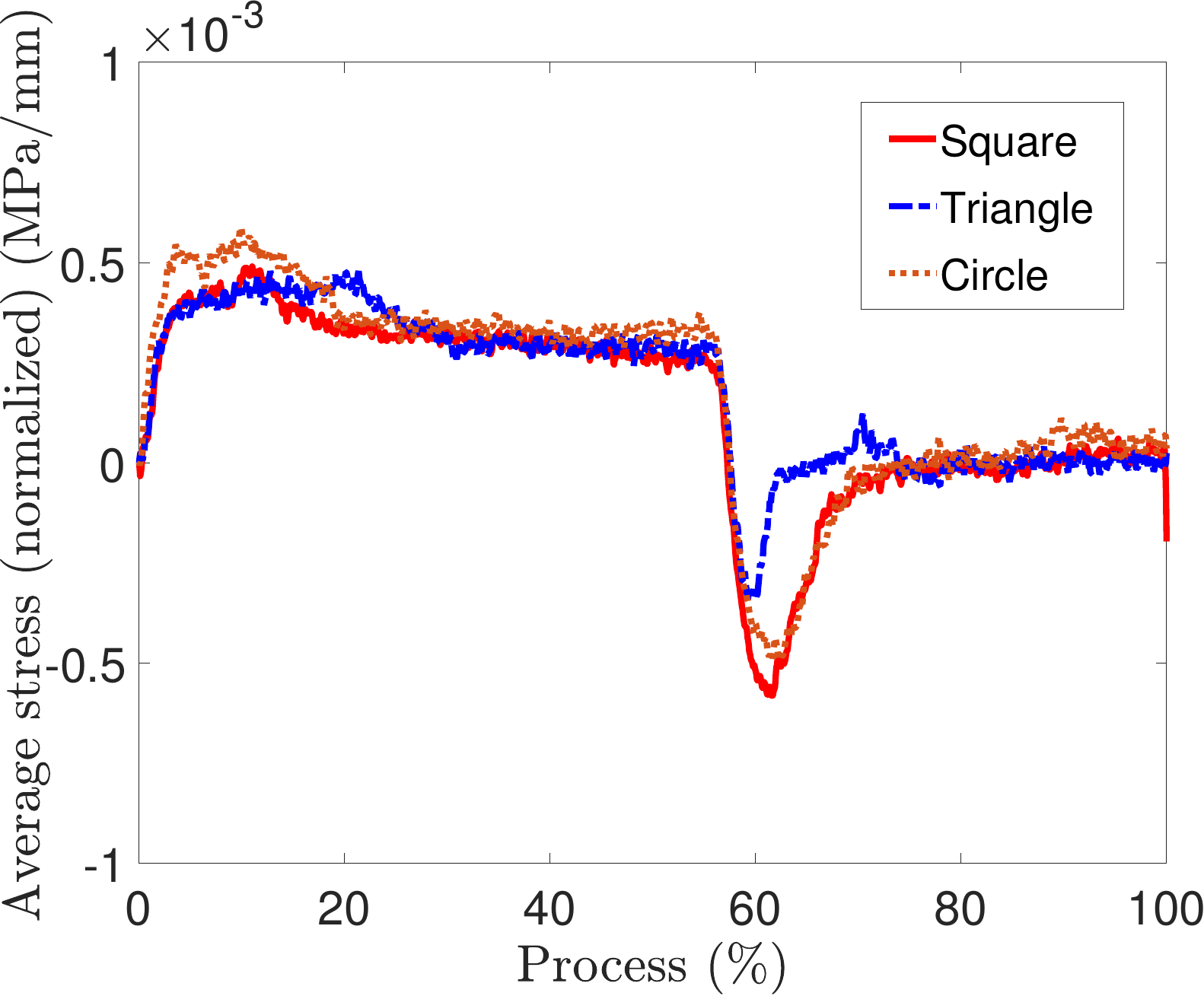}
		\label{fig:plate:b}}
		\hspace{-2mm}
	\subfigure[]{
		\includegraphics[width=1.6in]{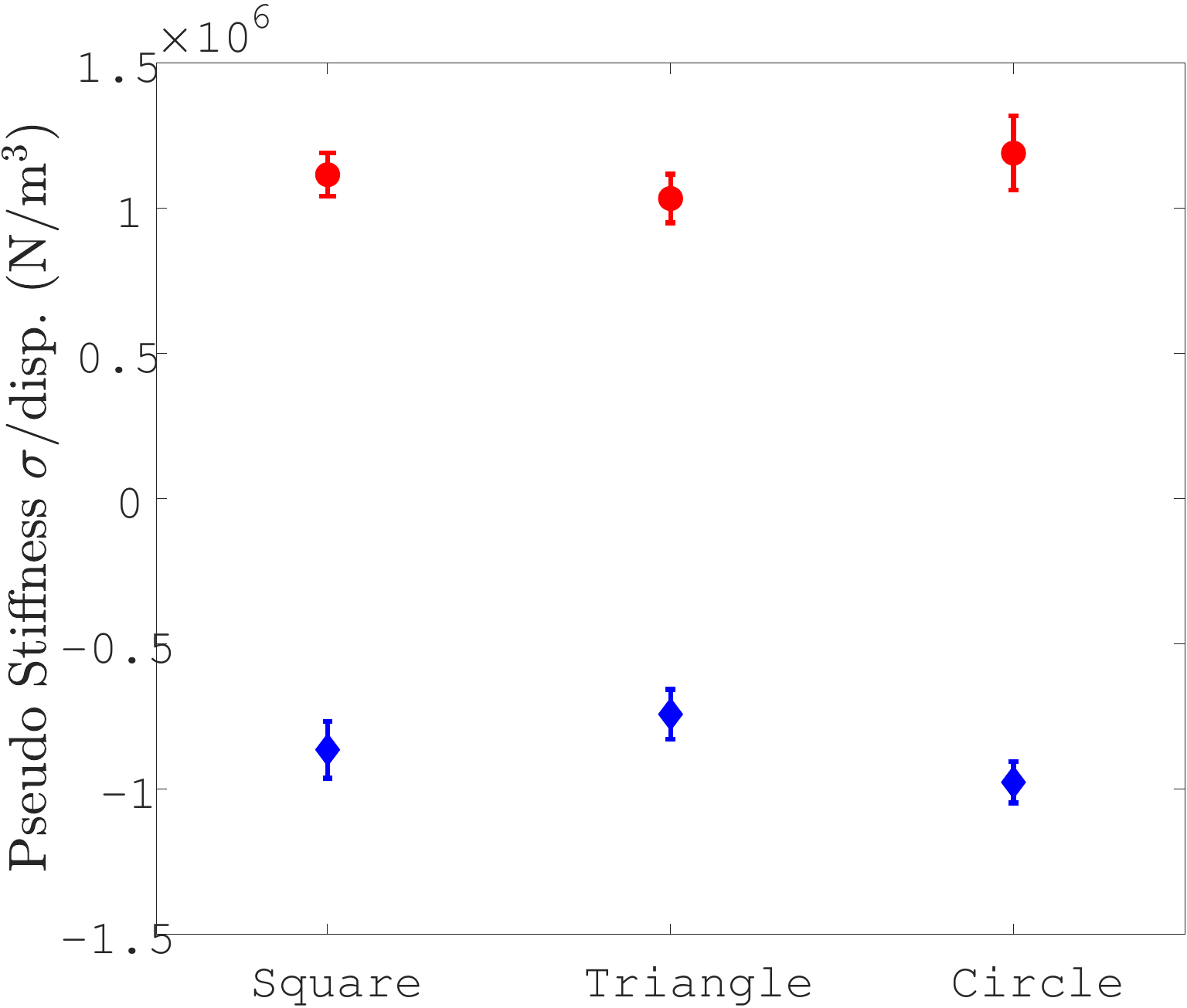}
		\label{fig:plate:c}}
		\hspace{-2mm}
	\subfigure[]{
		\includegraphics[width=1.6in]{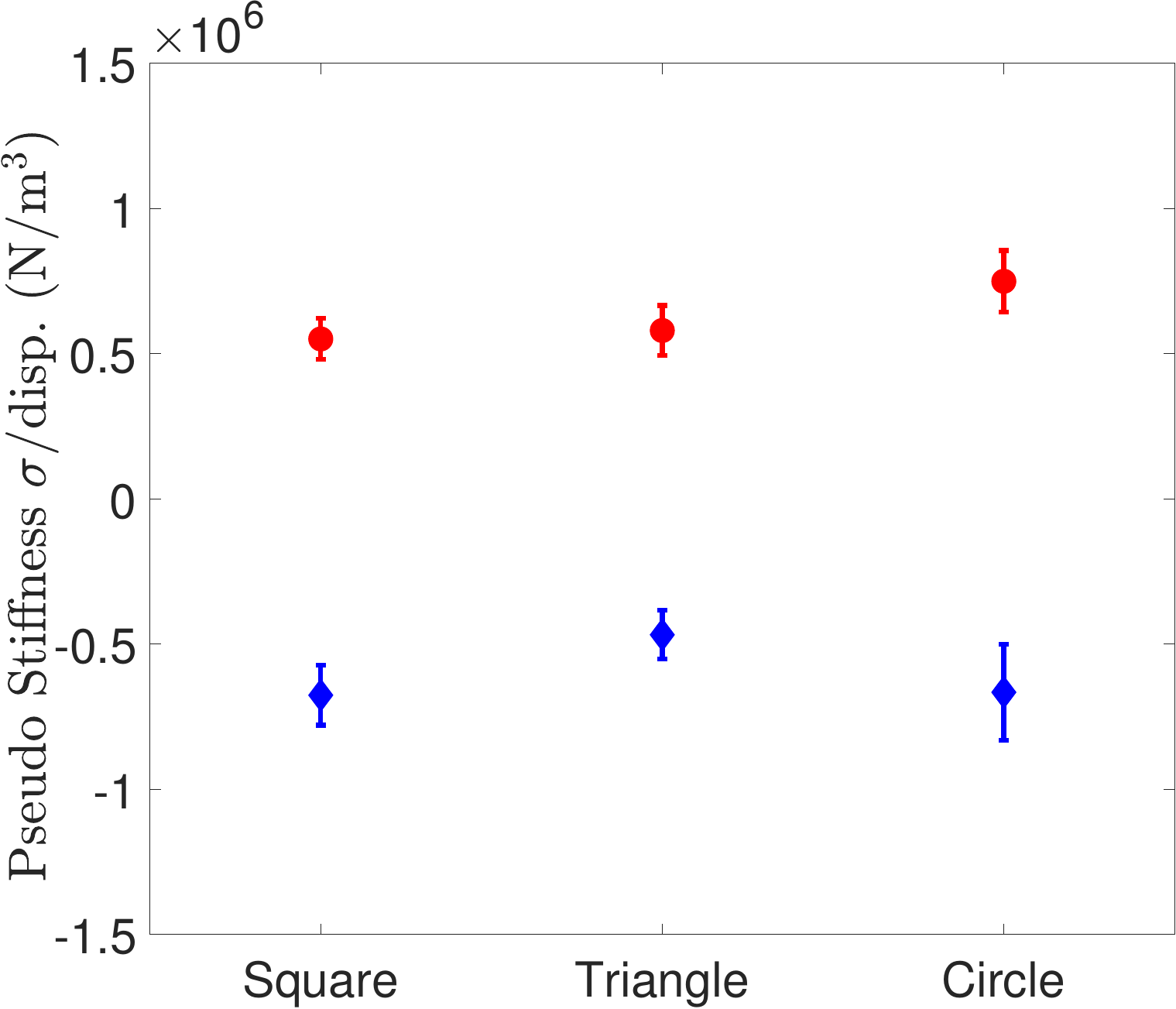}
		\label{fig:plate:d}}
		\vspace{-2mm}
	\caption{Results of the vertical and horizontal intrusion E1 and E2. (a) Vertical and (b) horizontal intrusion stresses (normalized by the corresponding maximum displacements). Pseudo-stiffness of the mud media in the (c) vertical and (d) horizontal directions (red: intrusion resistive; blue: suction).}
	\label{fig:resultPlate}
\vspace{-0mm}
\end{figure}

For experiments E1 and E2 with the plate of different shapes, Fig.~\ref{fig:resultPlate} shows the average resistive stress results of the single plate intrusion in the vertical and horizontal directions. We normalized the surface stress by the maximum intrusion displacement in the corresponding direction. We further defined the pseudo-stiffness of mud in both the vertical and horizontal directions by using the maximum intrusion/suction stress divided by the penetration/retraction displacement. Figs.~S\ref{fig:plate:a} and~S\ref{fig:plate:b} show the stress profiles in the vertical and horizontal directions, respectively. The results confirm a clear consistent pattern: immediate increase of resistance when pushing the mud media followed by a stress relaxation plus a stable plateau when the intruder was stationary. Suction happened and held when retraction started until the mud media broke. From these results, we found that the resistive stress of mud does not possess significant shape-dependency. Figs.~S\ref{fig:plate:c} and~S\ref{fig:plate:d} show the pseudo-stiffness of the intrusion resistive and suction in the vertical and horizontal directions, respectively. The results also indicate a small variance regarding the shape of the intruder plates.

\xunjie{First, we clarify that the model parameters are grouped according to the physical phases of the foot-mud interaction process, rather than being interpreted as an undifferentiated set. They are: (a) parameters governing the immediate resistive response during intrusion, $\Pi_{\mathrm{b}} = \{\alpha,n\}$; (b) parameters associated with time-dependent thixotropic behavior such as stress relaxation and hysteresis, $\Pi_{\mathrm{th}} = \{G_{\mathrm{m}},\eta_{\mathrm{m}},\eta_{\infty},\dot{\gamma}_c\}$ where we define $\lambda = \eta_\mathrm{m}/G_\mathrm{m}$, $\dot{\gamma}_c= k_a/k_r$, and yielding stress is $\sigma_{Y} = (\eta_{\mathrm{m}}-\eta_{\infty})\dot{\gamma}_c$; and (c) parameters of suction, $\Pi_{\mathrm{s}} = \{\tau_{\mathrm{build}},\tau_{\mathrm{leak}},\varepsilon, \gamma_0\}$. We acknowledge parameter sets $\Pi_{\mathrm{b}}$, $\Pi_{\mathrm{th}}$, and $\Pi_{\mathrm{s}}$ are intrinsic to mud media as they are associated with mud conditions such as substrate composition, water content, loading/shearing direction. The experimental results also indicated the small variance with respect to intruder plate shape; see Fig.~\ref{fig:resultPlate}. By contrast, the angular scaling functions $f_{\mathrm{n}}$ and $f_{\mathrm{t}}$ are considered as directional, empirical weighting terms. They are used to extend the mud infinitesimal-plate-scale stress to the resultant 3D force integration. Tables~\ref{tab:mdlPara_Z} and~\ref{tab:mdlPara_XY} list the parameter set for the vertical and horizontal direction, respectively.}

\renewcommand{\arraystretch}{1.2}
\setlength{\tabcolsep}{0.05in}
\begin{table}[h!]
	\centering
	\caption{List of model parameters for the vertical intrusion; $W$: water content}
    \vspace{-2mm}
	\label{tab:mdlPara_Z}
	\begin{tabular}{ccccccccccc}
	\toprule[1.2 pt]
	Water content & $\alpha$ (MPa) & $n$ & $\lambda$ & $\eta_{\mathrm{m}}$ (MPa/s) & $\eta_{\infty}$ (MPa/s) & $\dot{\gamma}_c$ (s$^{-1}$) & $\tau_{\mathrm{build}}$ (s) & $\tau_{\mathrm{leak}}$ (s) & $\varepsilon$ & $\sigma_{Y}$ (MPa)\\
			\midrule
		$W=25\%$ & 0.024 & 0.53 & 1.9 & 0.067 & 0.014 & 0.14 & 0.32 & 0.17 & 0.016 & 0.0071\\
		$W=35\%$ & 0.013 & 0.20 & 3.2 & 0.27 & 0.0031 & 0.042 & 0.49 & 0.089 & 0.054& 0.011\\
		$W=45\%$ & 0.010 & 0.18 & 2.6 & 0.31 & 0.0026 & 0.027 & 0.58 & 0.052 & 0.040 & 0.0086\\
	\bottomrule[1.2pt]
	\end{tabular}
\end{table}

\renewcommand{\arraystretch}{1.2}
\setlength{\tabcolsep}{0.055in}
\begin{table}[h!]
	\centering
	\caption{List of model parameters for the horizontal intrusion; $W$: water content}
    \vspace{-2mm}
	\label{tab:mdlPara_XY}
	\begin{tabular}{cccccccccccc}
	\toprule[1.2 pt]
	Water content & $\alpha$ (MPa) & $n$ & $\lambda$ & $\eta_{\mathrm{m}}$ (MPa/s) & $\eta_{\infty}$ (MPa/s) & $\dot{\gamma}_c$ (s$^{-1}$) & $\tau_{\mathrm{build}}$ (s) & $\tau_{\mathrm{leak}}$ (s) & $\varepsilon$ & $\sigma_{Y}$ (MPa)\\
			\midrule
		$W=25\%$ & 0.018 & 0.12 & 1.8 & 0.46 & 0.010 & 0.31 & 0.63 & 0.99 & 0.027 & 0.013 \\
		$W=35\%$ & 0.011 & 0.32 & 17 & 0.41 & 0.021 & 0.019 & 0.25 & 0.67 & 0.14 & 0.0094\\
		$W=45\%$ & 0.0078 & 0.95 & 8.7 & 0.63 & 0.065 & 0.028 & 0.21 & 0.022 & 0.12 & 0.0061\\
	\bottomrule[1.2pt]
	\end{tabular}
\end{table}

\begin{figure}[h!]
\centering
	\subfigure[]{ \includegraphics[width=3.20in]{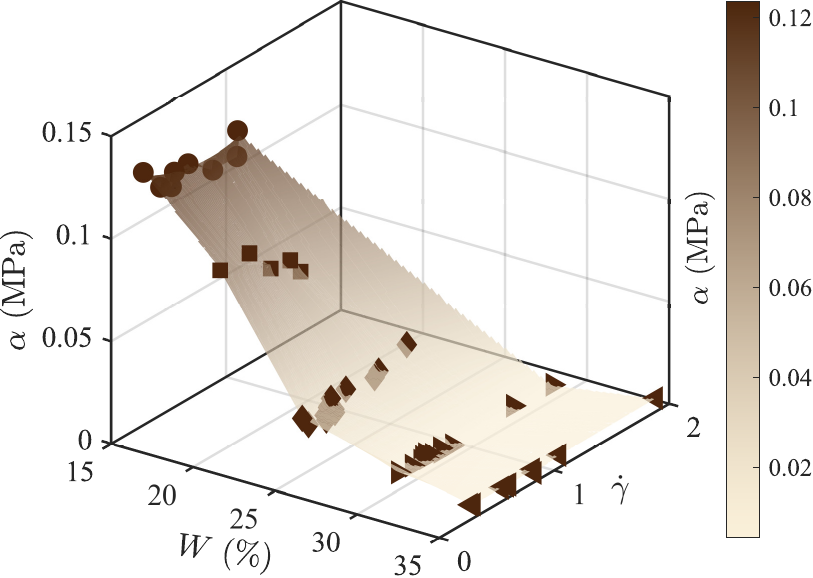}
		\label{fig:alpha}}
	\subfigure[]{
		\includegraphics[width=3.20in]{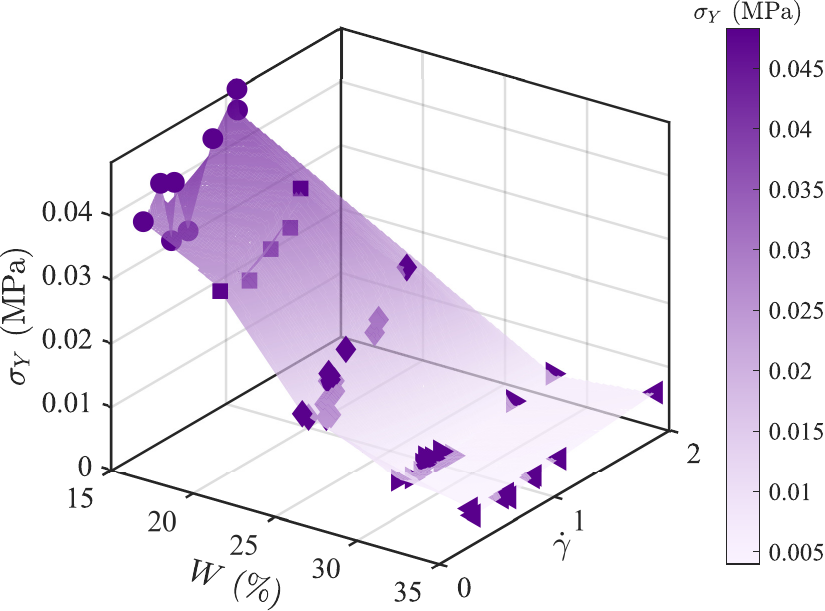}
		\label{fig:sigma}}
	\caption{\textcolor{black}{Model parameter variations under different intrusion speed ($\dot{\gamma}$) and water content ($W$) conditions. (a) Immediate resistive stiffness, $\alpha$; (b) Yielding stress, $\sigma_Y$.}}
	\label{fig:para}
\end{figure}

\xunjie{We found that mud stress became more sensitive to immediate resistance stiffness $\alpha$ of the vertical direction because of the gravity effect and compaction caused by the bottom. This leads to the result that $\alpha$ of the vertical direction is greater than that of the horizontal direction.
For the horizontal direction, mud substrates exhibited more fluidity such that the shearing viscosity became the main source of resistance in mud reflected in the larger magnitudes of $\eta_{\mathrm{m}}$ and $\eta_{\infty}$; see Tables~\ref{tab:mdlPara_Z} and~\ref{tab:mdlPara_XY}. This demonstrates that the mud stress is more sensitive to shearing viscosity in the horizontal direction.
By variations of mud conditions, i.e., the water content $W$, and shear speed represented by the dimensionless speed $\dot{\gamma}.$, Fig.~\ref{fig:para} shows the sensitivity of $\alpha$ and $\sigma_Y$. It is found that two mud-intrinsic parameters were not sensitive to the shear speed of the reported regime. However, water content considerably impacted the magnitude. The magnitudes of $\alpha$ and $\sigma_Y$ dropped significantly by $95\%$ and $88\%$, respectively, when the water content increased from $15\%$ to $35\%$. Therefore, they can be considered as the sensitive parameters requiring dedicated identifications for online deployment.}

From the above results, it is found that the critical displacement rate $\dot{\gamma}_c$ is much smaller than the foot locomotion velocity, that is, $\dot{\gamma}_c\ll\dot{\gamma}$. This means the mud viscosity becomes dominant for high-speed locomotion. Therefore, the mud thixotropy of the quasi-static condition is assumed to be $\sigma_{\mathrm{th,ss}}=\eta_\infty \dot{\gamma}$ with $\xi_{\mathrm{ss}}=0$. By this approximation, we obtain the total resistive stress for the intrusion as $\sigma_{\mathrm{tot}}\approx\alpha \gamma ^n + \eta_\infty \dot{\gamma}$, which is linear with parameters $\alpha$ and $\eta_\infty$. For the retraction of the foot from mud media, the maximum suction force is of interest and we assume that the sealing state function $\phi(\gamma)=1$ such that the suction pressure is $\sigma_{\mathrm{s}}=(1-e^{-\Delta t/\tau_{\mathrm{build}}})\sigma_{Y}$. Then, the total resistive stress for the retraction is $\sigma_{\mathrm{tot}} \approx (1-e^{-\Delta t/\tau_{\mathrm{build}}})\sigma_{Y} + \eta_\infty \dot{\gamma}$, which is also linear with parameters $\sigma_Y$ and $\eta_\infty$. With the above reasoning, model parameters $\Pi = \{\alpha,~\sigma_Y,~\eta_\infty,~n,~\tau_{\mathrm{build}}\}$ of primary interests can be potentially estimated through online identification process. \xunjie{Online} terrain classification and model parameter estimation is one of ongoing research tasks.

\subsection{High water content of mud}
\begin{figure}[h!]
\centering
	\subfigure[]{ \includegraphics[width=3.00in]{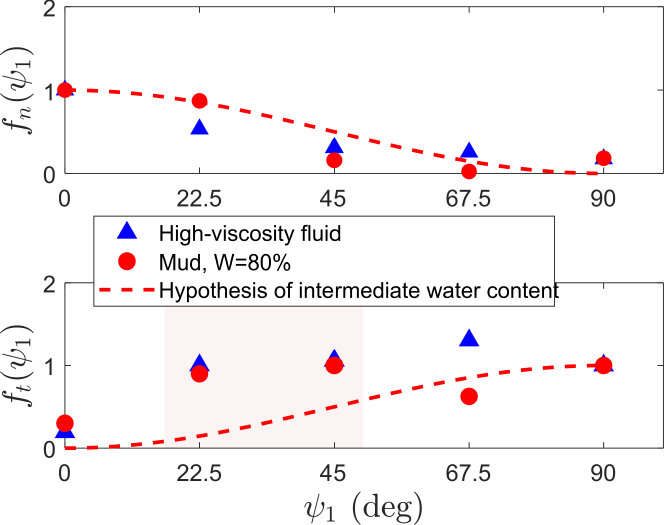}
		\label{fig:fnft:a}}
\hspace{2mm}
	\subfigure[]{
\includegraphics[width=3.00in]{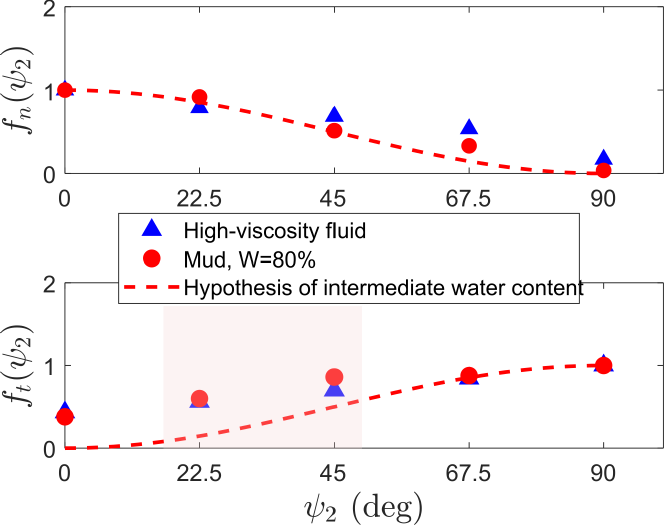}
		\label{fig:fnft:b}}
	\caption{\textcolor{black}{Two dimensionless scaling factors $f_\mathrm{n}$ and $f_\mathrm{t}$ for (a) vertical and (b) horizontal intrusion, respectively. Red dashed curves are empirical modeling of intermediate water content.}}
	\label{fig:fnft}
\end{figure}

\xunjie{We considered that $35\%$ was close to the upper limit beyond which the mud no longer provides sufficient supporting resistance, leading to the severe leg entrapment as a locomotion failure~\cite{liuRAL2023adaptation,ChenAIM24}. Therefore, we selected this range of primary and practical interest to reveal the most relevant and effective mud resistive mechanism.
We assume sufficiently low stiffness, low yielding stress, and dominated viscosity as primary features of mud if water content becomes really high.
Especially, the hypothesized empirical formulations of two scaling factors $f_{\mathrm{n}}$ and $f_{\mathrm{t}}$  vary and need to be revised. Therefore, we further implemented vertical- and horizontal-plate intrusion experiments with a high water content to check whether these hypothesises hold.

We selected $80\%$ water-content mud as a representative high-water-content case and compared with a classic high-viscosity liquid, i.e., gel. Fig.~\ref{fig:fnft} shows the compared results. We observed that normal and tangential force characteristics of high-water-content mud were very closed to the gel and the proposed $f_{\mathrm{n}}$ still matched the proposed fitting form for both vertical and horizontal direction. However, for $f_{\mathrm{t}}$ in the tangential direction, there was a significant discrepancy in the low penetration angle range; see the shaded regions in Fig.~\ref{fig:fnft}. This actual high scale of the tangential stress primarily shows that the shearing viscosity becomes dominated as water content increases.

The aforementioned results indicate that mud rheology under high water content possibly shares the similar modeling of classic fluid with high viscosity that has been well studied. This also confirmed the necessity of special modeling of mud in the intermediate water content regime, while rigorous modeling is still needed to identify the water-content-related boundary at which mud media changes the rheology significantly.}
\subsection{Morphing foot intrusion experiments}

\begin{figure}[h!]
	\centering
	\subfigure[]{
		\includegraphics[width=2.5in]{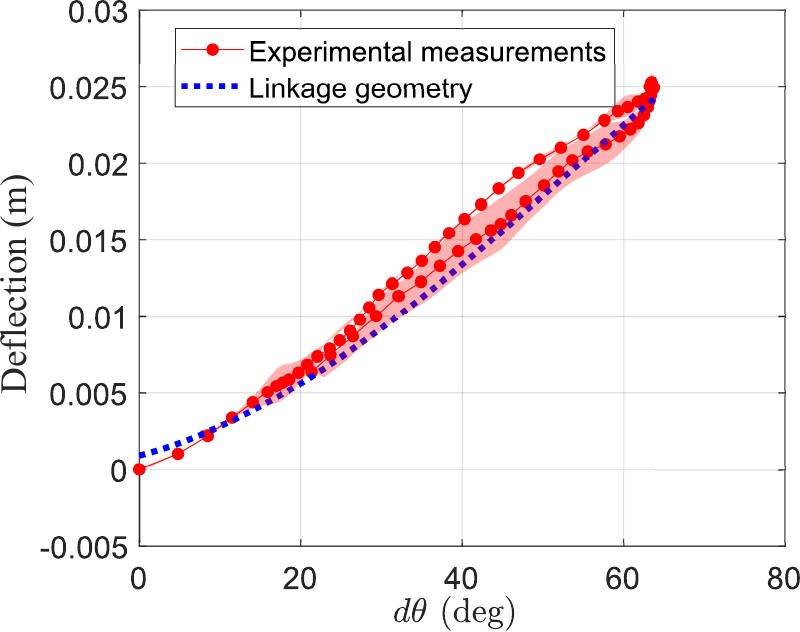}
		\label{fig:footCali:a}}
		\hspace{3mm}
	\subfigure[]{
		\includegraphics[width=2.55in]{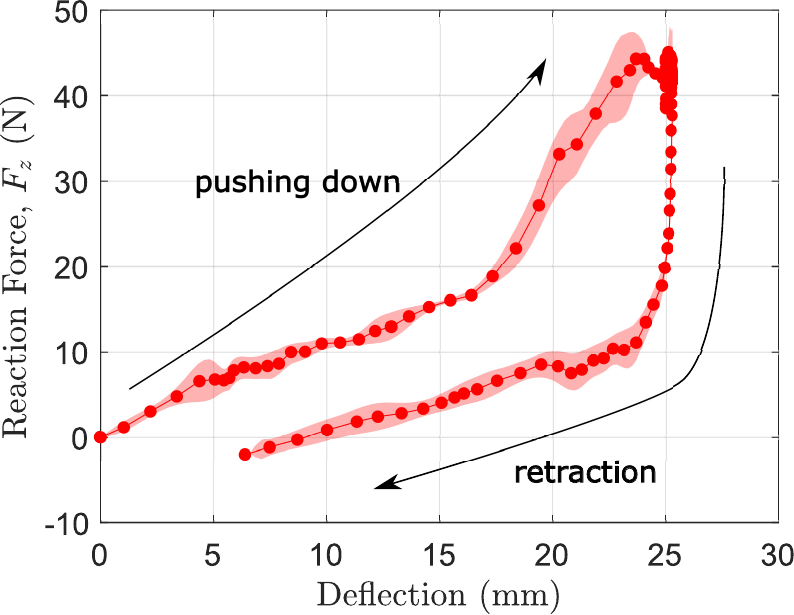}
		\label{fig:footCali:b}}
		\vspace{-3mm}
	\caption{Results of the morphing foot on the rigid ground and in mud. (a) Pad extending angle versus foot deflection. (b) Vertical resistive force versus foot deflection. }
	\label{fig:footCali}
		\vspace{-0mm}
\end{figure}

Using the morphing foot, we conducted vertical intrusion experiments on the rigid ground surface to characterize the contact stiffness. We recorded the pad expanding angle $d\theta$, the corresponding foot deflection, and the resultant vertical force $F_z$. Fig.~S\ref{fig:footCali:a} and S\ref{fig:footCali:b} show the foot deflection and foot stiffness profiles, respectively. The foot deflection is linear with the deformation angle, as shown in Fig.~S\ref{fig:footCali:a}. Because of the two cascaded springs used in the morphing foot design, during the foot touch-down, the soft spring was compressed first and then the harder one. We clearly observed two different stiffness regions as shown in Fig.~S\ref{fig:footCali:b}. The variable stiffness is attractive since it enables a  foot pad rapid expansion under a small contact force and a quick retraction under hard spring compression during foot lifting from the mud media.

\begin{figure}[h!]
	\vspace{-2mm}
	\centering
	\subfigure[]{
		\includegraphics[width=2.8in]{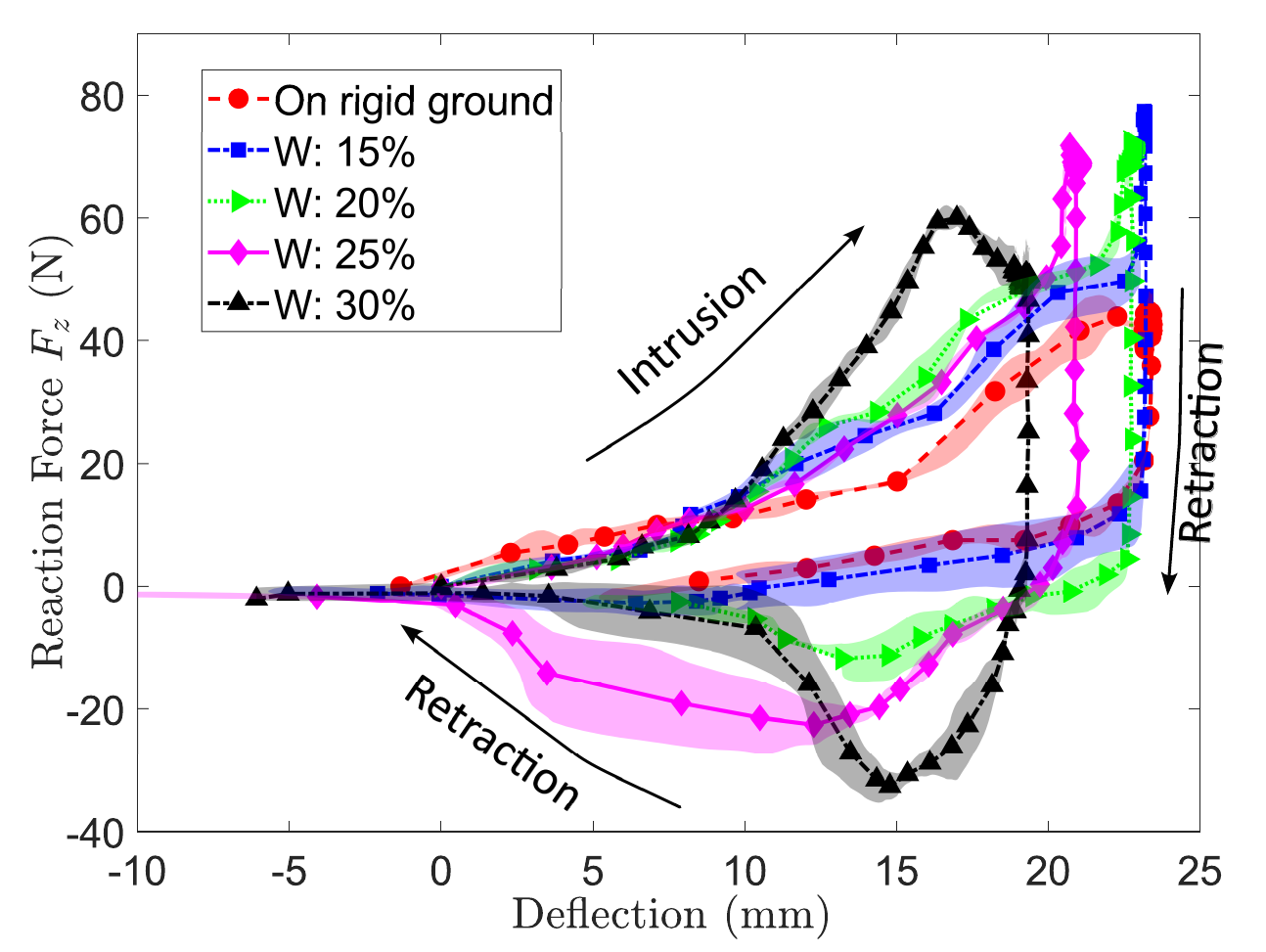}
		\label{fig:footcomp:a}}
	\subfigure[]{
		\includegraphics[width=2.9in]{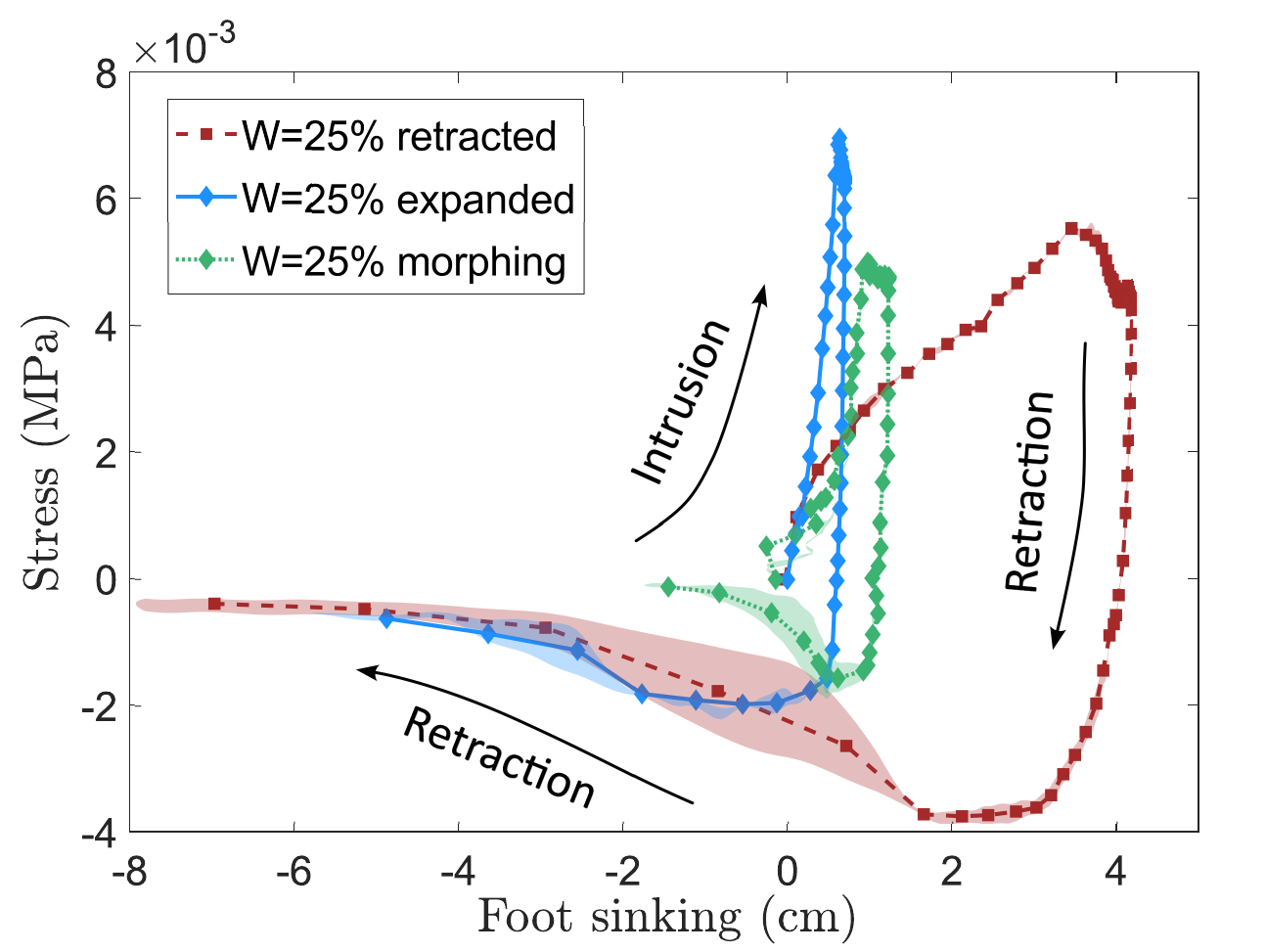}
		\label{fig:footcomp:b}}
	\vspace{-3mm}
	\caption{(a) Vertical resistive force comparisons of the morphing foot on the rigid ground surface and the muddy terrain with different water contents. (b) Reaction stress of contact surface of the morphing foot with three positions: fully retracted, fully expanded, and morphing on mud. Shaded areas represent the one-standard deviation.}
	\label{fig:footcomp}
	\vspace{-0mm}
\end{figure}

To observe the morphing foot performance, we conducted single-foot 1D (vertical) intrusion experiments on rigid ground surface and on mud with various water contents (from $15$\% to $30$\%). Fig.~S\ref{fig:footcomp:a} shows the vertical resistive force versus the vertical deflection of the morphing foot. Compared with on the rigid ground surface, the  morphing foot on muddy terrain with $15$\% and $20$\% water content fully expanded and generated much larger forces (almost doubled). However, when the water content increased to $25$\% and further to $30$\%, the morphing foot partially expanded, that is, deflection was smaller than that on the rigid ground surface, and the resistive forces were reduced but still higher than that on the rigid ground surface. Significantly large negative suction forces were also observed on muddy terrain with increased water content.

To further demonstrate the efficacy the morphing foot design, we repeated single-foot 1D (vertical) intrusion experiments on the ($25$\% water content) mud with three modes: foot was fixed at the fully expanded position (``expanded''), at the fully retracted position (``retracted''), and freely morphing mode (``morphing''). Fig.~S\ref{fig:footcomp:b} shows the vertical resistive stress versus the foot sinking depth. It is clear that the sinking depth was restricted with the large contact surface of the ``expanded'' mode. The ``retracted'' foot suffered a large sinking depth and the negative suction force in both the intrusion and retraction phases. This is because of the small surface contact area and added mud material mass on pads. The ``morphing'' mode clearly showed two advantages: a significant reduction of sinking on mud because of a quick expansion of pads increasing contact surface, and a reduction of the suction due to an active retraction of the pads breaking the sealing of the cavity underneath the foot during retraction.

\subsection{Gait energy and balance analysis of dynamic walking}

\begin{figure}[h!]
\centering
	\subfigure[]{
		\includegraphics[width=3.0in]{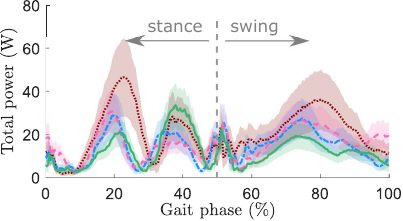}
		\label{fig:walking:a}}
	\subfigure[]{
		\includegraphics[width=3.0in]{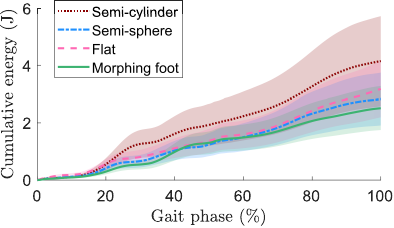}
		\label{fig:walking:b}}
	\subfigure[]{
		\includegraphics[width=6.5in]{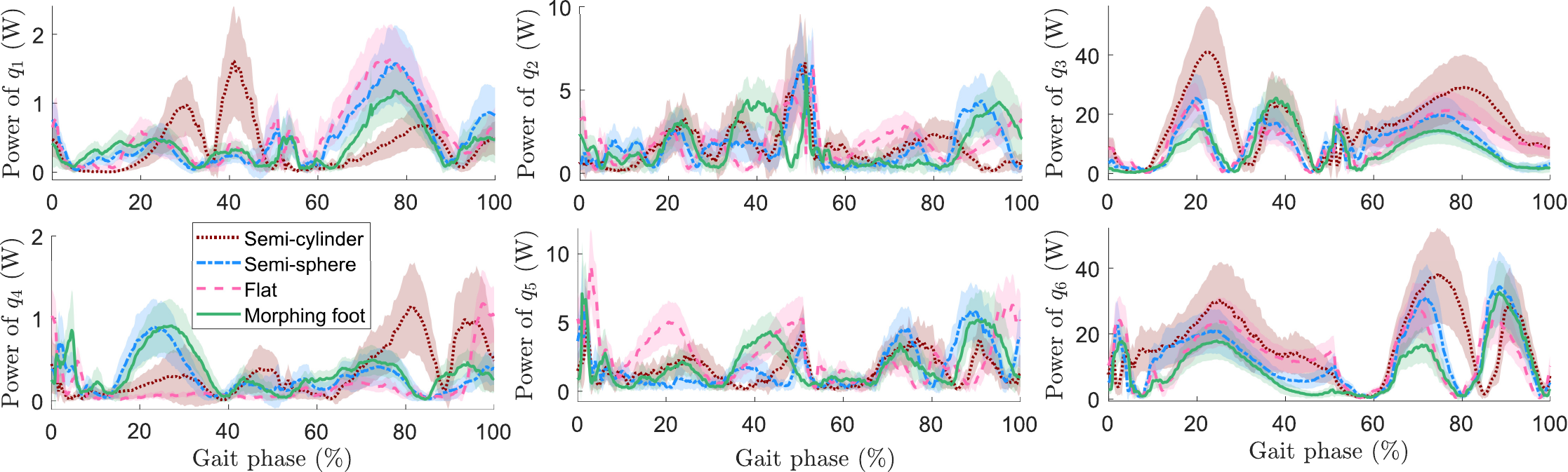}
		\label{fig:walking:c}}
		\vspace{-3mm}
	\caption{Results of dynamic walking experiments in mud. (a) Total power consumed by six joint actuators. (b) Cumulative energy in one gait phase. (c) Separated joint actuator power. $q_i$, $i=1,\ldots,6$ are joint angles of the robot lower-limb. Shade areas represent one-standard deviation.}
	\label{fig:walking}
\end{figure}

\xunjie{First,} we provided further details about the energy cost during biped dynamic walking on muddy terrain. Fig.~\ref{fig:walking} shows the comparisons of power and cumulative energy consumed by the joint actuators in one walking gait. Fig.~S\ref{fig:walking:a} shows the total power of the robot. It is clear that using the morphing foot leveraged the least power in the early stance phase and the entire swing phase with a high power in the late stance phase. As a result, the biped benefited from the morphing foot by the lowest total energy cost in the one gait cycle, as shown in Fig.~S\ref{fig:walking:b}. Fig.~S\ref{fig:walking:c} further shows the power profile of each joint actuator during one walking gait cycle.

\begin{figure}[h!]
  \centering
  \includegraphics[width=4.3in]{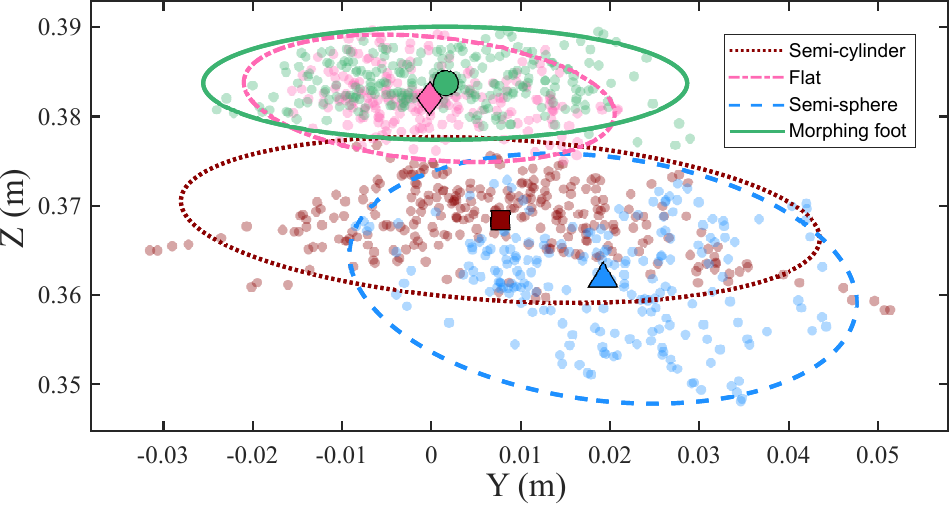}
  \caption{\textcolor{black}{Results of body sway in the robot frontal plane (i.e., Z-Y cluster of CoM) using different foot configurations.}}\label{fig:ZYsway}
\end{figure}

\begin{figure}[h!]
\centering
	\subfigure[]{ \includegraphics[width=4.4in]{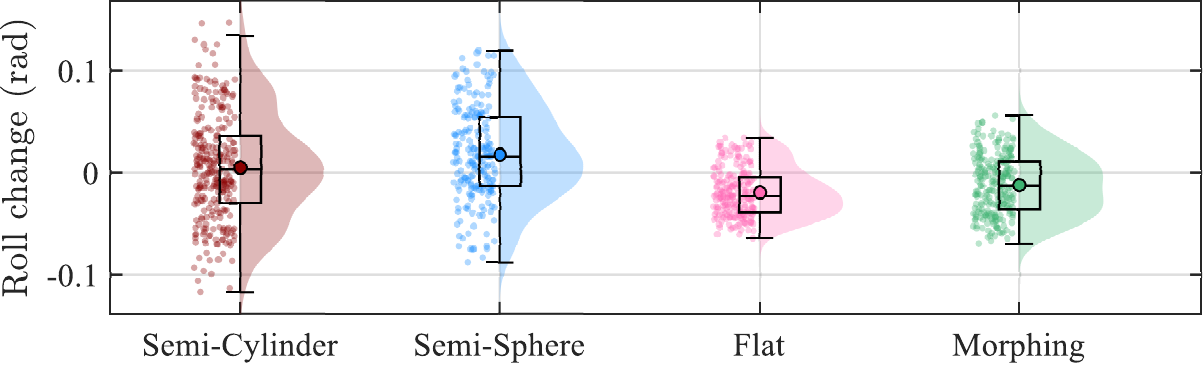}
		\label{fig:roll}}
	\subfigure[]{
		\includegraphics[width=4.4in]{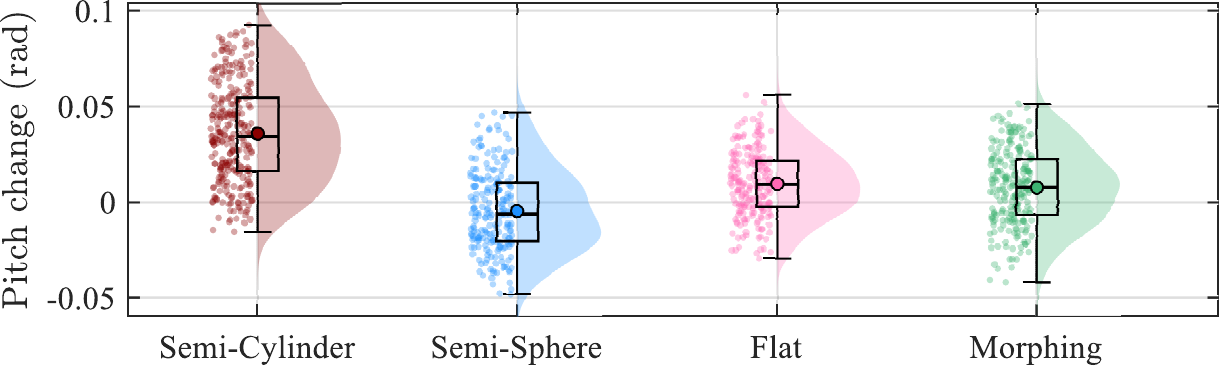}
		\label{fig:pitch}}
	\subfigure[]{
		\includegraphics[width=4.4in]{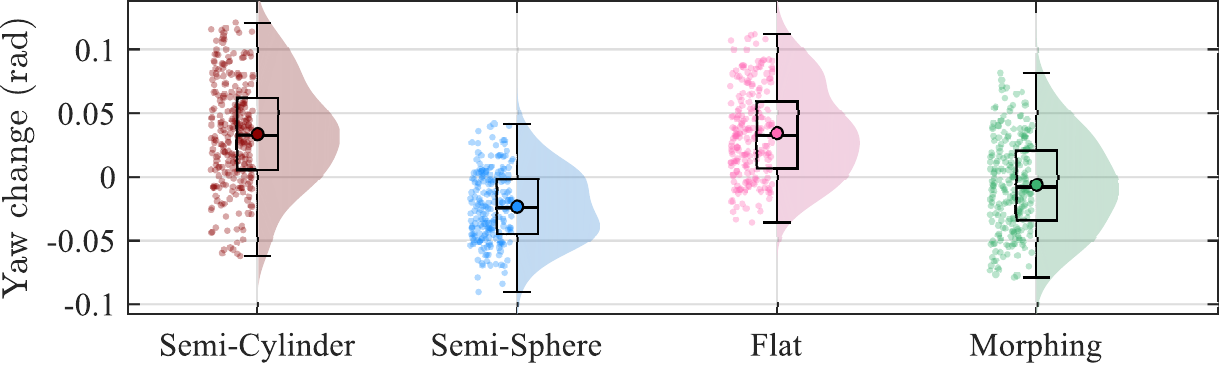}
		\label{fig:yaw}}
	\caption{\textcolor{black}{Results of body posture variation, namely, roll, pitch, and yaw deviations, using different foot configurations. (a) Roll; (b) Pitch; (c) Yaw.}}
	\label{fig:posture}
\end{figure}

\xunjie{Then, we acknowledge that the balance-related locomotion performance is also highly relevant. We also provided the recorded body sway in the frontal plane and posture variation (i.e., roll, pitch, and yaw deviations) as two representative aspects to evaluate the effect of the foot on walking balance. Fig.~\ref{fig:ZYsway} shows body sway results in the robot frontal plane of walking locomotion under the mud water content $25\%$. It is clear to see that the robot generally experienced the least sinking using the  morphing foot followed by using the flat foot. The semi-spherical foot led to the most significant sinking. This is consistent to the results shown in Fig.~12 in the manuscript. We also observed that the morphing foot moderated the range of the body sway compared to the semi-cylindrical and semi-spherical foot, while the flat foot had the best performance (i.e., the smallest sway area) in this sense.}
\xunjie{Fig.~\ref{fig:posture} shows the corresponding posture variation results of the robot body during walking. It shows that the morphing foot enabled the robot to have the smallest oscillations from the aspects of roll and yaw motions, while it performed similarly to the semi-spherical foot in terms of the pitch motion. Overall, the morphing foot still possesses the capability of balancing the body motion although it did not outperform considerably among all foot shapes for the posture variation.}

\xunjie{We consider that reducing excessive foot sinking and suction force in the mud can improve the stance consistency and reduce disturbance during foot retraction and swing. This potentially in turn benefits the whole-body balance and gait stability of mud-walking locomotion. However, it should be noted that the walking gait significantly affects the balance and energy consumption. The optimal locomotion gait design and control with the systematic evaluation of balance on muddy terrains are important topics for our future work. }

\subsection{Model comparisons}
\begin{figure}[h!]
\hspace{-2mm}
\subfigure[]{
	\label{fig:walkingForceComparison:a}
	\includegraphics[width=2.25in]{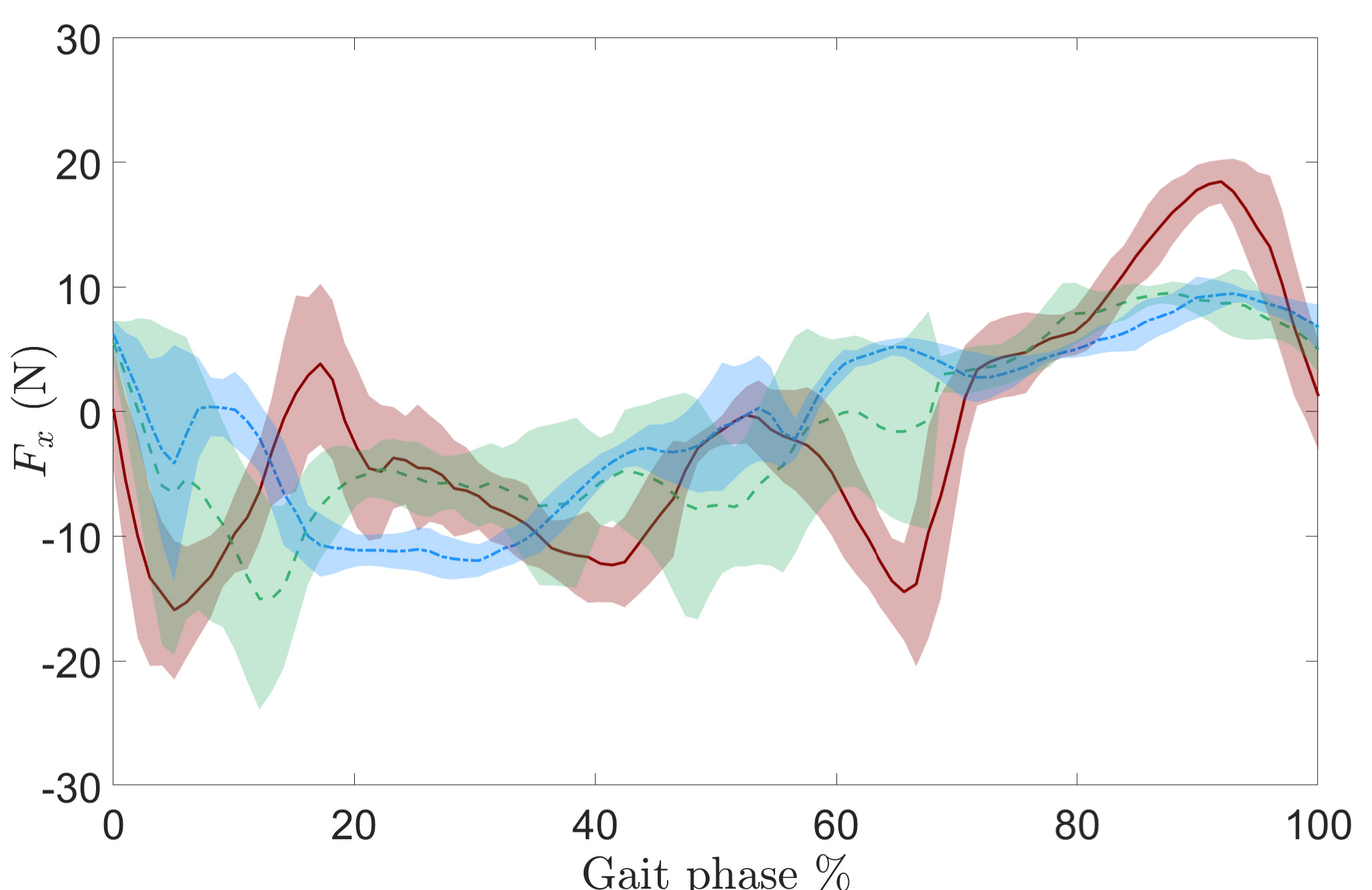}}
\hspace{-2mm}
\subfigure[]{
	\label{fig:walkingForceComparison:b}
	\includegraphics[width=2.25in]{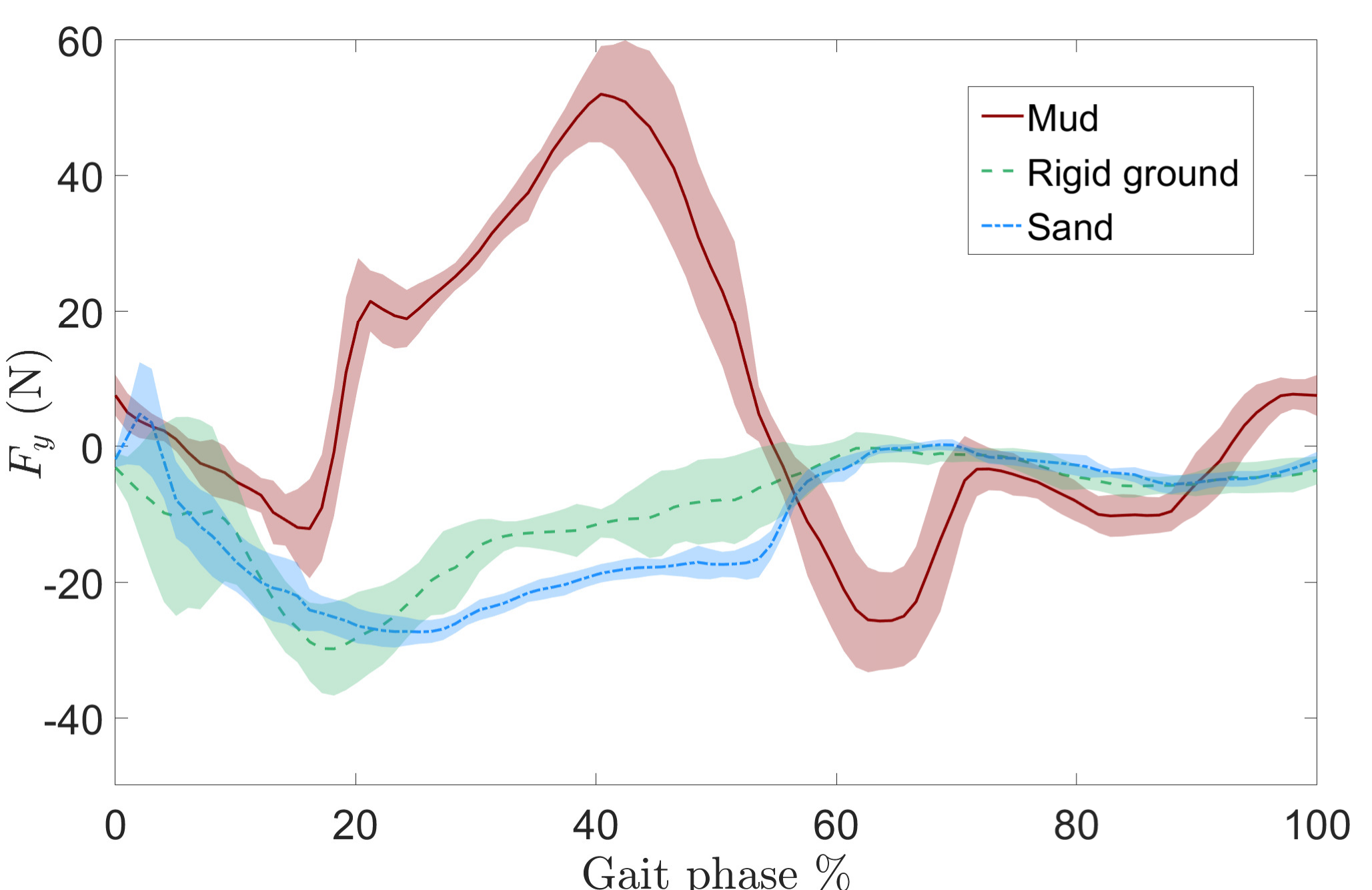}}
\hspace{-2mm}
\subfigure[]{
	\label{fig:walkingForceComparison:c}
	\includegraphics[width=2.25in]{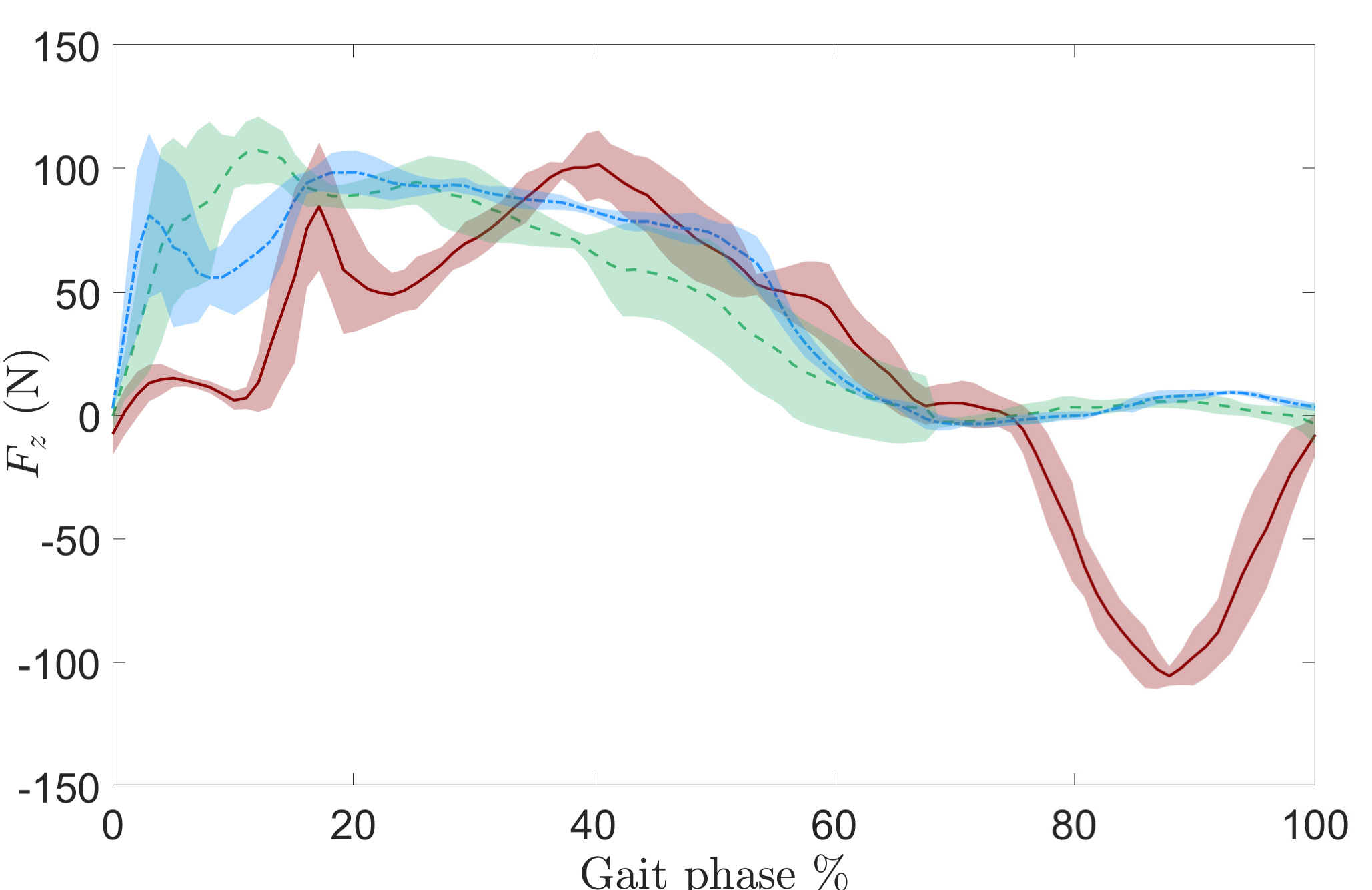}}
\vspace{-2mm}
	\caption{Comparisons of walking reaction forces on mud, sand and rigid ground surfaces. (a) $F_z$. (b) $F_y$, (c) $F_z$.}
	\label{fig:walkingForceComparison}
\end{figure}

We finally compared the mud resistance behavior with representative yielding terrains, i.e., sand and regular rigid ground surfaces. We conducted bipedal robot walking on three aforementioned terrains and extracted reaction forces in one gait cycle. Fig.~\ref{fig:walkingForceComparison} shows the corresponding three-dimensional (3D) force pattern comparisons. As discussed previously, one crucial feature of the vertical resultant force during walking on muddy terrain is the significantly negative suction effect; see Fig.~S\ref{fig:walkingForceComparison:c}. This is completely different for those on sand and rigid ground surfaces where $F_z$ decays to zero in late swing phase ($65$-$100$\%). It is also seen that there are large variations in terms of the longitudinal and lateral forces for walking in mud compared with rigid-ground and sand walking; see Figs.~S\ref{fig:walkingForceComparison:a} and~S\ref{fig:walkingForceComparison:b}. These observations confirm the necessity of an accurate foot-mud interactions model.

\begin{figure}[h!]
	\hspace{-2.5mm}
	\subfigure[]{
		\includegraphics[width=2.53in]{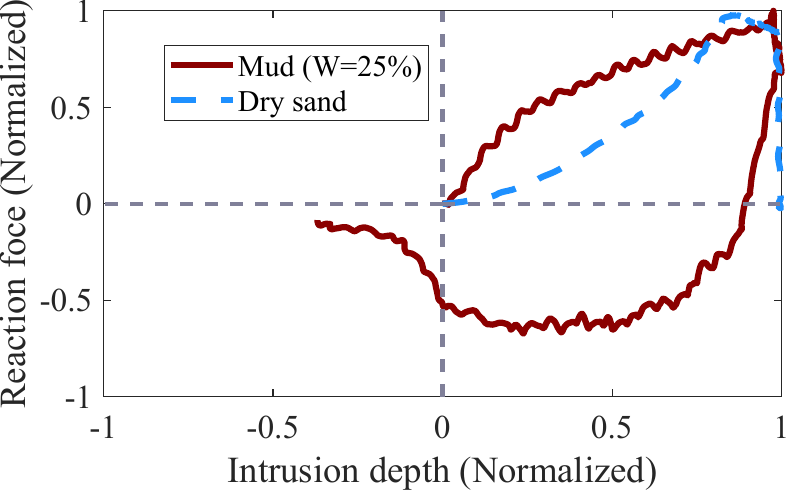}
		\label{fig:mdlComparison:a}}
	\hspace{-2.5mm}
	\subfigure[]{
		\includegraphics[width=2.06in]{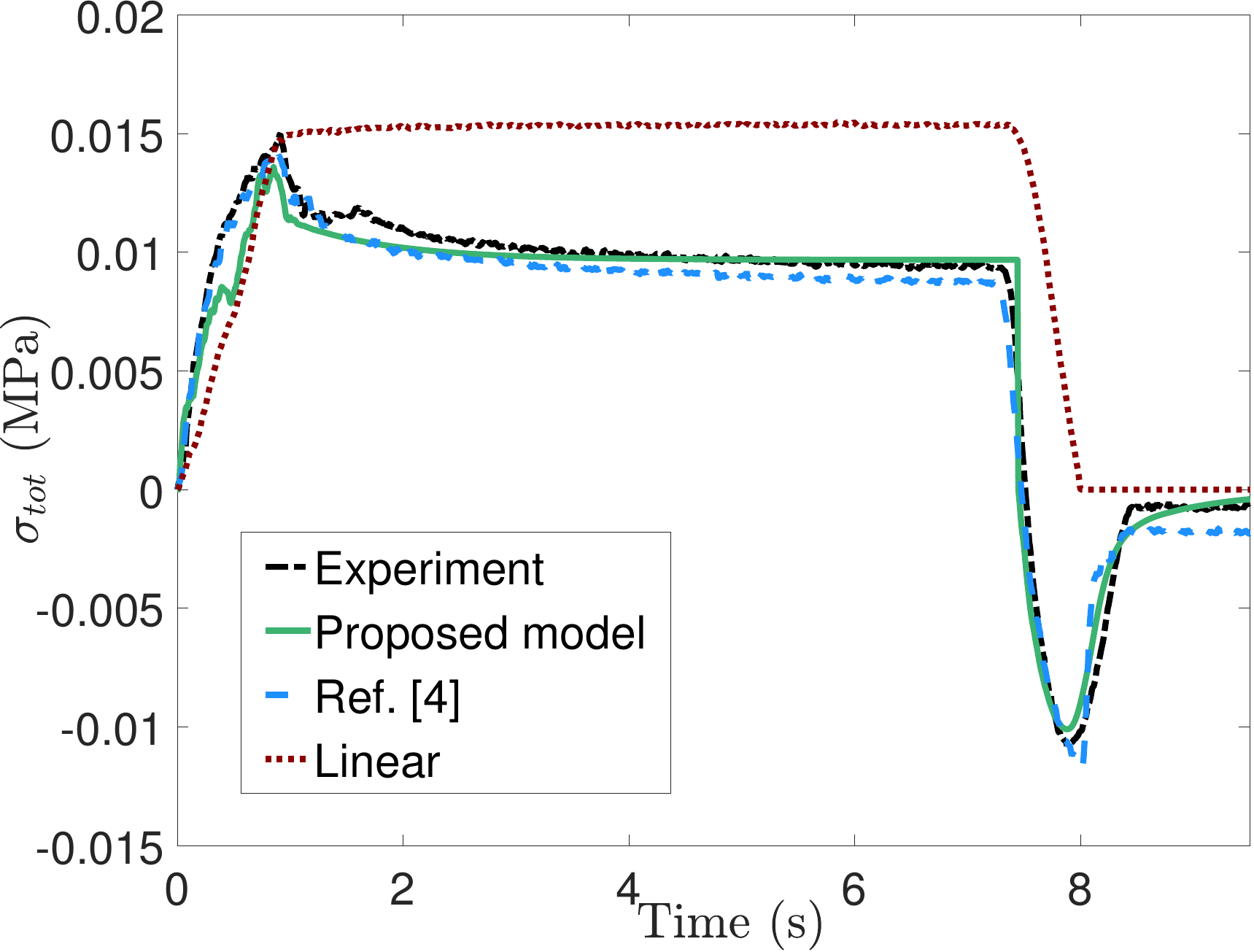}
		\label{fig:mdlComparison:b}}
	\hspace{-2.5mm}
	\subfigure[]{
		\includegraphics[width=2.1in]{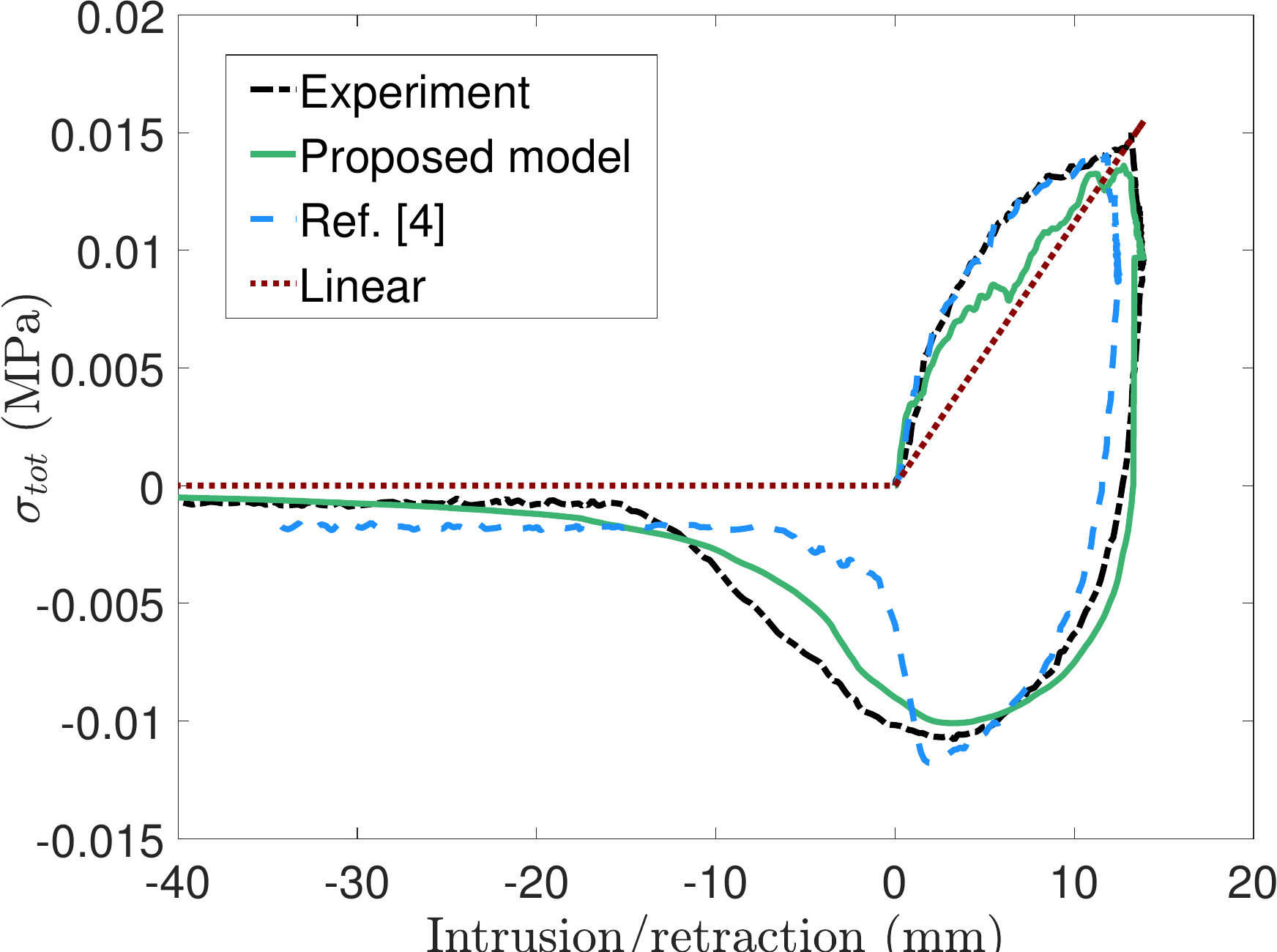}
		\label{fig:mdlComparison:c}}
\vspace{-3mm}
	\caption{Model comparisons of 1D (vertical) intrusion on yielding terrains. (a) Resistive hysteresis of stress in sand and mud. Different model predictions by the proposed model, the model in~\cite{ChenAIM24}, and the conventional linear model~\cite{GodonRAL2022} in (b) time trajectory and (c) stress-depth trajectory.}
	\label{fig:mdlComparison}
\end{figure}

For other yielding terrains such as sand, the foot-terrain interaction model mainly focused on resistive force theory (RFT)~\cite{li2013terradynamics,ChenTMECH2025Sand}. The resistance stresses at the contact interface are assumed to be proportional to the intrusion depth. We compared the resultant resistance on a horizontal plate from mud and sand in a vertical intrusion and retraction experiment. Fig.~S\ref{fig:mdlComparison:a} shows the normalized resultant force (by the maximum force magnitude) versus the (normalized) intrusion and retraction displacement. It is clear to see that there exists an obvious stress thickening for mud media during the intrusion while the sand possesses an approximately constant stiffness. This justifies the thixotropy modeling of mud in this work. Moreover, once the retraction happened, the resistance force dropped to zero immediately for sand intrusion, while suction became dominant for mud intrusion. With the featured suction effect that the granular media do not possess, the hysteresis loop is considerable for the force-depth relationship of the foot-mud interactions. Because of these unique complex nonlinear features (i.e., suction and hysteresis), the conventional RFT methods is no longer valid for applying to foot-mud interactions.

We also conducted a 1D intrusion/retraction experiment using a horizontal plate in mud and further compared the model predictions with the Maxwell-type fluid modeling in~\cite{ChenAIM24} and the conventional linear model in~\cite{GodonRAL2022} and~\cite{liuRAL2023adaptation}. Figs.~S\ref{fig:mdlComparison:b} and~S\ref{fig:mdlComparison:c} show the comparison results in time and stress-depth domains, respectively. It is clearly seen that the linear model cannot capture the thickening and stress relaxation behavior for the intrusion and stationary conditions. Moreover, the model missed the suction effect of mud for the retraction. Although the model in~\cite{ChenAIM24} captured these behaviors, the propose model has demonstrated several advantages. First, the proposed model demonstrated higher prediction accuracy than that in~\cite{ChenAIM24}; see Figs.~S\ref{fig:mdlComparison:c}. More importantly, the proposed model extended to 3D force prediction, while the model in~\cite{ChenAIM24} is only for 1D condition.

Additionally, the proposed suction pressure model by (5) in the manuscript, together with mud thixotropy, demonstrated superior properties in several aspects compared with the Maxwell-type mud fluid modeling in the previous work, including that in~\cite{ChenAIM24}. First, we interpreted the suction mechanism by describing cavity and sealing evolution from an intrinsic perspective and avoided using an artificial second-order filter as in~\cite{ChenAIM24}. The build-up and leaking effects are naturally coupled through a continuous differential equation with a smooth function indicating the sealing state. The used yield stress $\sigma_{Y}$ inherently governed by the thixotropy provided a natural threshold that is possibly estimated online, while such value has to be identified and determined a-priori by experiments in~\cite{ChenAIM24}.

\bibliography{ChenAIM_TMECH_Ref}